\documentclass{article}
\usepackage[dvipsnames,table,xcdraw]{xcolor}
\usepackage{microtype}
\usepackage{graphicx}
\usepackage{subfigure}
\usepackage{booktabs} %
\usepackage{tocbasic}
\usepackage{hyperref}

\usepackage[accepted]{icml2025}

\usepackage{amsmath}
\usepackage{amssymb}
\usepackage{mathtools}
\usepackage{amsthm}

\usepackage[capitalize,noabbrev]{cleveref}

\theoremstyle{plain}

\theoremstyle{definition}

\theoremstyle{remark}

\usepackage[textsize=tiny]{todonotes}

\usepackage{times}
\usepackage{xspace}
\usepackage{amsmath}
 \usepackage{enumitem}
\usepackage{mathtools}
\usepackage{hyperref}
\hypersetup{
    colorlinks=true,
	linkcolor=blue,
	filecolor=magenta,      
	urlcolor=blue,
	citecolor=blue,
}
\usepackage{url}

\usepackage{booktabs}
\usepackage{tabularx}
\usepackage{multicol}
\usepackage{multirow}

\usepackage{caption}
\usepackage{mwe}
\usepackage{wrapfig}
\usepackage[toc,page]{appendix}
\usepackage{mathtools}
\usepackage{titlesec}

\titlespacing{\section}{0pt}{\parskip}{0pt}
\titlespacing{\subsection}{0.5\parskip}{5pt}{0pt}
\def\Comments{0}  %

\newcommand{\tracc}{\mathsf{Acc}(f; \mathsf{Tr})}
\newcommand{\ttacc}{\mathsf{Acc}(f; \mathsf{Tst})}
\newcommand{\accd}[1][\mathcal{D}]{\mathsf{Acc}(f;#1)}

\newcommand{\consistr}{\mathsf{CR}}

\newcommand{\limem}{\mathsf{LiMem}}

\newcommand{\limemd}[1][f]{\mathsf{LiMem}(#1; \mathcal{D})}
\newcommand{\limemtr}[1][f]{\mathsf{LiMem}(#1; \mathsf{Tr})}
\newcommand{\limemtt}[1][f]{\mathsf{LiMem}(#1; \mathsf{Tst})}
\newcommand{\limemx}{\mathsf{LiMem}(f; \{x\})}

\newcommand{\kk}{K\&K\xspace}
\newcommand{\llamathree}{Llama3-8B\xspace}
\newcommand{\gptfouro}{GPT4o\xspace}
\newcommand{\gptfouromini}{GPT4o-mini\xspace}
\newcommand{\claude}{Claude-3.5-Sonnet\xspace}
\newcommand{\geminiflashtwo}{Gemini-1.5-Flash-002\xspace}
\newcommand{\geminiprotwo}{Gemini-1.5-Prof-002\xspace}
\newcommand{\ft}{Direct FT\xspace}
\newcommand{\cotft}{CoT FT\xspace}

\crefname{section}{\S}{\S}
\crefname{appendix}{\S}{\S}
\crefname{table}{Tab.}{Tabs.}
\crefname{figure}{Fig.}{Figs.}
\Crefname{theorem}{Thm.}{Thms.}
\Crefname{proposition}{Prop.}{Props.}
\crefname{algorithm}{Alg.}{Algs.}
\Crefname{assumption}{Asm.}{Asms.}
\crefname{mechanism}{Mech.}{Mechs.}
\Crefname{definition}{Def.}{Defs.}
\Crefname{equation}{Eq.}{Eqs.}
\usepackage{caption}
\usepackage[font=footnotesize, labelfont=footnotesize]{caption} 

\definecolor{lemon}{RGB}{255,247,0}
\definecolor{maize}{RGB}{250,237,94}
\definecolor{mustard}{RGB}{255,219,89}
\definecolor{ocre}{RGB}{241,103,35}
\definecolor{c0}{cmyk}{1,0.3968,0,0.2588} 
\definecolor{c1}{cmyk}{0,0.6175,0.8848,0.1490} 
\definecolor{c2}{cmyk}{0.1127,0.6690,0,0.4431} 
\definecolor{c3}{cmyk}{0.3081,0,0.7209,0.3255} 
\definecolor{c4}{RGB}{164, 16, 52}
\definecolor{orange}{HTML}{E66100}
\definecolor{bluex}{HTML}{0C7BDC}
\definecolor{yellow}{HTML}{FFC20A}
\definecolor{lightpurple}{HTML}{E6E6FA}
\definecolor{lightbluee}{HTML}{e8f4f8}
\definecolor{blush}{rgb}{0.87, 0.36, 0.51}
\definecolor{c5}{HTML}{EE4E4E}
\definecolor{gggggg}{HTML}{EFEFEF}
\definecolor{Gred}{RGB}{219, 50, 54}
\definecolor{Ggreen}{RGB}{60, 186, 84}
\definecolor{Gblue}{RGB}{72, 133, 237}
\definecolor{Gyellow}{RGB}{247, 178, 16}
\definecolor{ToCgreen}{RGB}{0, 128, 0}
\definecolor{myGold}{RGB}{231,141,20}
\definecolor{myBlue}{rgb}{0.19,0.41,.65}
\definecolor{myPurple}{RGB}{175,0,124}

\providecommand{\Comments}{1}
\ifnum\Comments>0
\usepackage[colorinlistoftodos,prependcaption,textsize=scriptsize]{todonotes}
\paperwidth=\dimexpr \paperwidth + 0.8cm\relax
\marginparwidth=\dimexpr\marginparwidth + 3.8cm\relax %
\else
\usepackage{todonotes}
\fi
\newcommand{\mytodo}[1]{\ifnum\Comments=1{#1}\fi}

\usepackage{inconsolata}
\usepackage[T1]{fontenc}
\usepackage[scaled=0.95]{biolinum}

\usepackage{soul}

\newcommand{\revise}[1]{{#1}}
\usepackage{tcolorbox}
\definecolor{chart}{HTML}{1f77b4}

\newtcolorbox{example}[1][]{
  colback=chart!5!white,
  colframe=chart,
  floatplacement=floating,
  title=\centering \textsf{\small #1}
}

\newtcbox{\hlprimarytab}{on line, box align=base, colback=BlueGreen!20,colframe=blue,size=fbox,arc=3pt, before upper=\strut, top=-2.5pt, bottom=-4.5pt, left=-2pt, right=-2pt, boxrule=0pt}
\newtcbox{\hlsecondarytab}{on line, box align=base, colback=WildStrawberry!10,colframe=orange,size=fbox,arc=3pt, before upper=\strut, top=-2.5pt, bottom=-4.5pt, left=-2pt, right=-2pt, boxrule=0pt}
\newtcbox{\hlwhite}{on line, box align=base, colback=WildStrawberry!8,colframe=white,size=fbox,arc=2pt, before upper=\strut, top=-3pt, bottom=-4.5pt, left=-2pt, right=-2pt, boxrule=0pt}
\newtcbox{\hlyellow}{on line, box align=base, colback=BlueGreen!10,colframe=white,size=fbox,arc=2pt, before upper=\strut, top=-3pt, bottom=-4.5pt, left=-2pt, right=-2pt, boxrule=0pt}

\usepackage{titletoc}
\usepackage{fontawesome}

\usepackage{tcolorbox}

\newtcolorbox{promptbox}[1][]{%
  colback=blue!10,
  colframe=blue!30!black,
  fonttitle=\bfseries,
  title=#1, %
  sharp corners,
    fontlower=\scriptsize, %
  boxsep=1mm, %
}

\icmltitlerunning{On Memorization of Large Language Models in Logical Reasoning}

\begin{document}

\twocolumn[
\icmltitle{On Memorization of Large Language Models in Logical Reasoning}

\icmlsetsymbol{equal}{*}

\begin{icmlauthorlist}
\icmlauthor{Chulin Xie}{sch}
\icmlauthor{Yangsibo Huang}{comp,pri}
\icmlauthor{Chiyuan Zhang}{comp}\\
\icmlauthor{Da Yu}{comp}
\icmlauthor{Xinyun Chen}{comp}
\icmlauthor{Bill Yuchen Lin}{yyy}
\icmlauthor{Bo Li}{sch}
\icmlauthor{Badih Ghazi}{comp}
\icmlauthor{Ravi Kumar}{comp}
\end{icmlauthorlist}

\icmlaffiliation{comp}{Google}
\icmlaffiliation{sch}{University of Illinois Urbana-Champaign}
\icmlaffiliation{pri}{Princeton University}
\icmlaffiliation{yyy}{University of Washington}

\icmlkeywords{Machine Learning, ICML}

\vskip 0.3in
]

\printAffiliationsAndNotice{}  %

\begin{abstract}

Large language models (LLMs) achieve good performance on challenging reasoning benchmarks, yet could also make basic reasoning mistakes. This contrasting behavior is puzzling when it comes to understanding the mechanisms behind LLMs' reasoning capabilities. One hypothesis is that the increasingly high and nearly saturated performance on common reasoning benchmarks could be due to the memorization of similar problems. In this paper, we systematically investigate this hypothesis with a quantitative measurement of memorization in reasoning tasks, using a dynamically generated logical reasoning benchmark based on Knights and Knaves (\kk) puzzles. We find that LLMs could interpolate and memorize the training puzzles (achieving near-perfect accuracy) after fine-tuning, yet they struggle with slight variations of these puzzles. On the other hand, we show that while fine-tuning leads to heavy memorization, it also consistently improves generalization performance. Through in-depth analyses with perturbation tests, cross difficulty-level transferability, probing model internals, and fine-tuning with wrong answers, we establish that LLMs develop reasoning skills on \kk puzzles alongside memorization.
Finally, our analysis based on a per-sample memorization score sheds light on how LLMs switch between reasoning and memorization when solving logical puzzles. 
\end{abstract}

\section{Introduction}
Modern Large Language Models (LLMs) show impressive reasoning capabilities 
that allow them to solve a wide range of challenging problems including commonsense reasoning and mathematical reasoning. 
In the meantime, LLMs also make mistakes on some of the most basic problems, such as comparing which number is bigger---13.11 or 13.8~\citep{1311vs138}, and counting the number of sisters that Alice’s brother has~\citep{nezhurina2024alice}.
This contrast is puzzling when it comes to understanding how exactly LLMs solve reasoning tasks. This question is important both scientifically and practically: understanding how LLMs reason could shed light on their learning and generalization behaviors.  It is also crucial for real-world applications where robust reasoning is required due to safety
and trustworthiness concerns~\citep{wang2023decodingtrust, wallace2024instruction, lee2024mechanistic, wei2024assessing}.

One hypothesis is that LLMs could be relying on \emph{memorization} when solving those reasoning tasks, especially when measured by popular benchmarks that could be accidentally leaked into various massive internet-crawled pre-training datasets. Previous work~\citep{tirumala2022memorization, carlini22quantifying} show that LLMs could indeed memorize the training data, which may lead to potential privacy~\citep{carlini2021extracting} or copyright~\citep{karamolegkou2023copyright,wei2024evaluating} concerns. Additional evidences of potential memorization come from extensive studies on data contamination in LLMs~\citep{magar2022data, balloccu2024leak, shidetecting, xu2024benchmarking, oren2023proving}. To mitigate the issue of benchmark saturation potentially due to memorization, some papers focus on designing dynamic benchmarks~\citep{roberts2023cutoff,zhu2023dyval,srivastava2024functional,jain2024livecodebench, wu2024conceptmix} or alternative evaluation protocols~\citep{zeng2024mrgsm8kmetareasoningbenchmarklarge, zhang2024careful, xu2024benchmarking, srivastava2024functional}. %

\begin{figure*}[t]
    \centering
    \includegraphics[width=.95\linewidth]{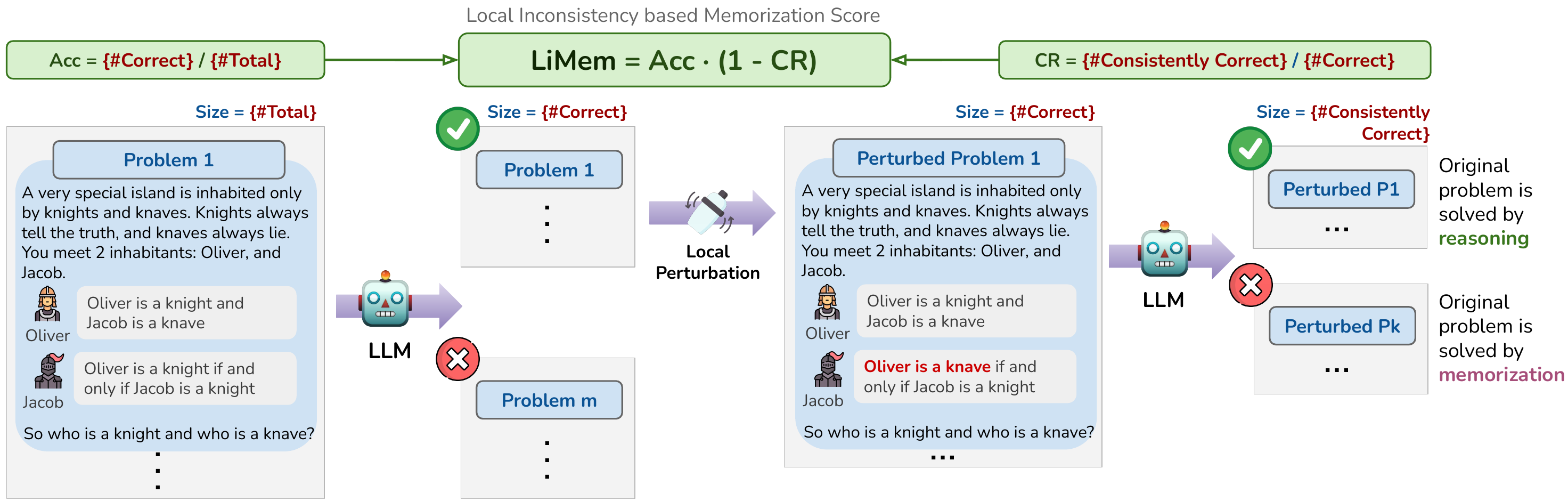}
    \vspace{-2mm}
    \caption{
    \small Illustration of the definition of Local Inconsistency based Memorization Score, $\limem$. \revise{High level of memorization occurs when the model shows high accuracy in solving some problems but fails to consistently solve those problems under local perturbations that require similar underlying reasoning principles.}
    }
    \vspace{-4mm}
    \label{fig:mem_reason_explain}
\end{figure*}

In this paper, we take a direct approach to quantify the memorization behaviors of LLMs in reasoning tasks within a controlled setting. Specifically, we seek to understand: (i) whether LLMs rely on memorization to solve reasoning tasks, and (ii) whether memorization is only detrimental to learning to reason. Both questions are inspired by human behavior. For instance, when a student works hard on the preparation material for an exam, the preparation could help them get familiarized with the problems, and their ability to solve new problems could usually improve with enough exercises. However, without genuinely understanding the principles, they might fail when the same problem is slightly changed despite doing well on prepared problems. 
Our metric of memorization \revise{$\limem$}, illustrated in \Cref{fig:mem_reason_explain}, is based on this intuition: \revise{an LLM shows a high level of memorization when it solves reasoning problems with high accuracy but struggles to consistently solve those problems under local perturbations requiring similar mathematical principles (i.e., low consistency).}
We note that a similar perturbation (mostly at language-level) idea has been used in previous work, especially in detecting contamination~\citep{golchin2023data, yang2023rethinking,xu2024benchmarking}.
However, given our focus on understanding memorization in logical reasoning tasks, we further consider problem-level perturbation that slightly changes the mathematical structure of a puzzle, in addition to language-level perturbations.
To facilitate our study, we propose a new logical reasoning benchmark that supports automatic problem-level perturbation.
With this tool, we evaluate the reasoning power of \revise{17} off-the-shelf LLMs. We then fine-tune \llamathree and \gptfouromini to quantify their memorization in reasoning tasks, and reveal interesting interplay: \revise{while models indeed tend to memorize many training logical puzzles, they also develop reasoning capabilities during fine-tuning} (even directly on question-answer pairs without reasoning steps), and the reasoning performance improves when memorizing more training puzzles.

In the following, we 
summarize our key contributions:
\begin{itemize}[leftmargin=*, itemsep=0pt, topsep=0pt, partopsep=0pt, parsep=0.5pt]
    \item To quantify memorization in reasoning tasks, we define a memorization score based on the notions of performance inconsistency under local perturbation, inspired by human behavior (\Cref{sec:limem-def}).
    \item To facilitate the measurement, we propose  a new logical reasoning benchmark based on the \emph{Knights and Knaves}~\citep[\kk,][]{KKorig, johnson1990meta} puzzles, that can generate new puzzles at different difficulty levels, locally perturb existing puzzles, and automatically synthesize detailed reasoning steps to solve a given puzzle (\Cref{sec:proposal_dataset}). %
    \item We show that \kk puzzles are challenging, and only the most advanced LLMs could solve them well. 
    Moreover, our analysis suggests those models exhibit some level of memorization (\Cref{sec:measure_mem}). 
    \item By fine-tuning on \kk samples, we confirm that modern LLMs are capable of memorizing a large collection of puzzles, and reach high memorization score when interpolating~\citep[i.e., fitting,][]{belkin2018overfitting} the training set. We observe that the models' generalization accuracies continue to improve as memorization grows (\Cref{sec:learn-to-reason-answer}).
    \item We design various in-depth analyses (\Cref{subsec:direct-ft-generalization}$\sim$\Cref{subsec:wrong-ans-ft}) to verify that \revise{LLMs developed improved reasoning capabilities (i.e., generalization) 
    after fine-tuning even with only question-answer pairs,   via local perturbation tests, cross difficulty-level transferability,  fine-tuning with wrong answers,
    and model internal probing.}
    \item We show that fine-tuning with detailed reasoning steps can further boost the generalization on \kk puzzles, even when fine-tuned with wrong reasoning steps (\Cref{sec:learn-to-reason-cot}).
    \item To analyze the interplay between memorization and reasoning, we measure per-sample memorization and study how LLMs switch between memorization and reasoning  to solve a puzzle (\Cref{sec:mem-reason-classification}).
\end{itemize}

\section{Measuring Memorization in Reasoning}
\label{sec:proposal}
\subsection{Memorization Metrics for Reasoning Tasks}
\label{sec:limem-def}

\begin{figure*}[t]
    \centering

    \begin{minipage}[t]{0.63\linewidth}
       \centering
    \includegraphics[width=1\linewidth]{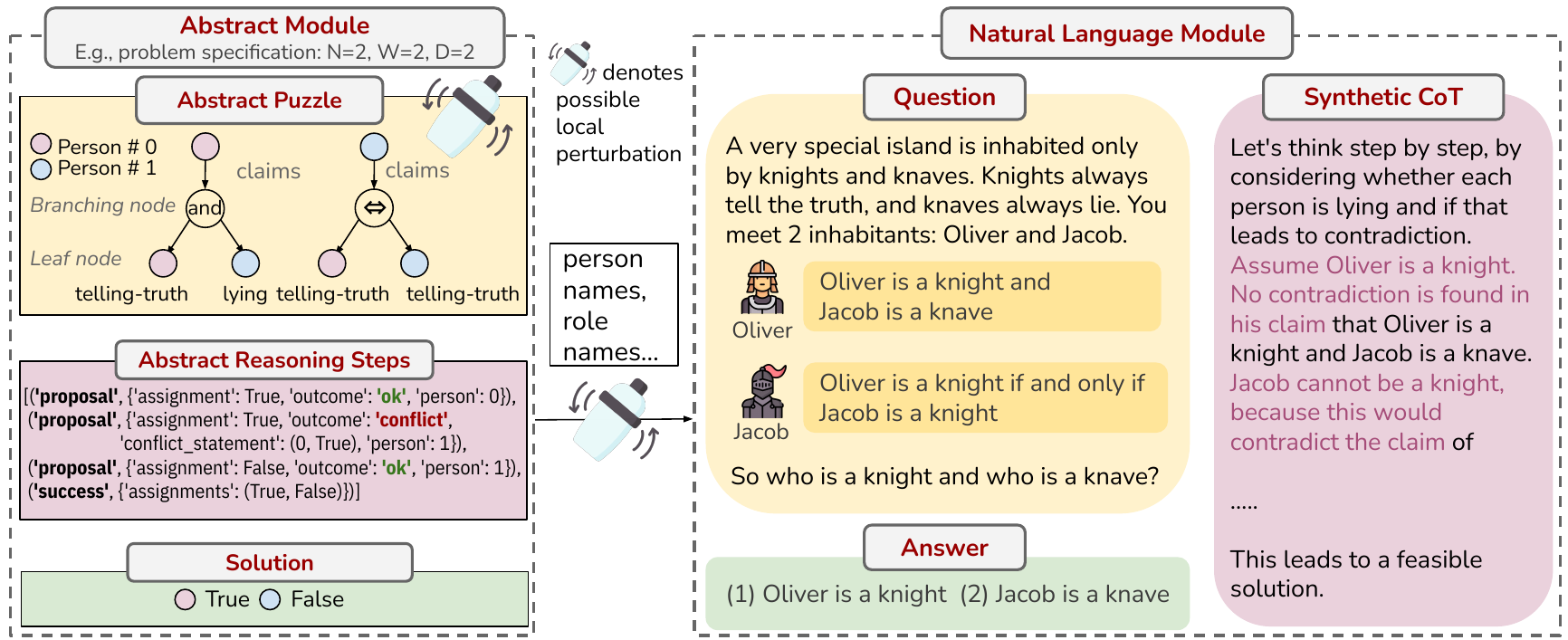}
    \vspace{-6mm}
    \caption{
    \small  \kk data generation framework employs abstract module and natural language module to generate question-answer pair and synthetic CoTs for each \kk puzzle, based on the problem specification: number of persons ($N$), tree width ($W$), and depth ($D$). 
    Perturbers in these modules can alter the math structure and language description, respectively, and recompute the question-answer pair.}
    \vspace{-3mm}
    \label{fig:data_gen}
    \end{minipage}
    \hfill
     \begin{minipage}[t]{0.35\linewidth}
    \centering
   \includegraphics[width=1\linewidth]{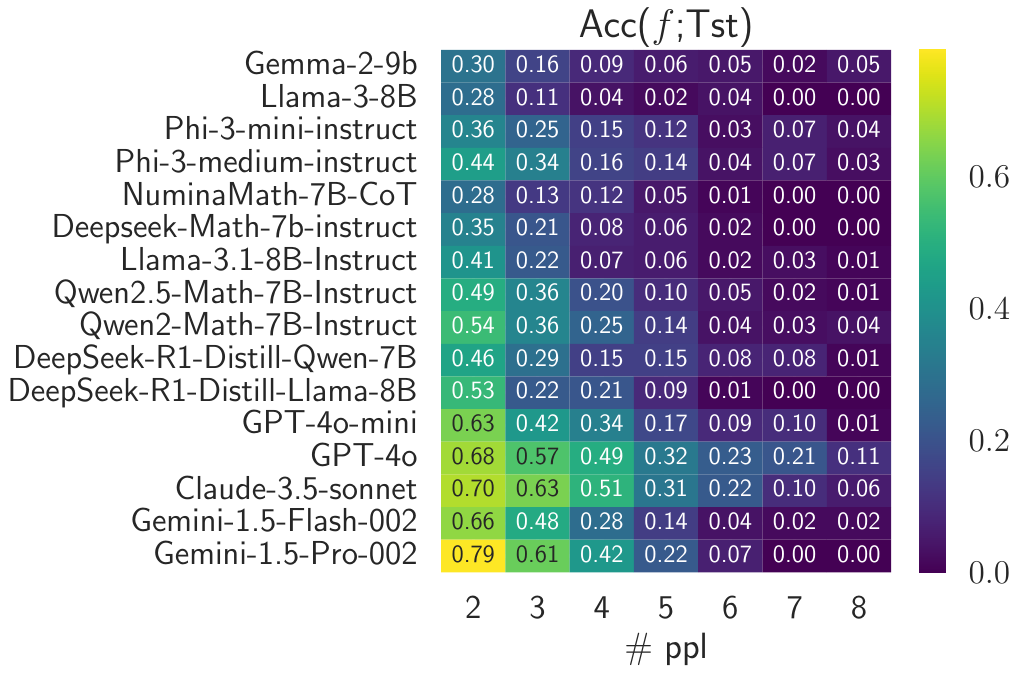}
    \vspace{-6mm}
    \caption{
    \small Test acc of off-the-shelf LLMs under 0-shot direct prompting drops with increasing puzzle complexity. \revise{For reference, OpenAI o1 model with test-time compute achieves an acc of 0.86  on 8-ppl task and 0.67 on 18-ppl task.
    }
    }
    \label{fig:eval_offshelf_lp_gap_test}
    \end{minipage}
    \vspace{-4mm}
\end{figure*}

Memorization of LLMs has been studied in various contexts such as privacy~\citep{carlini22quantifying}, copyright ~\citep{carlini2021extracting,karamolegkou2023copyright,wei2024evaluating, he2024fantastic}, and solving knowledge intensive tasks~\citep{hartmann2023sok}. In this paper, we are specifically interested in measuring the level of memorization when solving reasoning tasks, by borrowing intuition from human behavior.
For example, when preparing for an exam, a student may not be able to fully digest the underlying principles due to various reasons or constraints. But when (luckily) facing the same problem the student had prepared for, they would still be able to solve it.  A key characteristic of this type of memorization is: (A) high accuracy on observed problems and (B) low accuracy when the problem is slightly changed.
Based on this intuition, for a dataset $\mathcal{D}$ of reasoning puzzles, we combine the following two quantities to measure memorization: %
\begin{enumerate}[leftmargin=*, itemsep=0pt, topsep=0pt, partopsep=0pt, parsep=1pt]
    \item 
    For (A), we measure the accuracy of a target model $f$ on $\mathcal{D}$, denoted as $\accd$. We are especially interested in measuring on the set of \emph{observed puzzles}, i.e., the training set, $\tracc$. We say $f$ \textit{interpolates}~\citep{belkin2018overfitting,muthukumar2020harmless,belkin2021fit,bartlett2021deep} the training puzzles if $\tracc\approx 100\%$.
    \item 
    For (B), we measure a \emph{consistency ratio}  $\consistr(f;\mathcal{D})$ between the number of \emph{consistently solved puzzles}  after some \emph{local perturbations},
    and the number of solved puzzles (without perturbation). We are interested in local perturbations that make minimal changes to the puzzle and maintain the same underlying principle for solving it, and a similar difficulty level (to be specified in \cref{sec:proposal_dataset}).  %
\end{enumerate}
We combine the two factors to define a \textit{Local Inconsistency-based Memorization Score} $\limemd\in[0,1]$: 
\begin{align}
    \limemd &= \accd \cdot (1 - \consistr(f;\mathcal{D}))  \label{eq:limem} \\
            &= \frac{\texttt{\#Correct}- \texttt{\#Consistently\_Correct}}{\texttt{\#Total}}.  \notag
\end{align}
When there is no ambiguity, we call it the memorization score. A larger score provides stronger evidence of memorization (\revise{i.e., a larger proportion of memorized examples in the given dataset}). 
Specifically, a high $\limemtr$ matches the characteristic behavior of human memorizing observed puzzles, and in this case we say $f$ \textit{memorized} the training puzzles. 
Note that the $\accd$ factor is necessary, as there can be three types of behaviors: (i) solving by memorization, (ii) solving by reasoning, (iii) not solving (e.g., random guessing). A high $\limemd$ indicates (i), but a low $\limemd$ would only indicate (ii) if we separately check that $\accd$ is high.

To effectively measure the memorization score $\limemd$, we need a principled way to (1) locally perturb the puzzle while maintaining its difficulty level; (2) compute the new correct answer after perturbation. Towards this goal, we design and implement a functional dataset based on the Knights and Knaves puzzles~\citep{KKorig, johnson1990meta}. 

\begin{figure*}[t]
    \centering
    \begin{minipage}[t]{0.37\linewidth}
        \centering
         \vspace{-15mm} %
        \includegraphics[width=0.45\linewidth]{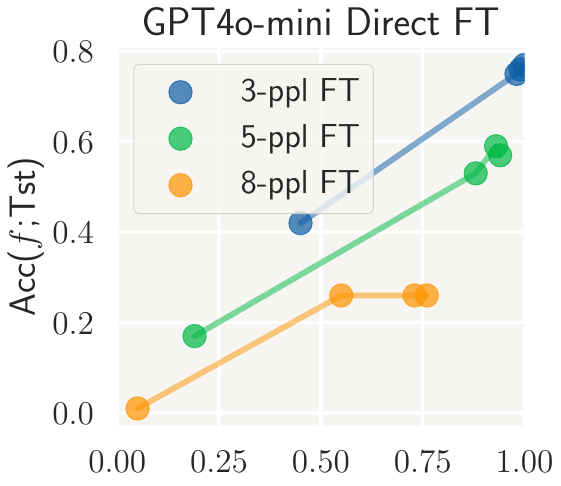}
        \includegraphics[width=0.45\linewidth]{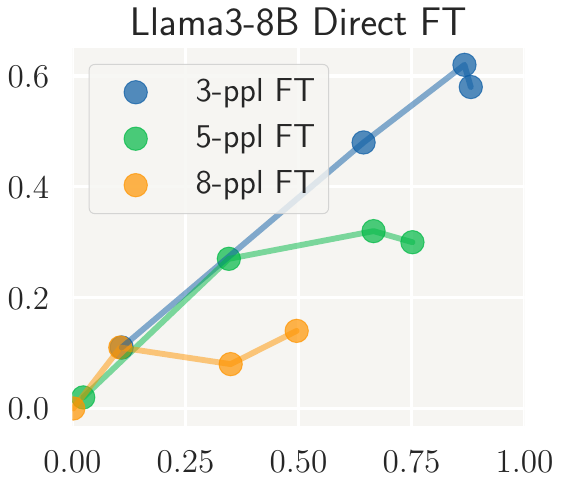}
        \includegraphics[width=0.45\linewidth]{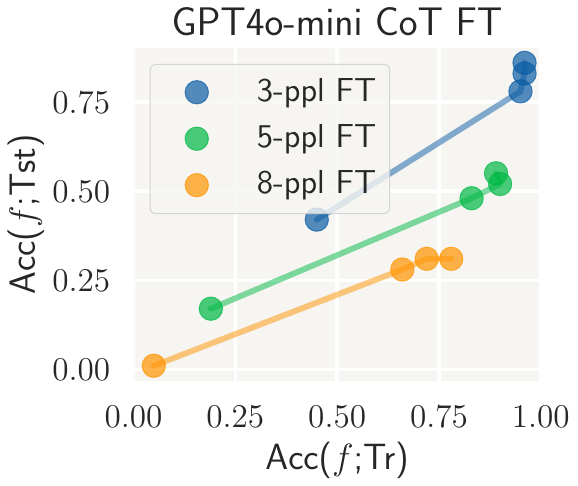}
        \includegraphics[width=0.45\linewidth]{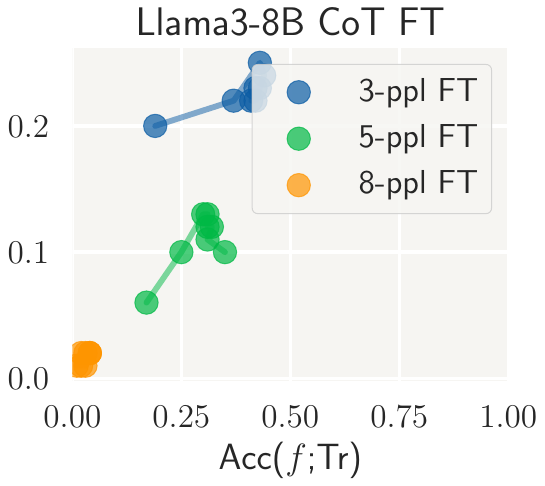}
        \vspace{-2mm}
        \caption{\small Train \& test accuracy increases over the epochs.  FTed LLMs can achieve interpolation ($\approx100\%$ train accuracy) for easy tasks, e.g., 3/5-ppl puzzles. \llamathree struggles with CoT FT on \kk tasks, likely due to limited model capacity.
        }
        \label{fig:xtrain_ytest_acc_ft}
    \end{minipage}
    \hfill
     \begin{minipage}[t]{0.60\linewidth}
        \centering
         \includegraphics[width=1\linewidth]{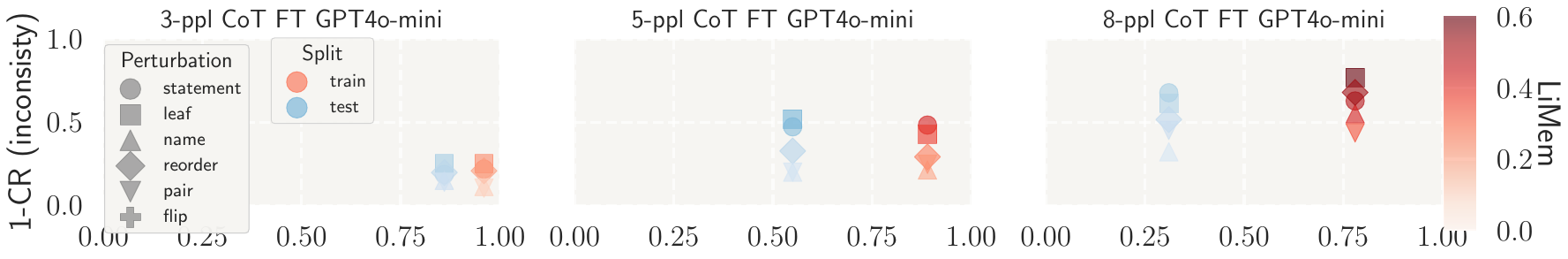}
        \includegraphics[width=1\linewidth]{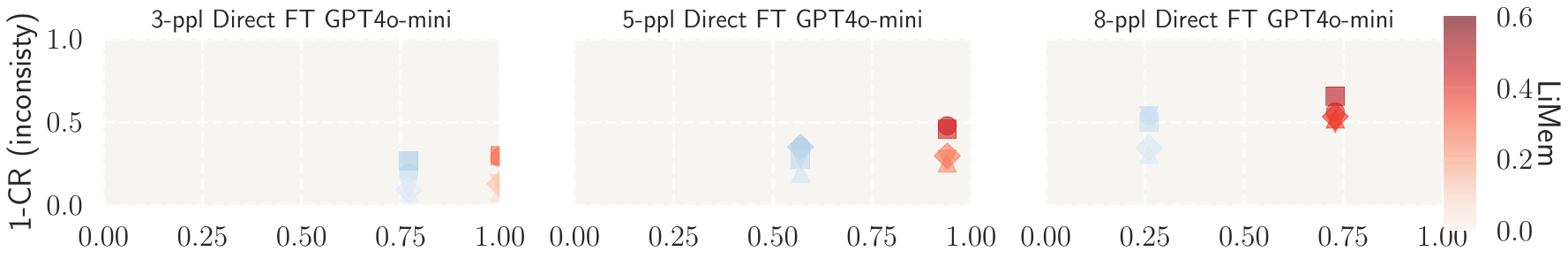}
        \includegraphics[width=1\linewidth]{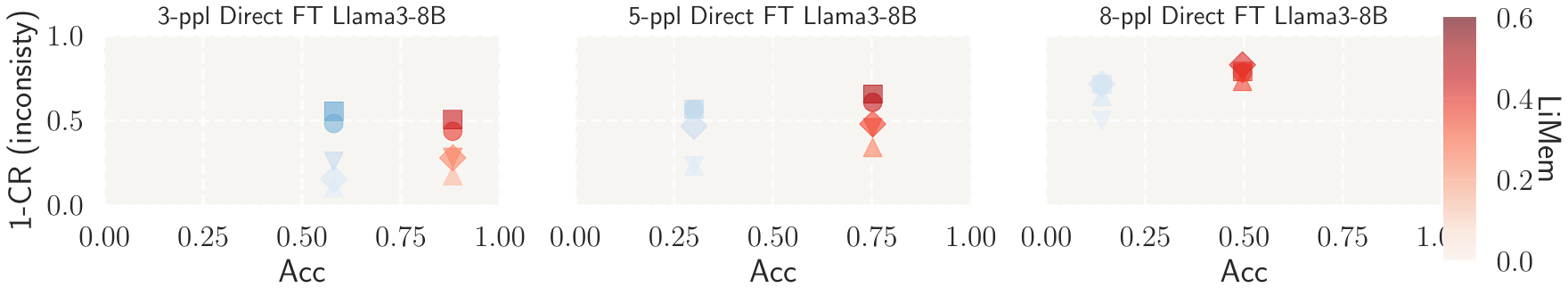}
        \vspace{-8mm}
        \caption{\small \revise{Fine-tuned LLMs generally exhibit 
       both higher clean accuracy (x-axis) \& inconsistency ratio under perturbations (y-axis) on the train set than test set, resulting in a higher memorization score (color spectrum). 
        LLMs show stronger memorization under math-level perturbations (statement/leaf) than language level.
        We separately report memorization score in \cref{fig:perturb_ablation_direct_ft_seperate_mem_score} and consistency ratio in \cref{fig:perturb_ablation_direct_ft_consist_ratio}, and results under combined math \& language-level perturbations in \cref{fig:perturb_ablation_direct_ft_stack_mem_score}.
        }
        }
        \label{fig:perturb_ablation_direct_ft_mem_score}
    \end{minipage}
    \vspace{-5mm}
\end{figure*}

\subsection{Knights and Knaves Reasoning Benchmark}
\label{sec:proposal_dataset}

\emph{Knights and Knaves (\kk)} is a type of logical puzzle, \revise{where the goal is to infer each character $i$’s truthfulness $B_i$ (Boolean value) by judging the logical consistency of the statements $S_i$ they made.} 
\cref{fig:data_gen} shows an example.

\revise{The principle underlying \kk is the Boolean satisfiability problem (SAT)~\citep{wiki:booleansat}. SAT was the first problem proven to be NP-complete and many well-known problems can be translated into SAT, such as hardware and software verification and theorem proving~\citep{wiki:satsolver}. Hence, the performance of a model on SAT (i.e., \kk puzzles) can be important indicative of its reasoning capabilities. 
Specifically, consider a \kk puzzle involving $N$ people, and a possible Boolean value assignments to $\{B_i\}_{i=1}^N$, where $B_i$ indicates whether the $i$th person is telling the truth,
i.e., their statement $S_i$ is true. Therefore, a valid solution to a \kk puzzle is an assignment such that the following formula is true: $(B_1\Leftrightarrow S_1)\wedge(B_2\Leftrightarrow S_2)\wedge\cdots\wedge(B_N\Leftrightarrow S_N)$.}

Based on the \kk puzzle, we design a \emph{dynamic} benchmark that supports \revise{generating new puzzles and local perturbations. 
Our benchmark has 2 modules (See \Cref{fig:data_gen} for an overview):} 

\revise{\textbf{The Abstract Module} contains 4 components that generate and manipulate \kk puzzles in an abstract form: The \emph{Generator} that produce random puzzles; the \emph{Solver} that find valid solutions algorithmically; the \emph{Reasoner} that generate human-like reasoning steps (chain-of-thoughts, CoT); and the \emph{Perturber} that maps a given puzzle to a local perturbation.
Each puzzle involves $N$ people, each making a statement forming a logical tree with max width/depth of $W$/$D$, using the logical operations \emph{and}, \emph{or}, \emph{not}, \emph{implication}, and \emph{equivalence}. The \emph{Perturber} is the most important part that is not usually supported in existing benchmark. It generates perturbation by replacing a statement or a leaf node in a statement with a newly sampled one, and ensures the perturbation has a different solution. This support is crucial for making our memorization measurements.}

\revise{\textbf{The Natural Language Module} converts the puzzles and the generated CoTs into natural language. It uses random names and templates to diversify the generated puzzles, and also supports \emph{language level perturbation} to a given puzzle.}

\revise{See \Cref{app:dataset_detail} for more details. We generate and release a core dataset of $1000$/$100$/$50$ train/test/validation puzzles for each $2\leq N\leq 8$ people\footnote{We only include $200$ training examples for 2-ppl puzzles due to the limited problem space. Note the problem space is huge as $N$ increases. For example, for 8-ppl puzzles ($D,W=2,2$), there are \textasciitilde$10^{24}$ unique problems, and \textasciitilde$30\%$ of them has a unique solution based on empirical estimation from 100k random generations.}.
By default we use a maximum tree width of $W=2$ and depth $D=2$.
For each puzzle, we also generate six perturbed variants (2 problem level and 4 language level): \{\textit{perturbed statement, perturbed leaf node, random role-pair name, uncommon person name, reordered statement, flipped role}\}. 
We release the code for generation additional puzzles and perturbations if needed.}

\textbf{\kk is challenging for off-the-shelf models.} 
We use 0-shot direct prompting with task-specific instructions for open-ended question-answering (details in \Cref{app:exp_details}).\footnote{Note that even under direct prompting, capable LLMs can generate Chain of Thought~\citep[][CoT]{wei2022chain}. Our evaluation mainly considers the $0$-shot setting to avoid biases from in-context examples~\citep{Zhao2021CalibrateBU}, but we provide results for CoT prompting, 1-shot prompting \& self-consistency prompting in \cref{app:add_exp_results}.} Accuracy is determined by keyword matching and requires correctly identifying \emph{all} characters in the conclusion. We evaluate 17 leading models known for strong reasoning performance.
\cref{fig:eval_offshelf_lp_gap_test} shows that \kk puzzles are highly challenging---even for the simplest 2-ppl puzzles, the best models \revise{(except o1 model)} achieve at most 70\% accuracy, which drops to just 11\% for 8-ppl puzzles. 
\revise{In \cref{appsubsec:eval_benchmark}, we show that various prompting techniques like CoT/1-shot/self-consistency~\cite{wangself2023} cannot fundamentally improve performance on challenging \kk tasks.}

\begin{figure*}[t]
    \centering
    \begin{minipage}{0.24\textwidth}
        \centering
        \includegraphics[width=0.82\linewidth]{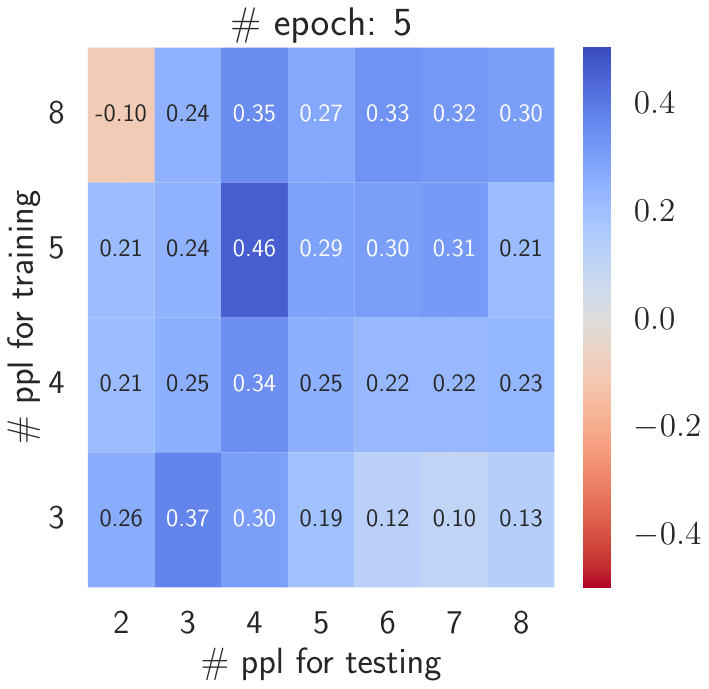}
        \subfigure{\small (a) \gptfouromini  \cotft}
    \end{minipage}
    \begin{minipage}{0.24\textwidth}
        \centering
        \includegraphics[width=0.82\linewidth]{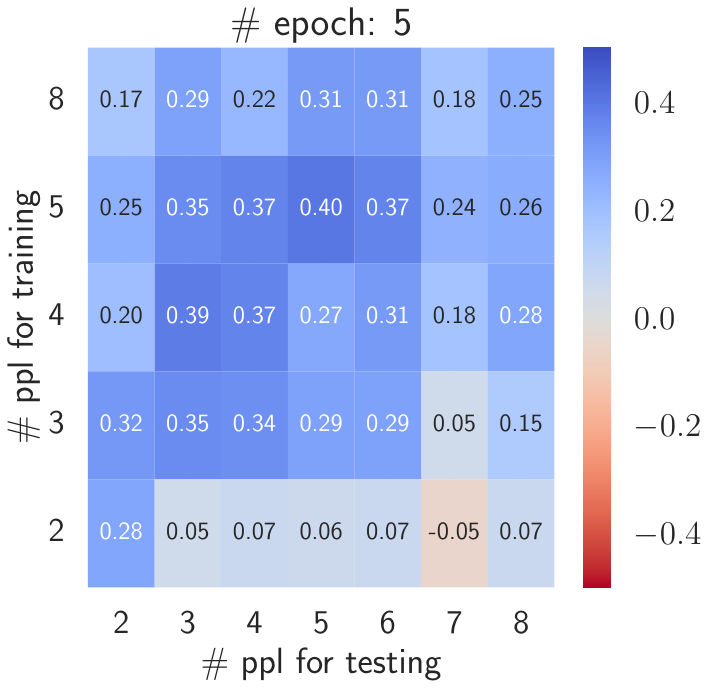}
        \subfigure{\small (b) \gptfouromini  \ft}
    \end{minipage}
    \hfill
    \begin{minipage}{0.5\textwidth}
        \centering
        \includegraphics[width=0.82\linewidth]{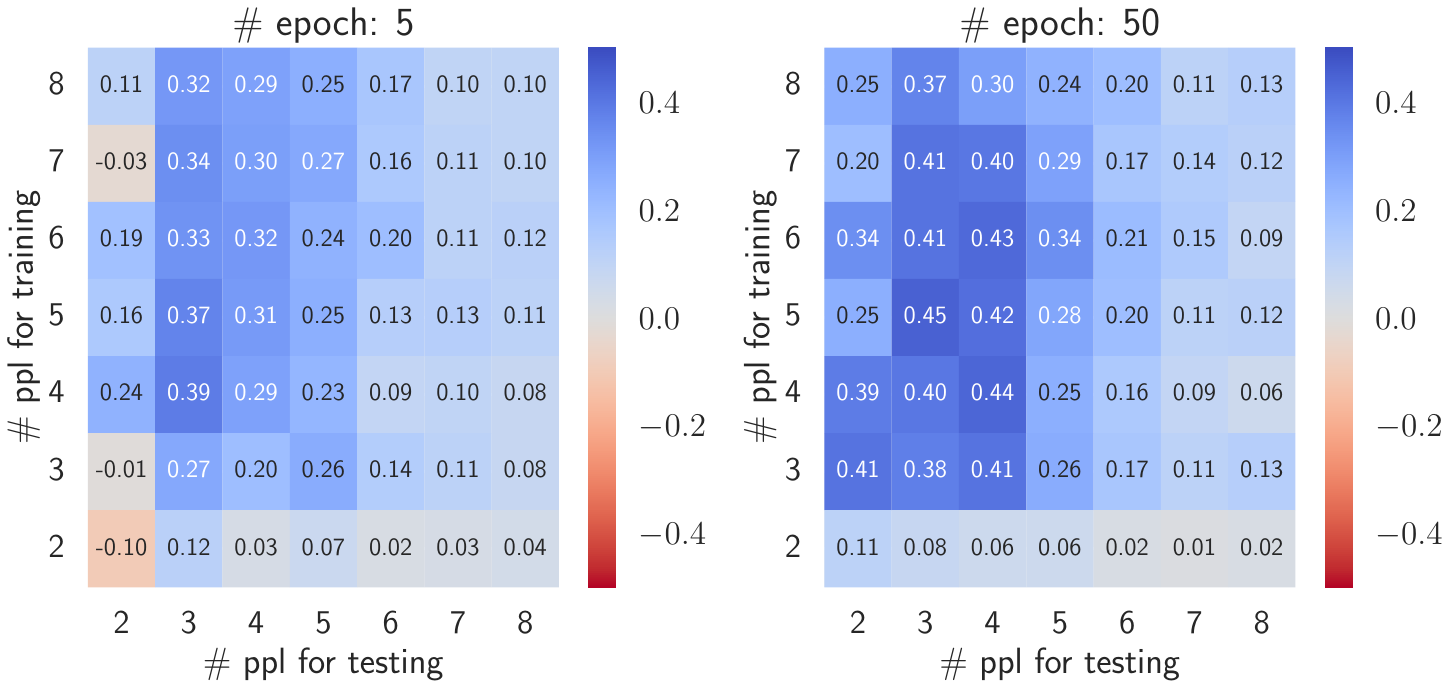}
        \subfigure{\small (c) \llamathree \ft}
    \end{minipage}
    \hfill
    \vspace{-3mm}
    \caption{Test accuracy improvement on $N$-people problems for LLMs fine-tuned on $M$-people problems, compared to the unfine-tuned model, under 0-shot direct prompting. Most grid values are above 0, indicating transferability and enhanced reasoning abilities across unseen tasks. Results for more epochs are in \cref{appsubsec:eval_reason}.
    }
    \label{fig:transfer_grid_direct_ft}
    \vspace{-5mm}
\end{figure*}

\vspace{-2mm}
\section{Quantifying Memorization in Reasoning}
\label{sec:measure_mem}

Here, we study a model's memorization behavior when fine-tuned on \kk puzzles.

\textbf{Fine-tuning setup.}
We fine-tune the models for each $N$-people task separately, with $n_{\text{train}}=1,000$ for $3\leq N \leq 8$, and $n_{\text{train}}=200$ for $2$-people task due to limited number of combinations.
We take \llamathree and \gptfouromini and run \emph{supervised fine-tuning} (SFT) on a set of \kk training puzzles disjoint from the test set. 
We consider two fine-tuning paradigms:
(1) Fine-tuning on detailed CoT steps (\textbf{\cotft}): during SFT, the model observes the concatenation of the question, synthetic CoT steps, and the answer for each puzzle; the loss is computed on the CoT steps and the answer part.  
(2) Fine-tuning on the answers (\textbf{\ft}) where the model observes the question-answer pair for each puzzle, and the loss is only computed on the answer part. 
Examples of \cotft/\ft training instances are provided in \cref{app:ft_details}.
We fine-tune \llamathree for \revise{$50$} epochs\footnote{\revise{We fine-tune \llamathree for max 100 epochs in \cref{fig:train-acc} and find that it typically converges at 50 epochs.}} and \gptfouromini for $5$ epochs (due to budget constraints) via the OpenAI fine-tune API (details in \cref{app:exp_details}). 
During the evaluation, we follow the same prompting paradigm as FT paradigm, i.e., direct/CoT prompting for direct/CoT-FTed model, which is shown effective in \cref{appsubsec:eval_reason}.

\textbf{LLMs interpolate \kk training puzzles}. 
In \cref{fig:xtrain_ytest_acc_ft}, we present the training accuracy of models trained on each task on the $x$-axis (each dot represents a training epoch). We find that models exhibit high training accuracy in tasks such as $3/5$-people puzzles. The higher capacity model \gptfouromini nearly achieves interpolation ($\tracc\approx100\%$) using both \ft and \cotft.

\textbf{Interpolating LLMs have large memorization scores on training examples}. From \cref{fig:perturb_ablation_direct_ft_mem_score},  
\textbf{(1)} we observe high $\limemtr$ memorization score on training samples (e.g., $\sim 50\%$ on 8-people task)  under various perturbations. It shows significant gaps between accuracy on the original sample and the consistent accuracy under perturbation, suggesting a heavy reliance on memorization.
\textbf{(2)} $\limemtr$  is higher for more difficult tasks (e.g., 5/8-people), which could mirror human behavior, where memorization is often used to tackle challenging tasks that people do not fully understand.   
\textbf{(3)} More capable model \gptfouromini, in general, show lower memorization scores than \llamathree.

\begin{figure*}[t]
    \vspace{-2mm}
    \centering
     \begin{minipage}[t]{\linewidth}
        \centering
        \includegraphics[width=0.95\linewidth]{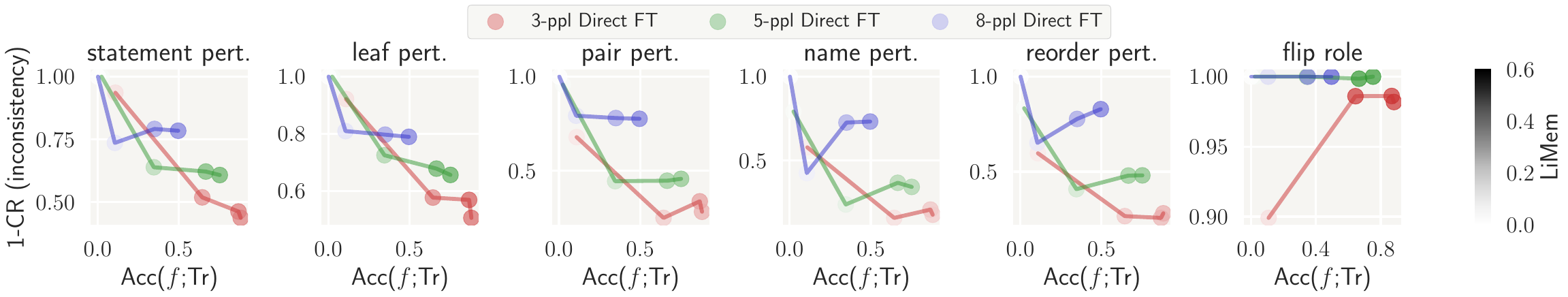}
        \vspace{-2mm}
        \caption{\revise{Inconsistency ratio (y-axis) on correctly solved training puzzles of fine-tuned \llamathree decreases over epochs (x-axis), even as the proportion of memorized training puzzles increases, as indicated by the larger $\limemtr$ values (color bar).
        }
        }
        \label{fig:test_acc_mem_llama3_direct_ft}
    \end{minipage}
    \vspace{-4mm}
\end{figure*}

\textbf{Ablation on local perturbations.}
Comparing different perturbations in \cref{fig:perturb_ablation_direct_ft_mem_score},  we find that \textbf{(1)} LLMs exhibit a higher memorization score when evaluated with math-level perturbations (e.g., statement/leaf) compared to language-level, which indicates that LLMs 
can compose the language understanding capability to solve the same puzzle in alternative phrasing.
\textbf{(2)} LLMs get nearly zero accuracy on role-flipped samples (e.g., when a knight, typically viewed as truthful, is defined as always lying), and memorization score $\limemtr$ under role-flipping for \llamathree is $\sim 80\%$ as shown in \cref{fig:test_acc_mem_llama3_direct_ft}. This could be due to an internal bias or commonsense understanding that knights are inherently good characters (e.g., truthful), and thus LLMs disregard the altered puzzle statement.

\vspace{-3mm}
\section{Learn to Reason by Fine-tuning With Answers Only}
\label{sec:learn-to-reason-answer}

\begin{figure*}[ht]
    \centering
    \includegraphics[width=\linewidth]{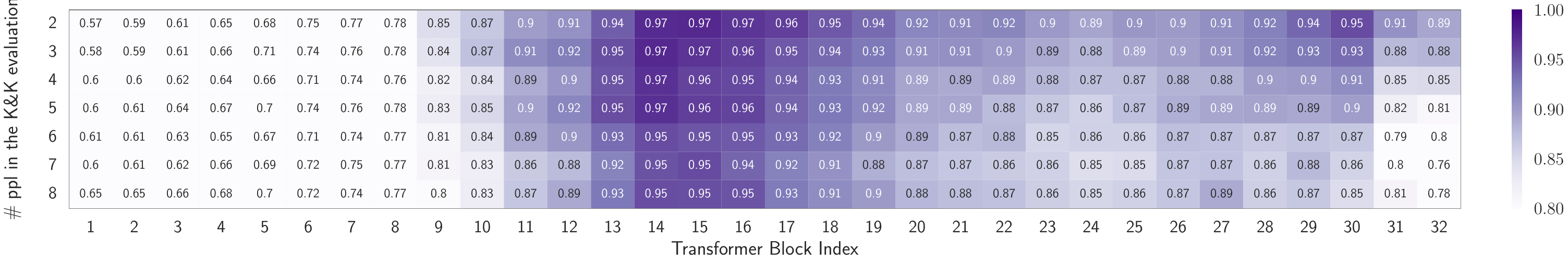}
    \vspace{-5mm}
    \caption{Probing accuracy of \kk puzzles with different number of people in testing puzzles across different layers of the \llamathree transformer model. Results for un-FTed models are shown in \Cref{fig:probe-base}  in \Cref{app:add_exp_results}.
    }
    \vspace{-4mm}
    \label{fig:probe}
\end{figure*}
\cref{sec:measure_mem} shows that fine-tuned models exhibit memorization when solving \kk reasoning tasks. Does it mean that those models do not have reasoning capabilities at all? In this section, we show that LLMs can do both, and the reasoning capability consistently improves as the memorization level increases when the models are fine-tuned on \kk puzzles.

We focus on analyzing {\ft} in this section and discuss \cotft in \Cref{sec:learn-to-reason-cot}.
For humans, solving \kk tasks without understanding the underlying logic is difficult. %
However, after observing the step-by-step reasoning steps, people can understand the procedure and solve the puzzles more easily. Similarly, compared to \cotft, learning from only answers ({\ft}) without detailed reasoning steps is intuitively more challenging for LLMs, as the models need to come up with the reasoning procedures on their own. Therefore, the models might be more likely to rely on memorization in this case. 
Surprisingly, from \Cref{fig:perturb_ablation_direct_ft_mem_score}, we did not observe {\ft}ed \gptfouromini models exhibiting consistently higher memorization score than {\cotft}ed ones. It turns out that models can learn to reason \kk puzzles well directly from observing only question-answer pairs, as we will show in \cref{subsec:direct-ft-generalization}. To better understand what the model learns through \ft, we conduct a probing analysis on model internals in \cref{subsec:wrong-ans-ft}  and an ablation study with incorrect answers fine-tuning in \cref{subsec:wrong-ans-ft}.

\vspace{-2mm}
\subsection{Reasoning capabilities of {\ft}-ed model}
\label{subsec:direct-ft-generalization}

\textbf{Fine-tuned model generalizes across different difficulty levels}. We evaluate LLMs' transferability by fine-tuning on $M$-people puzzles and testing on $N$-people puzzles. When $M \neq N$, the testing is out-of-distribution compared to  training and solving it requires reasoning.
The $N\times M$ test accuracy improvement grid (compared to the un-FTed model) in \cref{fig:transfer_grid_direct_ft} shows: 
\textbf{(1)} Training on any $M$-people puzzle generally improves test accuracy on any $N$-people puzzles, suggesting that the model learns general task-solving rules after FT (to reason and solve both easier and harder unseen puzzles). 
\textbf{(2)} More training epochs (e.g., 50 vs. 5) improve generalization, especially for \llamathree.
\textbf{(3)} Accuracy gains are larger for $N \leq 6$ puzzles, though improvements on harder tasks remain possible.

\revise{\textbf{Inconsistency ratio decreases despite increased memorization}. As shown in \cref{fig:test_acc_mem_llama3_direct_ft}, the inconsistency ratio ($y$-axis) of fine-tuned LLMs on correctly solved training puzzles decreases over epochs, even as the memorization score $\limemtr$ increases, indicating a higher proportion of memorized training puzzles (\cref{eq:limem}). This reduction in inconsistency suggests a potential improvement in the model’s generalization ability, aligning with its enhanced transferability observed in \cref{fig:transfer_grid_direct_ft}. The memorization score $\limemtr$ under role-flipping is significantly higher than other perturbation, possibly due to an internal bias that knights are truthful.
See \cref{fig:test_acc_mem_4omini} for results on \gptfouromini.
}

\begin{wrapfigure}{l}{0.5\linewidth}
    \centering
    \vspace{-5mm}
    \includegraphics[width=\linewidth]{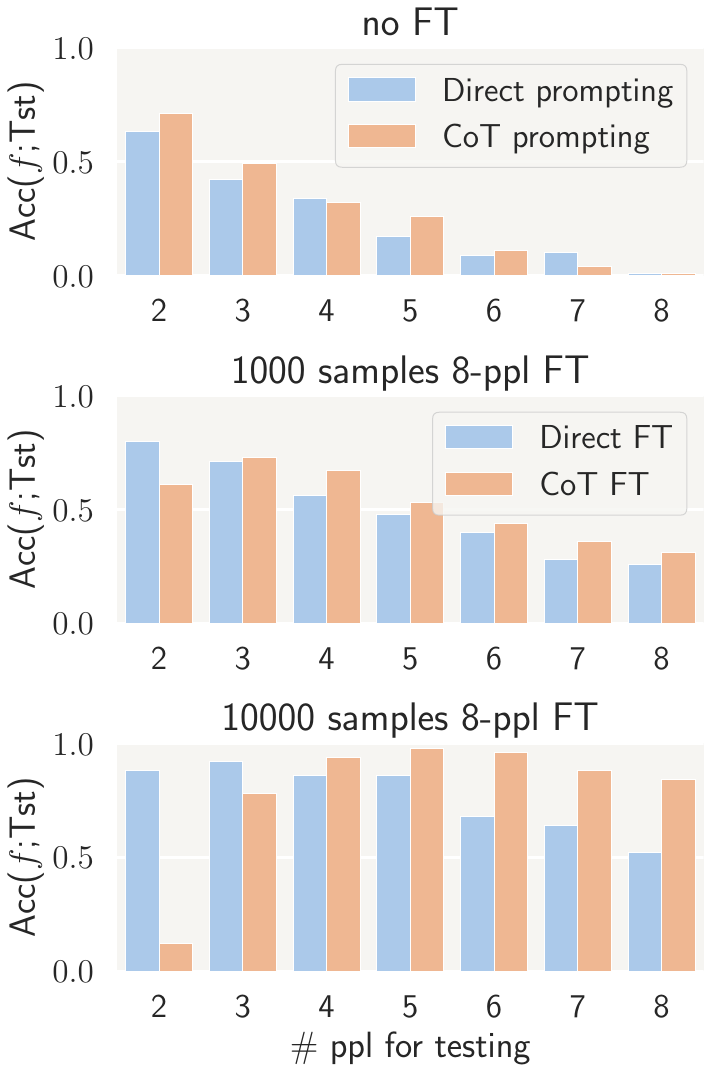}
    \vspace{-6mm}
    \caption{Transferability of 1$k$/10$k$ 8-ppl FTed \gptfouromini. \llamathree results are in \cref{fig:llama-10k-8ppl}.
    }
    \vspace{-3mm}
    \label{fig:8ppl-10k}
\end{wrapfigure}
\textbf{Fine-tuning with 10k 8-people puzzles.}
Given the significant performance improvement from fine-tuning, a natural question arises: can brute-force fine-tuning on a very large number of puzzles eventually solve the \kk puzzles, by observing/memorizing a variety of combinations of persons' claims and their corresponding answers? We \ft \gptfouromini on $1k$/$10k$ of the most challenging 8-people puzzles for 5 epochs. \cref{fig:8ppl-10k} 
shows that \textbf{(1)} $10k$-FT significantly outperforms $1k$-FT across all tasks, reaching $\sim90\%$ test accuracy on moderately difficult 4/5-people puzzles. \textbf{(2)} \cotft is generally more effective than \ft with $10k$ samples, likely due to the guidance provided by reasoning steps. \textbf{(3)} An exception is the 2-people task, where the training and testing distribution gap causes the {\cotft}ed model to occasionally get stuck in a loop of listing assumptions and contradictions, resulting in long, repetitive responses without reaching a conclusion\footnote{We observe similar accuracy drop on 2-people task for \llamathree  (see \cref{fig:llama-10k-8ppl}) when it is {\ft}ed for overly long epochs.
We provide more examples and discussions in \cref{appsubsec:4omini_eval_reason}.}. 
\textbf{(4)} \ft with $10k$ puzzles achieves surprisingly high test accuracy on all tasks, e.g., 52\% on 8-people tasks, where the un-FTed model scores near 0. Notably, the models do not see reasoning steps during training and rely solely on memorizing answers. %
We also observe high transferability for $10k$ {\ft}ed \llamathree in \cref{fig:llama-10k-8ppl}, e.g., $87\%$ test accuracy on 3-people puzzles.

\subsection{Probing {\ft}ed models}
\label{subsec:probing}

To investigate whether {\ft}ed models develop internal understanding of the skills 
necessary to solve \kk puzzles when learning only from the answers, we use probing techniques~\citep{adi2016fine, conneau2018you, hewitt2019designing, ye2024physics} to analyze their internal representations. Specifically, we study whether a {\ft}ed model's intermediate outputs provide evidence that it can distinguish between correct and incorrect statements for a given \kk puzzle, which is essential for solving the puzzle via reasoning. For a given model, we extract intermediate outputs from all transformer blocks for 200 correct and 200 incorrect statements, then check whether these outputs form distinct clusters by measuring the training accuracy of a logistic regression model fit on them (see  \Cref{app:probing} for details). For each $N$-people \kk puzzle, we report the per-layer probing accuracy averaged across seven {\ft}ed models, each FTed on an $M \in \{2,\ldots,8\}$-people task.

\Cref{fig:probe} shows \textbf{(1)} a clear trend of higher probing accuracy in deeper layers, peaking at around the 14th/15th layer.
The near-perfect peak accuracy suggests that the model's internal representations have a clear distinction between true/false statements about a given puzzle. \textbf{(2)} The probing accuracy is much higher than the un-FTed model (\Cref{fig:probe-base} in \Cref{app:add_exp_results}), suggesting that such representations are learned from the question-answer pairs during \ft. \textbf{(3)} Puzzles with more people seem to demand more internal computation, as evidenced by the point where probing accuracy surpasses $85\%$ shifting to later transformer blocks.

\subsection{{\ft} with Wrong Answers}
\label{subsec:wrong-ans-ft}

\begin{figure}[t]
    \centering
    \includegraphics[width=0.95\linewidth]{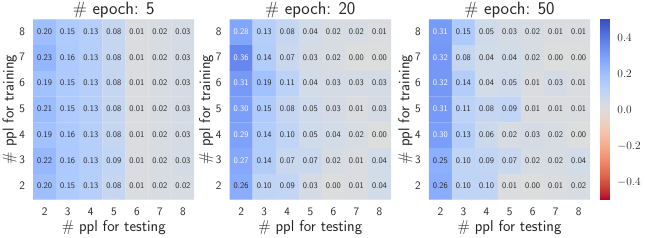}
    \vspace{-2mm}
    \caption{Test accuracy improves on $N$-people puzzles for \llamathree Direct FTed on $M$-people puzzles \textit{with completely wrong answers}, compared to the unfine-tuned model.
    This evaluation uses 1-shot direct prompting (see \Cref{fig:wrong-ans-ft-100} for results under different prompting setups).
    }
    \label{fig:wrong-ans-ft}
\end{figure}

To further explore what could the models learn from the question-answer pairs without detailed reasoning steps, we consider an extreme scenario of learning with incorrect answers:  for each $N$-people training puzzle, we randomly select $\tilde{N}$ from $[1,N]$ and flip the knight/knave identities of $\tilde{N}$ randomly chosen individuals.
Surprisingly, \Cref{fig:wrong-ans-ft} shows that \ft with incorrect answers still leads to non-trivial improvements for \llamathree. These improvements occur gradually over more epochs, suggesting that the model progressively developed reasoning skills during fine-tuning. 

Note that in this case the improved test accuracy could not have come from \revise{pure} memorization because 100\% of the training examples are incorrectly labeled. However, since in each wrong answer of a $N$-people puzzle, there are still $N-\tilde{N}$ correct role assignments where the random $\tilde{N}\geq1$. The model might have learned to reason from those partially correct role assignments in the wrong answer.

\begin{wrapfigure}{r}{0.62\linewidth}
    \centering
    \vspace{-5mm}
    \includegraphics[width=1\linewidth]{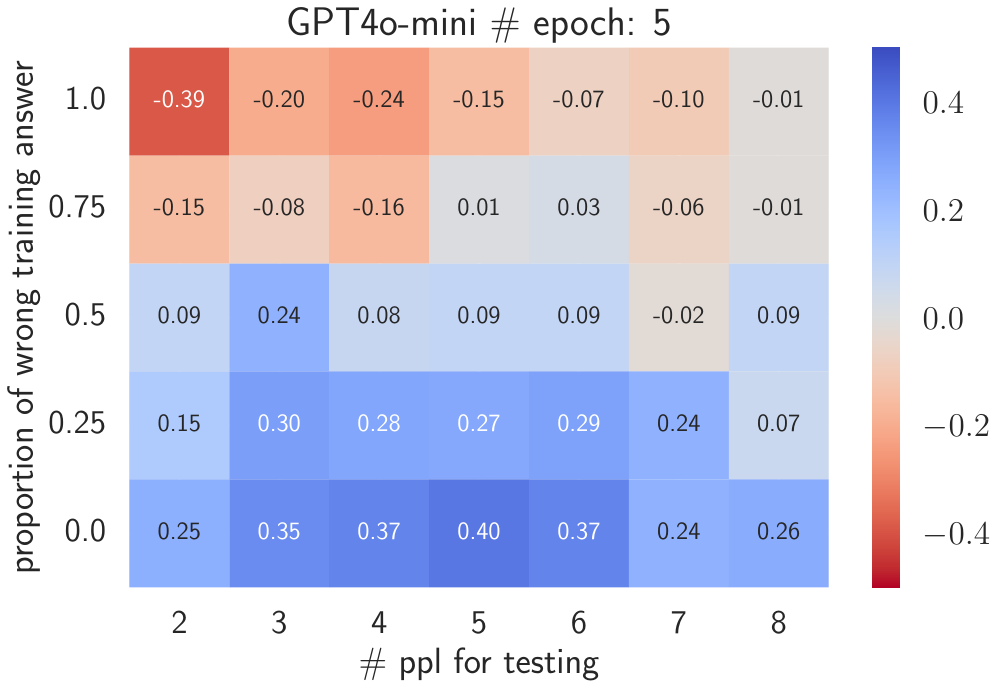}
    \vspace{-7mm}
     \caption{\ft w/ various wrong training answer proportions on 5-ppl task.
     }
    \vspace{-2mm}
     \label{fig:wrong-ans-ft-4omini}
\end{wrapfigure}
However, as shown in \cref{fig:wrong-ans-ft-4omini}, when applied to more capable model \gptfouromini, Direct FT on 5-people puzzles where 100\% training examples have corrupted answers does not lead to improvement. Moreover, the negative effects transfer to other tasks, notably easier ones (2/3/4-people). Nevertheless, as the percentage of corrupt-answer training examples reduces ($\leq 50\%$), the model could gain improved reasoning capabilities that generalize across different $N$-people tasks. We provide \gptfouromini results under more epochs in \cref{fig:wrong-ans-ft-4omini-ep345} and \llamathree results for partially wrong answer FT in  \Cref{fig:wrong-ans-ft-75,fig:wrong-ans-ft-50}.

\vspace{-2mm}
\section{Learn to Reason by Fine-tuning with CoTs}
\label{sec:learn-to-reason-cot}

Here we measure models' reasoning capabilities after fine-tuning with detailed reasoning steps.

\begin{wrapfigure}{l}{0.42\linewidth}
    \centering
     \vspace{-5mm}
\includegraphics[width=\linewidth]{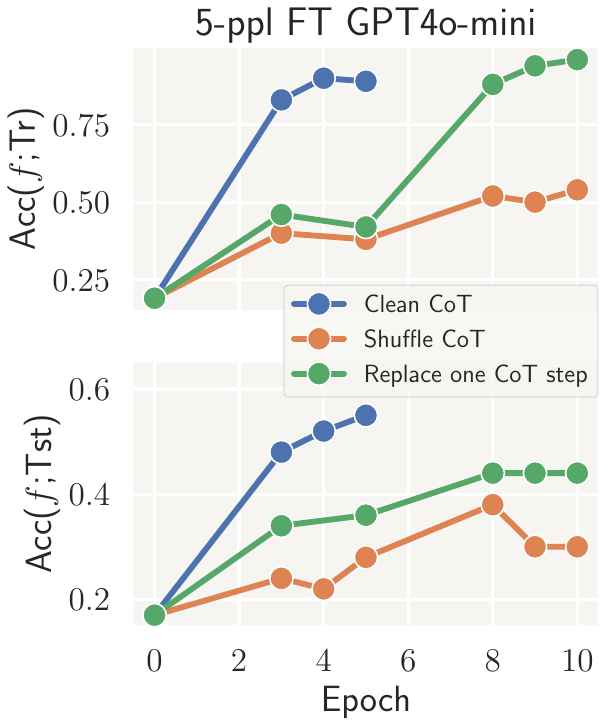}
    \vspace{-7mm}
    \caption{
    \small Wrong CoTs FT.
    }
    \vspace{-4mm}
    \label{fig:wrong_data_ft}
\end{wrapfigure}
\textbf{Model learns to reason on CoT when model capacity is large enough}. As shown in \cref{fig:xtrain_ytest_acc_ft},  
\textbf{(1)} training with reasoning steps as guidance improves test accuracy ($y$-axis) on unseen puzzles.  
\textbf{(2)} However, \llamathree struggles with \cotft, likely due to its limited capacity to effectively learn CoT skills with $\leq$1K training samples.
\textbf{(3)} Similar to \ft results in  \cref{sec:learn-to-reason-answer}, in \cotft,  memorization of training data is higher than test data (\cref{fig:perturb_ablation_direct_ft_mem_score}),  yet \revise{inconsistency ratio decreases despite that overall memorization score increases over training (\cref{fig:test_acc_mem_4omini})},  and the fine-tuned models show positive transferability to easier/harder tasks (\cref{fig:transfer_grid_direct_ft}). \textbf{(4)} Though LLMs can generalize surprisingly well under \ft, \cotft could lead to much higher test accuracy, especially with a larger training set (\Cref{fig:8ppl-10k}).

\textbf{Fine-tuning with wrong CoTs.}
The CoT training data includes both reasoning steps and answers. To understand the role of the CoT component in improving model generalization, we fine-tune \gptfouromini with two types of incorrect CoT data: \textbf{(a)} randomly shuffled CoT steps, disrupting the logic of the reasoning steps; and \textbf{(b)} CoTs with a single incorrect step, simulating genuine mistakes that people would sometimes make, where one step is randomly replaced with another puzzle’s CoT step (adjusting names to fit the current context). The results in \cref{fig:wrong_data_ft} show that 
\textbf{(1)} fine-tuning with a 100\% corrupted CoT dataset can still enhance test accuracy over the epochs, suggesting that the model learns to reason (potentially from the correct answers) despite CoT errors.
\textbf{(2)} Altering one CoT step slows convergence and reduces test accuracy compared to clean CoT. 
\textbf{(3)} Shuffling CoT steps further harms both convergence and generalization.
These also suggest that using correct logical chains in CoT can help LLMs to more effectively learn to reason.

\vspace{-2mm}
\section{Distinguish Memorization from Reasoning}
\label{sec:mem-reason-classification}

The findings above show that models' reasoning capabilities continue to improve as they memorize more training examples. In other words, the models use both memorization and reasoning to solve the puzzles. But how do they decide which examples to memorize or reason about? We can explore this by extending our memorization score to a per-example metric. Specifically, consider measuring \Cref{eq:limem} on a 1-point dataset $\mathcal{D}=\{x\}$. We skip the examples where $\accd[\{x\}]=0$ as the consistency ratio $\consistr(\{x\})$ is NaN in this case. Here, $\limemx\in{0,1}$ acts as a binary indicator: 0 means $x$ remains solvable after local perturbation, while 1 means it does not. Our goal is to determine if a clear rule separates these two types of puzzles.

\textbf{Setup.}  We collect training samples where the targeted LLM predicts correctly, and label each as either ``consistently solved'' (reasoning) or ``not consistently solved'' (memorization). We then train a logistic regression classifier using an 8:2 train/test split to distinguish between puzzles the model solves through reasoning versus memorization.

\begin{figure}[t]
    \centering
    \includegraphics[width=0.27\linewidth]{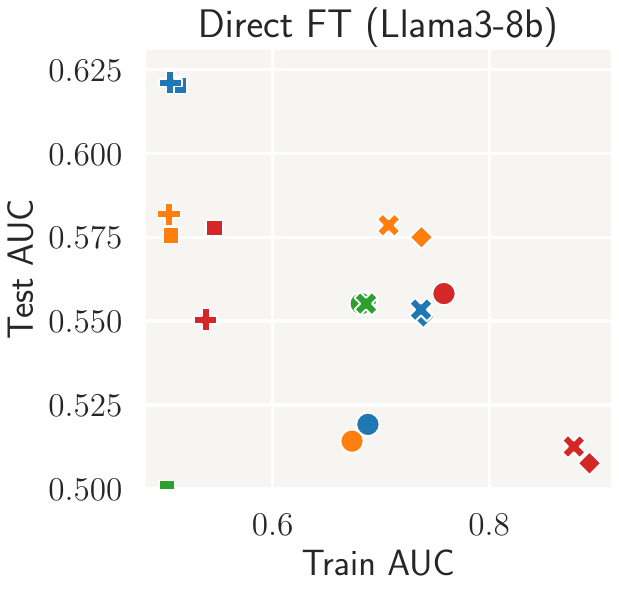}
     \includegraphics[width=0.71\linewidth]{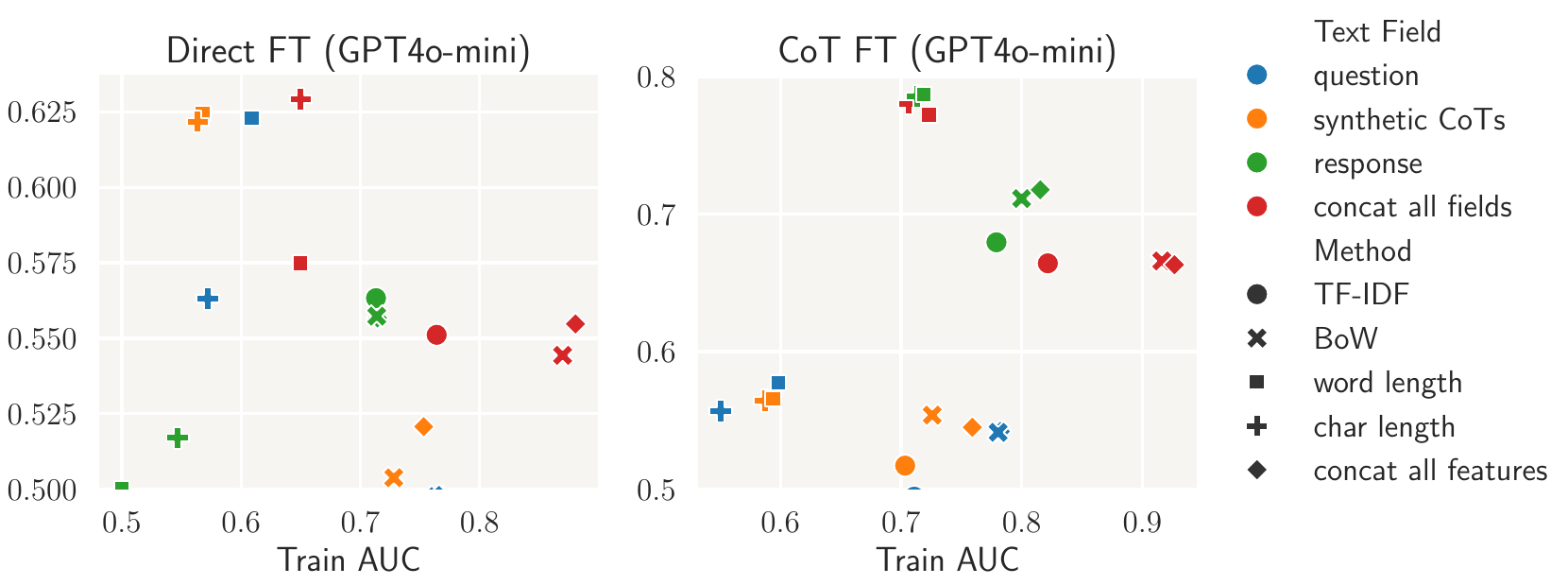}
    \caption{
    \small AUC for classifying 3-people puzzles under leaf perturbation based on puzzle-based indicators.  Results under more tasks and perturbations are in \Cref{fig:identify_robust_puzzle_more}.
    }
    \label{fig:identify_robust_puzzle}
\end{figure}

\textbf{Puzzle-based indicators.} We consider the following features: (1) TF-IDF; (2) Bag-of-Words; (3) Word Length; (4) Character Length; (5) concatenation of all. Each feature can be extracted from one of the following fields: (1) question; (2) synthetic CoT reasoning steps; 
(3) model response\footnote{Strictly speaking, this is a model-based indicator feature.}; (4) concatenation of the above fields. The train and test performance (measured with AUC as the dataset can be unbalanced) are shown in \cref{fig:identify_robust_puzzle}. We observe a test AUC of 0.629/0.787 for Direct/\cotft-ed \gptfouromini, and 0.627 for {\ft}-ed \llamathree. 
This indicates that the puzzle-based indicators could be informative, though not perfect, at determining which examples are reasoned vs. memorized.

\textbf{Model-based indicators.} 
We also study model-based indicators to test whether the internal activations of the fine-tuned model are informative for this categorization. Since we cannot access model internals of \gptfouromini, we conduct the experiment on \llamathree. Specifically, we feed each puzzle question into the fine-tuned model, extract average embeddings from each layer, and train a linear classifier per layer. \revise{Appendix} \Cref{fig:identify_robust_llama} shows test AUCs, where we also compare the fine-tuned model to its non-fine-tuned counterpart. We observe that \textbf{(1)} Lower-layer features poorly distinguish memorization from reasoning, but higher layers improve. \textbf{(2)} The features from the FTed model are consistently more informative than the un-FTed one, suggesting that the model's decision regarding memorization vs. reasoning on specific samples likely stems from the fine-tuning process. \textbf{(3)} The best embedding-based indicator (0.70 AUC) outperforms the puzzle-based indicator (\cref{fig:identify_robust_puzzle} left, 0.627 AUC) on 3-people puzzles.

\vspace{-3mm}
\section{Related Work}
\label{sec:related}
\vspace{-1mm}

\textbf{Memorization in LLMs.}
Previous work on LLM memorization primarily focused on near-verbatim training text regurgitation from the perspective of privacy or copyright \revise{concerns~\citep{carlini2021extracting,lee2022deduplicating,carlini22quantifying, lukas2023analyzing, biderman2024emergent,prashanth2024recite}}. In contrast, we focus on quantifying the memorization behavior of LLMs when solving reasoning tasks, using a metric computed with the help of local perturbation of reasoning puzzles.

\textbf{Benchmark Contamination and Logical Reasoning Evaluation.}
Recent research has revealed LLMs' significant performance decline when faced with altered versions of popular reasoning benchmarks~\citep{oren2023proving,xu2024benchmarking,yang2023rethinking,yao2024data,zhang2024careful, srivastava2024functional}, suggesting potential benchmark contamination. 
Various synthetic benchmarks have been developed to evaluate LLMs' logical reasoning capabilities, allowing for dynamic and scalable generation of samples with different configurations and difficulty levels~\citep{Clark2020TransformersAS, giadikiaroglou2024puzzle, Parmar2024LogicBenchTS,dziri2024faith,zebralogic2024,kazemi2024boardgameqa,mondorf2024liar}. TruthQuest~\citep {mondorf2024liar} is the most similar task to our work, which provides \kk-type of 3-6 person puzzles and answers. Our work provides more comprehensive {dynamic} set of \kk puzzles that support the automatic generation of perturbations, solutions, and detailed reasoning steps. Moreover, we define and measure memorization, and reveal its intricate relation to reasoning. 
\revise{We refer the readers to \cref{sec:extend_related} for a more comprehensive discussion of related work (e.g., grokking and adversarial perturbations).}

\vspace{-3mm}
\section{Conclusion}
\vspace{-1mm}

We propose a memorization metric $\limem$ based on the inconsistency when solving a locally perturbed logical reasoning puzzle, and quantitatively characterize the amount of memorization and reasoning. 
Through an in-depth analysis based on local perturbation, transferability, intermediate outputs probing, and fine-tuning with wrong answers, we find that LLMs learn to reason as they memorize more training examples. Furthermore, we study input and model-based 
signals that may determine which puzzles are solved by reasoning vs memorization. To support these studies, we create a feature-rich dynamic logical reasoning benchmark that not only enables our memorization study, but could also be useful for future studies related to LLM logical reasoning. 
\revise{We defer more discussion and future work to \cref{app:future_work}.}

\newpage

\section*{Acknowledgments}
We thank Yuntian Deng, Mingyang Deng,  Ziqi Wang, Tiancheng Yu, Mike Mozer, Rishabh Agarwal, Danqi Chen, Matthew Jagielski, Nikunj Saunshi, Wei Xiong and Minghao Chen for their valuable feedback and discussions. Part of this work was completed while Yangsibo Huang was a PhD student at Princeton, and she acknowledges the support of the Wallace Memorial Fellowship and the compute resources at Princeton Language and Intelligence. Bo Li acknowledges the support of NSF No. 2046726, NSF AI Institute ACTION No. IIS-2229876 and the Alfred P. Sloan Fellowship.

\bibliographystyle{icml2025}
\bibliography{ref}

\begin{thebibliography}{86}
\providecommand{\natexlab}[1]{#1}
\providecommand{\url}[1]{\texttt{#1}}
\expandafter\ifx\csname urlstyle\endcsname\relax
  \providecommand{\doi}[1]{doi: #1}\else
  \providecommand{\doi}{doi: \begingroup \urlstyle{rm}\Url}\fi

\bibitem[Adi et~al.(2017)Adi, Kermany, Belinkov, Lavi, and Goldberg]{adi2016fine}
Adi, Y., Kermany, E., Belinkov, Y., Lavi, O., and Goldberg, Y.
\newblock Fine-grained analysis of sentence embeddings using auxiliary prediction tasks.
\newblock In \emph{ICLR}, 2017.

\bibitem[Balloccu et~al.(2024)Balloccu, Schmidtov{\'a}, Lango, and Du{\v{s}}ek]{balloccu2024leak}
Balloccu, S., Schmidtov{\'a}, P., Lango, M., and Du{\v{s}}ek, O.
\newblock Leak, cheat, repeat: Data contamination and evaluation malpractices in closed-source {LLMs}.
\newblock \emph{arXiv preprint arXiv:2402.03927}, 2024.

\bibitem[Bartlett et~al.(2021)Bartlett, Montanari, and Rakhlin]{bartlett2021deep}
Bartlett, P.~L., Montanari, A., and Rakhlin, A.
\newblock Deep learning: a statistical viewpoint.
\newblock \emph{Acta Numerica}, 2021.

\bibitem[Belkin(2021)]{belkin2021fit}
Belkin, M.
\newblock Fit without fear: remarkable mathematical phenomena of deep learning through the prism of interpolation.
\newblock \emph{Acta Numerica}, 2021.

\bibitem[Belkin et~al.(2018)Belkin, Hsu, and Mitra]{belkin2018overfitting}
Belkin, M., Hsu, D.~J., and Mitra, P.
\newblock Overfitting or perfect fitting? {R}isk bounds for classification and regression rules that interpolate.
\newblock In \emph{NeurIPS}, 2018.

\bibitem[Biderman et~al.(2024)Biderman, Prashanth, Sutawika, Schoelkopf, Anthony, Purohit, and Raff]{biderman2024emergent}
Biderman, S., Prashanth, U., Sutawika, L., Schoelkopf, H., Anthony, Q., Purohit, S., and Raff, E.
\newblock Emergent and predictable memorization in large language models.
\newblock \emph{Advances in Neural Information Processing Systems}, 36, 2024.

\bibitem[{Boolean satisfiability problem}()]{wiki:booleansat}
{Boolean satisfiability problem}.
\newblock Boolean satisfiability problem --- {W}ikipedia{,} the free encyclopedia.
\newblock URL \url{https://en.wikipedia.org/wiki/Boolean_satisfiability_problem}.
\newblock [Online; accessed 20-Nov-2024].

\bibitem[Carlini et~al.(2021)Carlini, Tramer, Wallace, Jagielski, Herbert-Voss, Lee, Roberts, Brown, Song, Erlingsson, et~al.]{carlini2021extracting}
Carlini, N., Tramer, F., Wallace, E., Jagielski, M., Herbert-Voss, A., Lee, K., Roberts, A., Brown, T., Song, D., Erlingsson, U., et~al.
\newblock Extracting training data from large language models.
\newblock In \emph{USENIX Security}, 2021.

\bibitem[Carlini et~al.(2023)Carlini, Ippolito, Jagielski, Lee, Tramer, and Zhang]{carlini22quantifying}
Carlini, N., Ippolito, D., Jagielski, M., Lee, K., Tramer, F., and Zhang, C.
\newblock Quantifying memorization across neural language models.
\newblock In \emph{ICLR}, 2023.

\bibitem[Chen et~al.(2024)Chen, Chi, Wang, and Zhou]{chenpremise}
Chen, X., Chi, R.~A., Wang, X., and Zhou, D.
\newblock Premise order matters in reasoning with large language models.
\newblock In \emph{ICML}, 2024.

\bibitem[Chung et~al.(2024)Chung, Hou, Longpre, Zoph, Tay, Fedus, Li, Wang, Dehghani, Brahma, et~al.]{chung2024scaling}
Chung, H.~W., Hou, L., Longpre, S., Zoph, B., Tay, Y., Fedus, W., Li, Y., Wang, X., Dehghani, M., Brahma, S., et~al.
\newblock Scaling instruction-finetuned language models.
\newblock \emph{Journal of Machine Learning Research}, 2024.

\bibitem[Clark et~al.(2020)Clark, Tafjord, and Richardson]{Clark2020TransformersAS}
Clark, P., Tafjord, O., and Richardson, K.
\newblock Transformers as soft reasoners over language.
\newblock In \emph{IJCAI}, 2020.

\bibitem[Conneau et~al.(2018)Conneau, Kruszewski, Lample, Barrault, and Baroni]{conneau2018you}
Conneau, A., Kruszewski, G., Lample, G., Barrault, L., and Baroni, M.
\newblock What you can cram into a single vector: Probing sentence embeddings for linguistic properties.
\newblock In \emph{ACL}, 2018.

\bibitem[Deng et~al.(2023)Deng, Prasad, Fernandez, Smolensky, Chaudhary, and Shieber]{deng2023implicit}
Deng, Y., Prasad, K., Fernandez, R., Smolensky, P., Chaudhary, V., and Shieber, S.
\newblock Implicit chain of thought reasoning via knowledge distillation.
\newblock \emph{arXiv preprint arXiv:2311.01460}, 2023.

\bibitem[Deng et~al.(2024)Deng, Choi, and Shieber]{deng2024explicit}
Deng, Y., Choi, Y., and Shieber, S.
\newblock From explicit cot to implicit cot: Learning to internalize cot step by step.
\newblock \emph{arXiv preprint arXiv:2405.14838}, 2024.

\bibitem[Dziri et~al.(2024)Dziri, Lu, Sclar, Li, Jian, Lin, West, Bhagavatula, Bras, Hwang, Sanyal, Welleck, Ren, Ettinger, Harchaoui, and Choi]{dziri2024faith}
Dziri, N., Lu, X., Sclar, M., Li, X.~L., Jian, L., Lin, B.~Y., West, P., Bhagavatula, C., Bras, R.~L., Hwang, J.~D., Sanyal, S., Welleck, S., Ren, X., Ettinger, A., Harchaoui, Z., and Choi, Y.
\newblock Faith and fate: Limits of transformers on compositionality.
\newblock \emph{NeurIPS}, 2024.

\bibitem[Gao et~al.(2020)Gao, Biderman, Black, Golding, Hoppe, Foster, Phang, He, Thite, Nabeshima, et~al.]{gao2020pile}
Gao, L., Biderman, S., Black, S., Golding, L., Hoppe, T., Foster, C., Phang, J., He, H., Thite, A., Nabeshima, N., et~al.
\newblock The pile: An 800gb dataset of diverse text for language modeling.
\newblock \emph{arXiv preprint arXiv:2101.00027}, 2020.

\bibitem[Giadikiaroglou et~al.(2024)Giadikiaroglou, Lymperaiou, Filandrianos, and Stamou]{giadikiaroglou2024puzzle}
Giadikiaroglou, P., Lymperaiou, M., Filandrianos, G., and Stamou, G.
\newblock Puzzle solving using reasoning of large language models: A survey.
\newblock In \emph{IJCAI}, 2024.

\bibitem[Golchin \& Surdeanu(2023)Golchin and Surdeanu]{golchin2023data}
Golchin, S. and Surdeanu, M.
\newblock Data contamination quiz: A tool to detect and estimate contamination in large language models.
\newblock \emph{arXiv preprint arXiv:2311.06233}, 2023.

\bibitem[Gupta et~al.(2024)Gupta, Pantoja, Ross, Williams, and Ung]{gupta2024changing}
Gupta, V., Pantoja, D., Ross, C., Williams, A., and Ung, M.
\newblock Changing answer order can decrease {MMLU} accuracy.
\newblock \emph{arXiv preprint arXiv:2406.19470}, 2024.

\bibitem[Hartmann et~al.(2023)Hartmann, Suri, Bindschaedler, Evans, Tople, and West]{hartmann2023sok}
Hartmann, V., Suri, A., Bindschaedler, V., Evans, D., Tople, S., and West, R.
\newblock {SoK:} memorization in general-purpose large language models.
\newblock \emph{arXiv preprint arXiv:2310.18362}, 2023.

\bibitem[He et~al.(2024)He, Huang, Shi, Xie, Liu, Wang, Zettlemoyer, Zhang, Chen, and Henderson]{he2024fantastic}
He, L., Huang, Y., Shi, W., Xie, T., Liu, H., Wang, Y., Zettlemoyer, L., Zhang, C., Chen, D., and Henderson, P.
\newblock Fantastic copyrighted beasts and how (not) to generate them.
\newblock \emph{arXiv preprint arXiv:2406.14526}, 2024.

\bibitem[Hewitt \& Liang(2019)Hewitt and Liang]{hewitt2019designing}
Hewitt, J. and Liang, P.
\newblock Designing and interpreting probes with control tasks.
\newblock In \emph{EMNLP}, 2019.

\bibitem[Ho et~al.(2023)Ho, Schmid, and Yun]{ho2022large}
Ho, N., Schmid, L., and Yun, S.-Y.
\newblock Large language models are reasoning teachers.
\newblock In \emph{ACL}, 2023.

\bibitem[Hsieh et~al.(2023)Hsieh, Li, Yeh, Nakhost, Fujii, Ratner, Krishna, Lee, and Pfister]{hsieh2023distilling}
Hsieh, C.-Y., Li, C.-L., Yeh, C.-K., Nakhost, H., Fujii, Y., Ratner, A., Krishna, R., Lee, C.-Y., and Pfister, T.
\newblock Distilling step-by-step! outperforming larger language models with less training data and smaller model sizes.
\newblock In \emph{ACL}, 2023.

\bibitem[Jain et~al.(2024)Jain, Han, Gu, Li, Yan, Zhang, Wang, Solar-Lezama, Sen, and Stoica]{jain2024livecodebench}
Jain, N., Han, K., Gu, A., Li, W.-D., Yan, F., Zhang, T., Wang, S., Solar-Lezama, A., Sen, K., and Stoica, I.
\newblock {LiveCodeBench}: Holistic and contamination free evaluation of large language models for code.
\newblock \emph{arXiv preprint arXiv:2403.07974}, 2024.

\bibitem[Jiang et~al.(2024)Jiang, Xie, Hao, Wang, Mallick, Su, Taylor, and Roth]{jiang2024peek}
Jiang, B., Xie, Y., Hao, Z., Wang, X., Mallick, T., Su, W.~J., Taylor, C.~J., and Roth, D.
\newblock A peek into token bias: Large language models are not yet genuine reasoners.
\newblock \emph{EMNLP}, 2024.

\bibitem[Jin et~al.(2020)Jin, Jin, Zhou, and Szolovits]{jin2019bert}
Jin, D., Jin, Z., Zhou, J.~T., and Szolovits, P.
\newblock Is bert really robust? natural language attack on text classification and entailment.
\newblock \emph{AAAI}, 2020.

\bibitem[Johnson-Laird \& Byrne(1990)Johnson-Laird and Byrne]{johnson1990meta}
Johnson-Laird, P.~N. and Byrne, R.~M.
\newblock Meta-logical problems: Knights, knaves, and rips.
\newblock \emph{Cognition}, 1990.

\bibitem[Karamolegkou et~al.(2023)Karamolegkou, Li, Zhou, and S{\o}gaard]{karamolegkou2023copyright}
Karamolegkou, A., Li, J., Zhou, L., and S{\o}gaard, A.
\newblock Copyright violations and large language models.
\newblock In \emph{EMNLP}, 2023.

\bibitem[Kazemi et~al.(2024)Kazemi, Yuan, Bhatia, Kim, Xu, Imbrasaite, and Ramachandran]{kazemi2024boardgameqa}
Kazemi, M., Yuan, Q., Bhatia, D., Kim, N., Xu, X., Imbrasaite, V., and Ramachandran, D.
\newblock {BoardgameQA}: A dataset for natural language reasoning with contradictory information.
\newblock In \emph{NeurIPS}, 2024.

\bibitem[Kim et~al.(2023)Kim, Joo, Kim, Jang, Ye, Shin, and Seo]{kim2023cot}
Kim, S., Joo, S.~J., Kim, D., Jang, J., Ye, S., Shin, J., and Seo, M.
\newblock The cot collection: Improving zero-shot and few-shot learning of language models via chain-of-thought fine-tuning.
\newblock In \emph{EMNLP}, 2023.

\bibitem[Lee et~al.(2024)Lee, Bai, Pres, Wattenberg, Kummerfeld, and Mihalcea]{lee2024mechanistic}
Lee, A., Bai, X., Pres, I., Wattenberg, M., Kummerfeld, J.~K., and Mihalcea, R.
\newblock A mechanistic understanding of alignment algorithms: A case study on {DPO} and toxicity.
\newblock In \emph{ICML}, 2024.

\bibitem[Lee et~al.(2022)Lee, Ippolito, Nystrom, Zhang, Eck, Callison-Burch, and Carlini]{lee2022deduplicating}
Lee, K., Ippolito, D., Nystrom, A., Zhang, C., Eck, D., Callison-Burch, C., and Carlini, N.
\newblock Deduplicating training data makes language models better.
\newblock In \emph{ACL}, 2022.

\bibitem[Lin(2024)]{1311vs138}
Lin, B.~Y.
\newblock Math {O}lympiad becomes easier for {AI}; {C}ommon sense is still hard., 2024.
\newblock URL \url{https://x.com/billyuchenlin/status/1812948314360541302}.

\bibitem[Lin et~al.(2024)Lin, Bras, and Choi]{zebralogic2024}
Lin, B.~Y., Bras, R.~L., and Choi, Y.
\newblock {ZebraLogic:} benchmarking the logical reasoning ability of language models, 2024.
\newblock URL \url{https://hf.co/spaces/allenai/ZebraLogic}.

\bibitem[Liu et~al.(2022{\natexlab{a}})Liu, Kitouni, Nolte, Michaud, Tegmark, and Williams]{liu2022towards}
Liu, Z., Kitouni, O., Nolte, N.~S., Michaud, E., Tegmark, M., and Williams, M.
\newblock Towards understanding grokking: An effective theory of representation learning.
\newblock \emph{NeurIPS}, 2022{\natexlab{a}}.

\bibitem[Liu et~al.(2022{\natexlab{b}})Liu, Michaud, and Tegmark]{liu2022omnigrok}
Liu, Z., Michaud, E.~J., and Tegmark, M.
\newblock Omnigrok: Grokking beyond algorithmic data.
\newblock In \emph{The Eleventh International Conference on Learning Representations}, 2022{\natexlab{b}}.

\bibitem[Lukas et~al.(2023)Lukas, Salem, Sim, Tople, Wutschitz, and Zanella-B{\'e}guelin]{lukas2023analyzing}
Lukas, N., Salem, A., Sim, R., Tople, S., Wutschitz, L., and Zanella-B{\'e}guelin, S.
\newblock Analyzing leakage of personally identifiable information in language models.
\newblock In \emph{IEEE Symposium on Security and Privacy (SP)}, 2023.

\bibitem[Magar \& Schwartz(2022)Magar and Schwartz]{magar2022data}
Magar, I. and Schwartz, R.
\newblock Data contamination: From memorization to exploitation.
\newblock \emph{arXiv preprint arXiv:2203.08242}, 2022.

\bibitem[McCoy et~al.(2024)McCoy, Yao, Friedman, Hardy, and Griffiths]{mccoy2024embers}
McCoy, R.~T., Yao, S., Friedman, D., Hardy, M.~D., and Griffiths, T.~L.
\newblock Embers of autoregression show how large language models are shaped by the problem they are trained to solve.
\newblock \emph{Proceedings of the National Academy of Sciences}, 121\penalty0 (41):\penalty0 e2322420121, 2024.

\bibitem[Mirzadeh et~al.(2024)Mirzadeh, Alizadeh, Shahrokhi, Tuzel, Bengio, and Farajtabar]{mirzadeh2024gsm}
Mirzadeh, I., Alizadeh, K., Shahrokhi, H., Tuzel, O., Bengio, S., and Farajtabar, M.
\newblock Gsm-symbolic: Understanding the limitations of mathematical reasoning in large language models.
\newblock \emph{arXiv preprint arXiv:2410.05229}, 2024.

\bibitem[Mondorf \& Plank(2024)Mondorf and Plank]{mondorf2024liar}
Mondorf, P. and Plank, B.
\newblock Liar, liar, logical mire: A benchmark for suppositional reasoning in large language models.
\newblock \emph{arXiv preprint arXiv:2406.12546}, 2024.

\bibitem[Murty et~al.(2023)Murty, Sharma, Andreas, and Manning]{murty2023grokking}
Murty, S., Sharma, P., Andreas, J., and Manning, C.~D.
\newblock Grokking of hierarchical structure in vanilla transformers.
\newblock In \emph{ACL}, 2023.

\bibitem[Muthukumar et~al.(2020)Muthukumar, Vodrahalli, Subramanian, and Sahai]{muthukumar2020harmless}
Muthukumar, V., Vodrahalli, K., Subramanian, V., and Sahai, A.
\newblock Harmless interpolation of noisy data in regression.
\newblock \emph{IEEE Journal on Selected Areas in Information Theory}, 2020.

\bibitem[Nezhurina et~al.(2024)Nezhurina, Cipolina-Kun, Cherti, and Jitsev]{nezhurina2024alice}
Nezhurina, M., Cipolina-Kun, L., Cherti, M., and Jitsev, J.
\newblock Alice in wonderland: Simple tasks showing complete reasoning breakdown in state-of-the-art large language models.
\newblock \emph{arXiv preprint arXiv:2406.02061}, 2024.

\bibitem[Nie et~al.(2020)Nie, Williams, Dinan, Bansal, Weston, and Kiela]{nie2020adversarial}
Nie, Y., Williams, A., Dinan, E., Bansal, M., Weston, J., and Kiela, D.
\newblock Adversarial {NLI}: A new benchmark for natural language understanding.
\newblock In \emph{ACL}, pp.\  4885--4901. Association for Computational Linguistics, 2020.
\newblock \doi{10.18653/v1/2020.acl-main.441}.
\newblock URL \url{https://aclanthology.org/2020.acl-main.441/}.

\bibitem[Oren et~al.(2024)Oren, Meister, Chatterji, Ladhak, and Hashimoto]{oren2023proving}
Oren, Y., Meister, N., Chatterji, N., Ladhak, F., and Hashimoto, T.~B.
\newblock Proving test set contamination in black box language models.
\newblock \emph{ICLR}, 2024.

\bibitem[Pan et~al.(2023)Pan, Albalak, Wang, and Wang]{pan2023logic}
Pan, L., Albalak, A., Wang, X., and Wang, W.~Y.
\newblock Logic-lm: Empowering large language models with symbolic solvers for faithful logical reasoning.
\newblock In \emph{The 2023 Conference on Empirical Methods in Natural Language Processing}, 2023.

\bibitem[Parmar et~al.(2024)Parmar, Patel, Varshney, Nakamura, Luo, Mashetty, Mitra, and Baral]{Parmar2024LogicBenchTS}
Parmar, M., Patel, N., Varshney, N., Nakamura, M., Luo, M., Mashetty, S., Mitra, A., and Baral, C.
\newblock {LogicBench:} towards systematic evaluation of logical reasoning ability of large language models.
\newblock In \emph{ACL}, 2024.

\bibitem[Power et~al.(2022)Power, Burda, Edwards, Babuschkin, and Misra]{power2022grokking}
Power, A., Burda, Y., Edwards, H., Babuschkin, I., and Misra, V.
\newblock Grokking: Generalization beyond overfitting on small algorithmic datasets.
\newblock \emph{arXiv preprint arXiv:2201.02177}, 2022.

\bibitem[Prabhakar et~al.(2024)Prabhakar, Griffiths, and McCoy]{prabhakar2024deciphering}
Prabhakar, A., Griffiths, T.~L., and McCoy, R.~T.
\newblock Deciphering the factors influencing the efficacy of chain-of-thought: Probability, memorization, and noisy reasoning.
\newblock \emph{arXiv preprint arXiv:2407.01687}, 2024.

\bibitem[Prashanth et~al.(2024)Prashanth, Deng, O'Brien, SV, Khan, Borkar, Choquette-Choo, Fuehne, Biderman, Ke, et~al.]{prashanth2024recite}
Prashanth, U.~S., Deng, A., O'Brien, K., SV, J., Khan, M.~A., Borkar, J., Choquette-Choo, C.~A., Fuehne, J.~R., Biderman, S., Ke, T., et~al.
\newblock Recite, reconstruct, recollect: Memorization in lms as a multifaceted phenomenon.
\newblock \emph{arXiv preprint arXiv:2406.17746}, 2024.

\bibitem[Puerto et~al.(2024)Puerto, Chubakov, Zhu, Madabushi, and Gurevych]{puerto2024fine}
Puerto, H., Chubakov, T., Zhu, X., Madabushi, H.~T., and Gurevych, I.
\newblock Fine-tuning with divergent chains of thought boosts reasoning through self-correction in language models.
\newblock \emph{arXiv preprint arXiv:2407.03181}, 2024.

\bibitem[Razeghi et~al.(2022)Razeghi, Logan~IV, Gardner, and Singh]{razeghi2022impact}
Razeghi, Y., Logan~IV, R.~L., Gardner, M., and Singh, S.
\newblock Impact of pretraining term frequencies on few-shot numerical reasoning.
\newblock In \emph{Findings of EMNLP 2022}, pp.\  840--854, 2022.

\bibitem[Roberts et~al.(2023)Roberts, Thakur, Herlihy, White, and Dooley]{roberts2023cutoff}
Roberts, M., Thakur, H., Herlihy, C., White, C., and Dooley, S.
\newblock To the cutoff... and beyond? {A} longitudinal perspective on {LLM} data contamination.
\newblock In \emph{ICLR}, 2023.

\bibitem[Robinson \& Wingate(2023)Robinson and Wingate]{robinson2023leveraging}
Robinson, J. and Wingate, D.
\newblock Leveraging large language models for multiple choice question answering.
\newblock In \emph{ICLR}, 2023.

\bibitem[Saparov \& He(2023)Saparov and He]{saparov2023language}
Saparov, A. and He, H.
\newblock Language models are greedy reasoners: A systematic formal analysis of chain-of-thought.
\newblock In \emph{The Eleventh International Conference on Learning Representations}, 2023.
\newblock URL \url{https://openreview.net/forum?id=qFVVBzXxR2V}.

\bibitem[{SAT solver}(2024)]{wiki:satsolver}
{SAT solver}.
\newblock Sat solver --- {W}ikipedia{,} the free encyclopedia, 2024.
\newblock URL \url{https://en.wikipedia.org/wiki/SAT_solver}.
\newblock [Online; accessed 20-Nov-2024].

\bibitem[Shi et~al.(2024)Shi, Ajith, Xia, Huang, Liu, Blevins, Chen, and Zettlemoyer]{shidetecting}
Shi, W., Ajith, A., Xia, M., Huang, Y., Liu, D., Blevins, T., Chen, D., and Zettlemoyer, L.
\newblock Detecting pretraining data from large language models.
\newblock In \emph{ICLR}, 2024.

\bibitem[Smullyan(1978)]{KKorig}
Smullyan, R.
\newblock \emph{What is the Name of this Book?}
\newblock Prentice-Hall, 1978.

\bibitem[Srivastava et~al.(2024)Srivastava, PV, Menon, Sukumar, Philipose, Prince, Thomas, et~al.]{srivastava2024functional}
Srivastava, S., PV, A., Menon, S., Sukumar, A., Philipose, A., Prince, S., Thomas, S., et~al.
\newblock Functional benchmarks for robust evaluation of reasoning performance, and the reasoning gap.
\newblock \emph{arXiv preprint arXiv:2402.19450}, 2024.

\bibitem[Tirumala et~al.(2022)Tirumala, Markosyan, Zettlemoyer, and Aghajanyan]{tirumala2022memorization}
Tirumala, K., Markosyan, A., Zettlemoyer, L., and Aghajanyan, A.
\newblock Memorization without overfitting: Analyzing the training dynamics of large language models.
\newblock \emph{NeurIPS}, 2022.

\bibitem[Wallace et~al.(2024)Wallace, Xiao, Leike, Weng, Heidecke, and Beutel]{wallace2024instruction}
Wallace, E., Xiao, K., Leike, R., Weng, L., Heidecke, J., and Beutel, A.
\newblock The instruction hierarchy: Training {LLMs} to prioritize privileged instructions.
\newblock \emph{arXiv preprint arXiv:2404.13208}, 2024.

\bibitem[Wang et~al.(2021)Wang, Xu, Wang, Gan, Cheng, Gao, Awadallah, and Li]{wang2021adversarial}
Wang, B., Xu, C., Wang, S., Gan, Z., Cheng, Y., Gao, J., Awadallah, A.~H., and Li, B.
\newblock Adversarial {GLUE}: A multi-task benchmark for robustness evaluation of language models.
\newblock In \emph{NeurIPS Datasets and Benchmarks Track}, 2021.
\newblock URL \url{https://openreview.net/forum?id=GF9cSKI3A_q}.

\bibitem[Wang et~al.(2023{\natexlab{a}})Wang, Chen, Pei, Xie, Kang, Zhang, Xu, Xiong, Dutta, Schaeffer, et~al.]{wang2023decodingtrust}
Wang, B., Chen, W., Pei, H., Xie, C., Kang, M., Zhang, C., Xu, C., Xiong, Z., Dutta, R., Schaeffer, R., et~al.
\newblock {DecodingTrust:} a comprehensive assessment of trustworthiness in {GPT} models.
\newblock In \emph{NeurIPS}, 2023{\natexlab{a}}.

\bibitem[Wang et~al.(2024{\natexlab{a}})Wang, Yue, Su, and Sun]{wang2024grokking}
Wang, B., Yue, X., Su, Y., and Sun, H.
\newblock Grokking of implicit reasoning in transformers: A mechanistic journey to the edge of generalization.
\newblock In \emph{The Thirty-eighth Annual Conference on Neural Information Processing Systems}, 2024{\natexlab{a}}.
\newblock URL \url{https://openreview.net/forum?id=D4QgSWxiOb}.

\bibitem[Wang et~al.(2024{\natexlab{b}})Wang, Zhao, Qiang, Qin, and Liu]{wang2024beyond}
Wang, H., Zhao, S., Qiang, Z., Qin, B., and Liu, T.
\newblock Beyond the answers: Reviewing the rationality of multiple choice question answering for the evaluation of large language models.
\newblock \emph{arXiv preprint arXiv:2402.01349}, 2024{\natexlab{b}}.

\bibitem[Wang et~al.(2023{\natexlab{b}})Wang, Wei, Schuurmans, Le, Chi, Narang, Chowdhery, and Zhou]{wangself2023}
Wang, X., Wei, J., Schuurmans, D., Le, Q.~V., Chi, E.~H., Narang, S., Chowdhery, A., and Zhou, D.
\newblock Self-consistency improves chain of thought reasoning in language models.
\newblock In \emph{The Eleventh International Conference on Learning Representations}, 2023{\natexlab{b}}.

\bibitem[Wei et~al.(2024{\natexlab{a}})Wei, Huang, Huang, Xie, Qi, Xia, Mittal, Wang, and Henderson]{wei2024assessing}
Wei, B., Huang, K., Huang, Y., Xie, T., Qi, X., Xia, M., Mittal, P., Wang, M., and Henderson, P.
\newblock Assessing the brittleness of safety alignment via pruning and low-rank modifications.
\newblock In \emph{ICML}, 2024{\natexlab{a}}.

\bibitem[Wei et~al.(2024{\natexlab{b}})Wei, Shi, Huang, Smith, Zhang, Zettlemoyer, Li, and Henderson]{wei2024evaluating}
Wei, B., Shi, W., Huang, Y., Smith, N.~A., Zhang, C., Zettlemoyer, L., Li, K., and Henderson, P.
\newblock Evaluating copyright takedown methods for language models.
\newblock In \emph{NeurIPS Datasets and Benchmark}, 2024{\natexlab{b}}.

\bibitem[Wei et~al.(2022)Wei, Wang, Schuurmans, Bosma, Xia, Chi, Le, Zhou, et~al.]{wei2022chain}
Wei, J., Wang, X., Schuurmans, D., Bosma, M., Xia, F., Chi, E., Le, Q.~V., Zhou, D., et~al.
\newblock Chain-of-thought prompting elicits reasoning in large language models.
\newblock \emph{NeurIPS}, 2022.

\bibitem[Weng et~al.(2023)Weng, Zhu, Xia, Li, He, Liu, Sun, Liu, and Zhao]{weng2023large}
Weng, Y., Zhu, M., Xia, F., Li, B., He, S., Liu, S., Sun, B., Liu, K., and Zhao, J.
\newblock Large language models are better reasoners with self-verification.
\newblock In \emph{The 2023 Conference on Empirical Methods in Natural Language Processing}, 2023.

\bibitem[Wu et~al.(2024{\natexlab{a}})Wu, Yu, Huang, Russakovsky, and Arora]{wu2024conceptmix}
Wu, X., Yu, D., Huang, Y., Russakovsky, O., and Arora, S.
\newblock {ConceptMix}: A compositional image generation benchmark with controllable difficulty.
\newblock In \emph{NeurIPS Datasets and Benchmark}, 2024{\natexlab{a}}.

\bibitem[Wu et~al.(2024{\natexlab{b}})Wu, Qiu, Ross, Aky{\"u}rek, Chen, Wang, Kim, Andreas, and Kim]{wu2024reasoning}
Wu, Z., Qiu, L., Ross, A., Aky{\"u}rek, E., Chen, B., Wang, B., Kim, N., Andreas, J., and Kim, Y.
\newblock Reasoning or reciting? exploring the capabilities and limitations of language models through counterfactual tasks.
\newblock In \emph{Proceedings of the 2024 Conference of the North American Chapter of the Association for Computational Linguistics: Human Language Technologies (Volume 1: Long Papers)}, pp.\  1819--1862, 2024{\natexlab{b}}.

\bibitem[Xu et~al.(2024)Xu, Wang, Fan, and Liu]{xu2024benchmarking}
Xu, R., Wang, Z., Fan, R.-Z., and Liu, P.
\newblock Benchmarking benchmark leakage in large language models.
\newblock \emph{arXiv preprint arXiv:2404.18824}, 2024.

\bibitem[Yang et~al.(2023)Yang, Chiang, Zheng, Gonzalez, and Stoica]{yang2023rethinking}
Yang, S., Chiang, W.-L., Zheng, L., Gonzalez, J.~E., and Stoica, I.
\newblock Rethinking benchmark and contamination for language models with rephrased samples.
\newblock \emph{arXiv preprint arXiv:2311.04850}, 2023.

\bibitem[Yao et~al.(2024)Yao, Zhuang, Sun, Xu, Kumar, and Shang]{yao2024data}
Yao, F., Zhuang, Y., Sun, Z., Xu, S., Kumar, A., and Shang, J.
\newblock Data contamination can cross language barriers.
\newblock \emph{arXiv preprint arXiv:2406.13236}, 2024.

\bibitem[Ye et~al.(2024)Ye, Xu, Li, and Allen-Zhu]{ye2024physics}
Ye, T., Xu, Z., Li, Y., and Allen-Zhu, Z.
\newblock Physics of language models: Part 2.1, grade-school math and the hidden reasoning process.
\newblock \emph{arXiv preprint arXiv:2407.20311}, 2024.

\bibitem[Zelikman et~al.(2022)Zelikman, Wu, Mu, and Goodman]{zelikman2022star}
Zelikman, E., Wu, Y., Mu, J., and Goodman, N.
\newblock Star: Bootstrapping reasoning with reasoning.
\newblock \emph{NeurIPS}, 35:\penalty0 15476--15488, 2022.

\bibitem[Zeng et~al.(2023)Zeng, Chen, Liu, Jiang, and Jia]{zeng2024mrgsm8kmetareasoningbenchmarklarge}
Zeng, Z., Chen, P., Liu, S., Jiang, H., and Jia, J.
\newblock {MR-GSM8K:} a meta-reasoning benchmark for large language model evaluation.
\newblock \emph{arXiv preprint arXiv:2312.17080}, 2023.

\bibitem[Zhang et~al.(2024)Zhang, Da, Lee, Robinson, Wu, Song, Zhao, Raja, Slack, Lyu, et~al.]{zhang2024careful}
Zhang, H., Da, J., Lee, D., Robinson, V., Wu, C., Song, W., Zhao, T., Raja, P., Slack, D., Lyu, Q., et~al.
\newblock A careful examination of large language model performance on grade school arithmetic.
\newblock \emph{arXiv preprint arXiv:2405.00332}, 2024.

\bibitem[Zhao et~al.(2021)Zhao, Wallace, Feng, Klein, and Singh]{Zhao2021CalibrateBU}
Zhao, T., Wallace, E., Feng, S., Klein, D., and Singh, S.
\newblock Calibrate before use: Improving few-shot performance of language models.
\newblock In \emph{ICML}, 2021.

\bibitem[Zhou et~al.(2024)Zhou, Wang, Xu, Chen, and Duan]{zhou-etal-2024-revisiting}
Zhou, W., Wang, Q., Xu, M., Chen, M., and Duan, X.
\newblock Revisiting the self-consistency challenges in multi-choice question formats for large language model evaluation.
\newblock In \emph{LREC-COLING}, 2024.

\bibitem[Zhu et~al.(2024)Zhu, Chen, Wang, Gong, Yang, and Xie]{zhu2023dyval}
Zhu, K., Chen, J., Wang, J., Gong, N.~Z., Yang, D., and Xie, X.
\newblock Dyval: Graph-informed dynamic evaluation of large language models.
\newblock In \emph{ICLR}, 2024.

\bibitem[Zong et~al.(2024)Zong, Yu, Zhao, Chavhan, and Hospedales]{zong2023fool}
Zong, Y., Yu, T., Zhao, B., Chavhan, R., and Hospedales, T.
\newblock Fool your (vision and) language model with embarrassingly simple permutations.
\newblock \emph{ICML}, 2024.

\end{thebibliography}

\newpage
\appendix
\onecolumn

\startcontents[appendix]
\printcontents[appendix]{ }{0}{\section*{Appendix}}

\section{Discussion and Future Work}
\label{app:future_work}
Our results reveal intricate phenomena of the interplay between reasoning and memorization, but challenging questions remain open: (i) While a model's reasoning capabilities improve during fine-tuning as it memorizes more training puzzles, it is unclear exactly how those capabilities develop, especially when fine-tuned on only question-answer pairs without detailed reasoning steps. (ii) While the models' reasoning capabilities can be significantly improved after fine-tuning, they have not reached 100\% test accuracy yet. Is it because the models only learned some ``shortcut rules'' that can only solve a specific subset of puzzles? If so, what are the shortcuts? (iii) Since some model-based indicators can approximately predict when the model is solving a specific puzzle by memorization vs by reasoning, can we further design intervention mechanisms to bias the model towards reasoning during inference or training time?
Exploring the open questions in further research would deepen our understanding of this space.

\section{Extended Related Work}
\label{sec:extend_related}

\textbf{Memorization in LLMs.}
Prior work has explored training data memorization in LLMs, primarily in the contexts of \textit{privacy} and \textit{copyright} concerns~\citep{carlini2021extracting,lukas2023analyzing, he2024fantastic}, 
focusing on how LLMs may reproduce text near-verbatim to their training data~\citep{lee2022deduplicating,carlini22quantifying, biderman2024emergent}. \revise{\citet{prashanth2024recite}  further introduces a taxonomy for memorization, categorizing it into Recitation, Reconstruction, and Recollection. They investigate the memorization behaviors of the Pythia model~\citep{biderman2024emergent} on the Pile dataset~\citep{gao2020pile}.}
In contrast, we examine memorization in the \textit{reasoning} context,
and focus on analyzing whether LLMs can accurately solve problems encountered during training but struggle to solve slightly perturbed variants. This allows us to better investigate the extent to which LLMs truly understand and generalize the \textit{underlying principles} of the reasoning problems they have been trained on, as opposed to merely memorizing the \textit{text}.

Recent research discusses signs of LLMs memorization in reasoning tasks by evaluating them on counterfactual reasoning tasks. These counterfactual tasks demand similar abstract reasoning skills as the original tasks but are less common in the training data. For instance, tasks such as reversing a sequence of words \citep{mccoy2024embers} show better performance on high-probability sequences than on low-probability sequences; shifting each letter by $n$ places in the alphabet (Rot-$n$) \citep{prabhakar2024deciphering, mccoy2024embers} demonstrates higher performance when $n=13$ than for other values, likely because ``Rot-13'' is commonly used in online forums. ~\cite{wu2024reasoning} presents 11 counterfactual tasks (e.g., 1-indexing in Python, base-9 arithmetic) that show significant performance declines. 
 \cite{jiang2024peek} changes some tokens in the reasoning task descriptions which leads to significant performance drops, suggesting that models might depend on recognizing superficial patterns with strong token bias. Moreover, \cite{razeghi2022impact} finds a strong correlation between the accuracy for a number on numerical reasoning tasks and its frequency in pretraining for GPT-J/GPT-Neo.
In our study, we formally define a memorization score to quantify performance variance under task perturbations, covering both counterfactual alterations (e.g., switching the roles of knights and knaves) and standard perturbations on language level and problem structure level.

\textbf{Detecting benchmark contamination.}
Recent work has shown that LLMs' performance drastically declines when faced with altered versions of popular reasoning benchmarks, suggesting potential contamination/memorization of these benchmarks. 
The benchmark variants include diverse forms such as altered multiple-choice questions formats~\citep{wang2024beyond,zong2023fool,gupta2024changing,zhou-etal-2024-revisiting,robinson2023leveraging}, rephrased or translated problems~\citep{xu2024benchmarking,yang2023rethinking,yao2024data}, shuffled example orderings~\citep{oren2023proving}, human-curated problems of comparable difficulty~\citep{zhang2024careful}, functional variants generating random instantiations~\citep{srivastava2024functional,mirzadeh2024gsm}, and problems beyond specific date cutoffs~\citep{roberts2023cutoff,jain2024livecodebench}.
Previous work either focus on surface level language perturbations or require extensive expert-level annotations for math level variations. In contrast, our benchmark support automatic problem-level perturbation, solution and reasoning procedure generation, and easily scale to different difficult levels and dataset sizes without extra human efforts.

\textbf{Logical reasoning benchmarks.}
To evaluate logical reasoning capabilities in LLMs, synthetic benchmarks have been developed to enable scalable generation of samples with varying configurations and difficulty levels~\citep{Clark2020TransformersAS, giadikiaroglou2024puzzle, Parmar2024LogicBenchTS}.
For instance,  DyVal~\citep{zhu2023dyval} uses directed acyclic graphs to dynamically generate samples on reasoning tasks including deductive, Boolean, and abductive reasoning. 
\cite{chenpremise} focus on propositional logical problems involving definite clauses, and synthetically generate variations with different premise orders, such as forward, backward, and shuffled. 
\cite{dziri2024faith} explore the limitations of LLMs in tasks requiring compositional reasoning, including multiplication, logic grid puzzles, and dynamic programming problems. ZebraLogic~\citep{zebralogic2024} is an extended benchmark that systematically tests logical reasoning capabilities. 
BoardgameQA~\citep{kazemi2024boardgameqa} presents a question-answering dataset characterized by contradictory facts and rules in the questions. 
\revise{PRONTOQA~\citep{saparov2023language} is a synthetic question-answering dataset where each example is generated from a synthetic world model represented in first-order logic. This dataset enables parsing the generated chain of thought into symbolic proofs, facilitating formal analysis.}
TruthQuest~\citep {mondorf2024liar} is the most similar task to our work, which provides evaluation samples based on \kk-type of puzzles involving 3-6 person. Our work provides more comprehensive \emph{dynamic} set of \kk puzzles that support automatic generation of perturbations, solutions and detailed reasoning steps. Moreover, based on this benchmark, we define and measure memorization in reasoning tasks, revealing intricate interplay between memorization and reasoning in LLMs.

\textbf{Improving reasoning via fine-tuning.}
Prior work has explored fine-tuning LLMs on synthetic reasoning data to enhance their performance on reasoning. 
DyVal~\citep{zhu2023dyval} shows that fine-tuning Llama2-13B-chat on their synthetic reasoning benchmark improves its performance on other popular reasoning benchmarks. 
BoardgameQA~\citep{kazemi2024boardgameqa} find that fine-tuning BERT-large and T5-XXL on their training dataset with synthetic proofs outperforms few-shot CoT prompting using PaLM. 
\cite{ye2024physics} pretrain GPT2 from scratch on synthetic math problems, synthetic CoT steps and solutions and show that model can solve problems from the same distribution and generalize to out-of-distribution (OOD) problems.  
However, \cite{dziri2024faith} show that while GPT-3 fine-tuned on their compositional reasoning tasks  with/without reasoning steps can solve in-distribution (ID) problems, it fails to generalize to OOD tasks with increased problem sizes.
Besides using synthetic CoTs, there are work using model-generated CoTs to enhance the models' reasoning capabilities~\citep{chung2024scaling}. STaR~\citep{zelikman2022star} uses model self-generated CoTs on correctly solved samples to iteratively fine-tune itself as a self-taught reasoner.  A number of work~\citep{puerto2024fine,kim2023cot,ho2022large,hsieh2023distilling} leverage CoTs generated from teacher models to train smaller student models.
Additionally, some recent efforts have focused on leveraging intermediate reasoning steps in CoT more implicitly. For instance, \citet{deng2023implicit} distill intermediate reasoning tokens into the network layers by representing reasoning steps as vectors and using them as targets; \citet{deng2024explicit} distill CoT by gradually removing the intermediate steps and fine-tuning the model to internalize these steps, predicting the answers based on partial CoT. Both studies show that full CoT fine-tuning may not be necessary for the model to achieve strong reasoning performance.

In our study, we employ both direct fine-tuning and CoT fine-tuning to achieve memorization on \kk training data. Notably, our findings show that the fine-tuned \gptfouromini and \llamathree models can effectively generalize to unseen OOD and ID \kk problems, contributing new insights to the topic of LLM fine-tuning for reasoning.

\revise{Orthogonal to our work, inference-time techniques have been explored to enhance reasoning performance such as self-consistency~\citep{wangself2023}, self-verification~\citep{weng2023large}, and integration with external symbolic solvers~\citep{pan2023logic}.}

\revise{\textbf{Grokking}. Our findings are related to Grokking, first identified by ~\cite{power2022grokking} on a small algorithmic dataset, where validation accuracy suddenly improves from random chance to near-perfect generalization long after severe overfitting. Follow-up studies expanded the range of tasks where grokking occurs and proposed various explanations ~\citep{liu2022towards,murty2023grokking,liu2022omnigrok}. Recently, ~\cite{wang2024grokking} observed grokking in the domain of complex knowledge-based tasks, showing that implicit reasoning over parametric knowledge emerges only after extensive overfitting.
In this work, we observe a related phenomenon but through the lens of memorization and logical reasoning. Through novel (math \& language-level) perturbation tests and transferability analyses, we verify that LLM's reasoning skills emerge alongside memorization.  Furthermore, our investigation focuses on logical reasoning, offering new insights into how LLMs acquire logical reasoning skills.
}

\revise{\textbf{Adversarial robustness under perturbations.} Language-level perturbations have been widely used to assess the adversarial robustness of language models, often involving manually annotated attacks, as seen in advGLUE~\cite{wang2021adversarial}, ANLI~\cite{nie2020adversarial}, and TextFooler~\cite{jin2019bert}. However, these approaches fundamentally differ from our proposed mathematical-level perturbations in purpose, methodology, and scope. Specifically, prior studies primarily focus on natural language understanding tasks, such as sentiment analysis and textual entailment, aiming to generate adversarial perturbations that cause misclassification without altering the ground truth.
In contrast, our proposed perturbation method operates at a mathematical level in logical reasoning tasks and modifies \textit{not only the problem, but also the ground-truth answer}. These mathematical perturbations ensure that the perturbed puzzle has a \textit{distinctly different solution} compared to the original puzzle, while remaining superficially similar and maintaining a comparable difficulty level. This is guaranteed by the Perturber, Reasoner, and Solver components in our data generation framework. This approach provides a direct evaluation of the models’ understanding of the underlying mathematical principles.
By addressing logical reasoning robustness through mathematical-level perturbations, our work contributes a novel perspective distinct from prior studies.
}

\section{Details on \kk Benchmark}
\label{app:dataset_detail}

\subsection{The Abstract Representation}
\label{app:kk-abstract-repr}

We use a simple internal representation using basic Python primitives (integer, string and tuple) to encode each \kk puzzle. This allows easy inter-operation with the json format to simplify saving and loading. Specifically, for a $N$-people puzzle, each person is represented by the integer $0,\ldots,N-1$. Each person's statement is represented by a tuple \texttt{(type, arguments, ...)}, where \texttt{type} indicate the statement type listed below:

\begin{itemize}
    \item \textbf{Leaf Statements}: It can be either \texttt{('lying', i)} or \texttt{('telling-truth', i)}, where \texttt{i} is an integer and this statement assert the \texttt{i}th person is lying or truthful.
    \item \textbf{Composite Statements}: It can take one or more statements as arguments, and has the following types:
    \begin{itemize}
        \item Negation \texttt{('not', statement)}
        \item Conjunction \texttt{('and', statement1, statement2, ...)}
        \item Disjunction \texttt{('or', statement1, statement2, ...)}
        \item Implication \texttt{('->', statement1, statement2)}
        \item Equivalence \texttt{('<=>', statement1, statement2)}
    \end{itemize}
\end{itemize}

\subsection{The Abstract Puzzle Module: Generator}
\label{app:kk-generator}

The Generator samples a problem based on a random seed and a difficulty level specification $(N, W, D)$, where $N$ indicates the number of people, $W$ indicates the max width of each statement, $D$ indicates the max depth of each person's statement. To instantiate the problem, we initialize a random number generator, and sample a statement for each person sequentially. We sample each statement type uniformly at random. For composite statement with variable number of sub-statements, we also randomize the number according to the max width $W$. We restrict the sampling to only leaf statements if the max depth is exhausted. We avoid (skip and resample) some invalid (e.g., asserting self is lying) or uninteresting cases (e.g., a \texttt{and} statement with identical sub-statements).

The following is an example \kk puzzle with 5 people in the abstract representation. We will use this example to illustrate various component in the rest of the section.

\begin{promptbox}[Example puzzle of 5 people in the abstract representation]
\footnotesize
\begin{verbatim}
(('and', ('lying', 3), ('telling-truth', 4)),
 ('<=>', ('lying', 3), ('telling-truth', 4)),
 ('telling-truth', 4),
 ('telling-truth', 0),
 ('<=>', ('telling-truth', 2), ('lying', 2)))
\end{verbatim}
\end{promptbox}

\subsection{The Abstract Puzzle Module: Solver and Reasoner}
\label{app:kk-solver-reasoner}

Each \kk problem can be transformed and solved as a Boolean satisfiability problem. Specifically, consider a puzzle involving $N$ people, a possible solution assign a Boolean value to $N$ variables $B_1,B_2,\ldots,B_N$, where the truth value of $B_i$ indicates whether the $i$th person is telling the truth. By definition, the $i$th person is telling the truth if and only if their statement $S_i$ is true. Therefore, a valid solution to a \kk puzzle is a Boolean assignment for $B_1,B_2,\ldots,B_N$ such that the following formula evaluates to true.

\begin{equation}
(B_1\Leftrightarrow S_1)\wedge(B_2\Leftrightarrow S_2)\wedge\cdots\wedge(B_N\Leftrightarrow S_N).
\label{eq:kk-sat}
\end{equation}

We implement our Solver and Reasoner based on this reduction. We take two different approaches here, because we want to find \emph{all} possible solutions in the Solver, and we want to generate \emph{intuitive} intermediate steps for the Reasoner. 

Specifically, we are primarily interested in evaluating \kk puzzles with a unique valid solution. Therefore, we design our Solver to use a simple brute-force search that enumerates all possible Boolean  assignments for $N$ people and count the number of assignments that evaluate  \Cref{eq:kk-sat} to true. In our dataset construction, we only include puzzles whose solution count is exactly one.

In the Reasoner, we are interested in procedurally generating intermediate reasoning steps that lead to the final solution. We note that when explaining the reasoning steps for \kk puzzles, human or off-the-shelf LLMs rarely use the brute-force assignment search approach adopted in our Solver. Instead, they tend to examine the statement from each person sequentially, construct a \emph{partial} assignment for the people examined so far, and backtrack when a contradiction is found. We design our Reasoner following the same procedure.

Specifically, we maintain a queue of people to be examined next, and a partial assignment of knight / knave for people that have been examined so far. In each step, we examine the next person from the queue by adding to the partial assignment the assumed knight / knave role for this person. Given the newly proposed assignment, we go through the known statements and check if there is a contradiction. (A) If a contradiction is found, we record the statement of contradiction as the explanation, and start backtracking. Backtracking will put people back into the to-be-examined queue until we reach a person who has an alternative unexamined role assignment. If no such person is found during backtracking, this means there is no valid solution for this problem. (B) If a contradiction is not found, we can proceed to examine the next person in the queue. Here we also implement a mechanism to reorder the queue so that it may match the human behavior better. For example, if the current person's statement is ``If Noah is a knight, then Lily is a knave.'' then we would bring Noah and Lily to the front of the to-be-examined queue, provided that they are in the queue (i.e.,  have not been previously examined).

The reasoning steps are generated and stored using a similar format as the abstract representation of the puzzle as described in \Cref{app:kk-abstract-repr}. The following snippet shows an example of the generated reasoning steps for the example puzzle shown above:

\begin{promptbox}[Example of generated reasoning steps in the abstract representation]
\scriptsize
\begin{verbatim}
[('proposal', {'assignment': True, 'outcome': 'ok', 'person': 0}),
 ('proposal', {'assignment': True, 'conflict_statement': (0, True), 'outcome': 'conflict', 'person': 3}),
 ('proposal', {'assignment': False, 'conflict_statement': (3, False), 'outcome': 'conflict', 'person': 3}),
 ('reconsider', {'exhausted': [3], 'person': 0}),
 ('proposal', {'assignment': False, 'outcome': 'ok', 'person': 0}),
 ('proposal', {'assignment': True, 'conflict_statement': (3, True), 'outcome': 'conflict', 'person': 3}),
 ('proposal', {'assignment': False, 'outcome': 'ok', 'person': 3}),
 ('proposal', {'assignment': True, 'conflict_statement': (0, False), 'outcome': 'conflict', 'person': 4}),
 ('proposal', {'assignment': False, 'outcome': 'ok', 'person': 4}),
 ('proposal', {'assignment': True, 'conflict_statement': (2, True), 'outcome': 'conflict', 'person': 2}),
 ('proposal', {'assignment': False, 'outcome': 'ok', 'person': 2}),
 ('proposal', {'assignment': True, 'conflict_statement': (1, True), 'outcome': 'conflict', 'person': 1}),
 ('proposal', {'assignment': False, 'outcome': 'ok', 'person': 1}),
 ('success', {'assignments': (False, False, False, False, False)})]
\end{verbatim}
\end{promptbox}

\subsection{The Abstract Puzzle Module: Perturber}
\label{app:kk-perturber}

To support memorization measurement, the \kk Puzzle Perturber generate an perturbed version of a given puzzle. We design the perturbation with the following considerations:

\begin{itemize}
    \item The perturbation should be ``local'', in the sense that the perturbed problem should be similar to the original problem when measured in some superficial ways, such as edit distance.
    \item The perturbation should keep the nature of the problem, i.e., the perturbed problem should be solvable using the same underlying principle, and the difficulty level should be roughly maintained. 
    \item The perturbed puzzle should have a unique solution, which should be different from the solution of the original puzzle.
\end{itemize}

With those consideration, we support two different perturbations:

\begin{itemize}
    \item Statement perturbation: randomly choose a person and sample a new statement tree for that person.
    \item Leaf perturbation: randomly choose a person, and from that person's existing statement tree, randomly choose one leaf node and change it.
\end{itemize}

\subsection{The Natural Language Module: NL-Generator}
\label{app:kk-nl-generator}

The NL-Generator generate a \kk puzzle in natural language by converting a given abstract-form puzzle into the familiar text form described in natural language. For example, the puzzle presented (in the abstract representation) above can be materialized as follows:

\begin{promptbox}[Example puzzle converted to natural language representation]
\footnotesize
A very special island is inhabited only by knights and knaves. Knights always tell the truth, and knaves always lie. You meet 5 inhabitants: David, Zoey, Alexander, Aurora, and Isabella. In David's words: "Aurora is a knave and Isabella is a knight". Zoey expressed that Aurora is a knave if and only if Isabella is a knight. Alexander said, "Isabella is a knight." Aurora commented, "David is a knight". According to Isabella, "Alexander is a knight if and only if Alexander is a knave". So who is a knight and who is a knave?
\end{promptbox}

Specifically, given a puzzle of $N$ people in the abstract representation, our natural language representation generator first sample $N$ human names, and then format each people's claim by plugging in the corresponding name mapping. We use some heuristics to make the conversion of the tree-structured logic statements to natural language sounds natural. Our current implementation randomly sample from 18 templates of making a statement claim and from the following common names --- but this can be easily extended to include more.

{\footnotesize
\begin{verbatim}
COMMON_NAMES = ['Emma', 'Liam', 'Olivia', 'Noah', 'Ava', 'Ethan', 'Sophia',
                'Mason', 'Isabella', 'William', 'Mia', 'James', 'Charlotte',
                'Benjamin', 'Amelia', 'Lucas', 'Harper', 'Henry', 'Evelyn',
                'Alexander', 'Abigail', 'Michael', 'Emily', 'Daniel', 'Elizabeth',
                'Jacob', 'Sofia', 'Logan', 'Avery', 'Jackson', 'Ella', 'Sebastian',
                'Scarlett', 'Jack', 'Grace', 'Aiden', 'Chloe', 'Owen', 'Victoria',
                'Samuel', 'Riley', 'Matthew', 'Aria', 'Joseph', 'Lily', 'Luke',
                'Aurora', 'David', 'Zoey', 'Oliver', 'Penelope']
\end{verbatim}
}

\subsection{The Natural Language Module: NL-Reasoner}
\label{app:kk-nl-reasoner}

The NL-Reasoner generates detailed reasoning steps in natural language by converting the output from the abstract Reasoner to natural language descriptions using a similar approach as the NL-Generator. The following show the generated reasoning steps in natural language for the puzzle shown above:

\begin{promptbox}[Reasoning steps generated by the Reasoner]\footnotesize
Let's think step by step, by considering whether each person is lying and if that leads to contradiction.
\begin{enumerate}[leftmargin=20pt, itemsep=0pt, topsep=0pt, partopsep=0pt]
\item Assume David is a knight. No contradiction is found in their claim that Aurora is a knave and Isabella is a knight.
\item Aurora cannot be a knight, because this would contradict the claim of David that Aurora is a knave and Isabella is a knight.
\item Aurora cannot be a knave, because this would contradict the false claim of their own that David is a knight.
\item We have exhausted all possibilities for Aurora, so let us go back and reconsider David.
\item Assume David is a knave. No contradiction is found in their false claim that Aurora is a knave and Isabella is a knight.
\item Aurora cannot be a knight, because this would contradict the claim of their own that David is a knight.
\item Assume Aurora is a knave. No contradiction is found in their false claim that David is a knight.
\item Isabella cannot be a knight, because this would contradict the false claim of David that Aurora is a knave and Isabella is a knight.
\item Assume Isabella is a knave. No contradiction is found in their false claim that Alexander is a knight if and only if Alexander is a knave.
\item Alexander cannot be a knight, because this would contradict the claim of their own that Isabella is a knight.
\item Assume Alexander is a knave. No contradiction is found in their false claim that Isabella is a knight.
\item Zoey cannot be a knight, because this would contradict the claim of their own that Aurora is a knave if and only if Isabella is a knight.
\item Assume Zoey is a knave. No contradiction is found in their false claim that Aurora is a knave if and only if Isabella is a knight.
\end{enumerate}
This leads to a feasible solution.
\end{promptbox}

\subsection{The Natural Language Module: NL-Perturber}
\label{app:kk-nl-perturber}

The NL-Perturber generates perturbed puzzles at the language level. Note unlike in the perturbations generated by the abstract Perturber, NL-Perturber keep the underlying abstract puzzle intact and only modify the materialization in natural language. Therefore, the solution to the perturbed puzzle is identical to the solution to the original puzzle. Specifically, the NL-Perturber supports the following perturbations:

With those consideration in mind, we provide two family of perturbations:
\begin{itemize}
    \item Uncommon name: replace the names of the characters with randomly sampled names from the set of uncommon names.
    \item Random role: change the role name from knight/knave to other pairs of role names. To avoid introducing bias, we sample from pairs of good/bad role names, including ``\textit{saint/sinner, hero/villain, angel/devil, altruist/egoist, sage/fool, pioneer/laggard}''.
    \item Reorder statement: shuffle the order of presenting each person's statement.
    \item Flip role: change the role from knight/knave to knave/knight, i.e.,  knave will be telling the truth while knight will be lying.
\end{itemize}

The uncommon names are sampled from the following list:

{\footnotesize
\begin{verbatim}
UNCOMMON_NAMES = [
  'Zephyr', 'Elowen', 'Caspian', 'Isolde', 'Osiris', 'Vesper', 'Thaddeus', 'Ondine',
  'Lysander', 'Xanthe', 'Oberon', 'Calliope', 'Leander', 'Eulalia', 'Florian', 'Forsythe',
  'Nephele', 'Peregrine', 'Ianthe', 'Lazarus', 'Elodie', 'Cillian', 'Ottoline', 'Evander',
  'Saffron', 'Caius', 'Zora', 'Cyprian', 'Amaryllis', 'Theron', 'Perdita', 'Ignatius',
  'Zephyrine', 'Balthazar', 'Melisande', 'Zinnia', 'Sylvester', 'Cosima', 'Leocadio',
  'Percival', 'Oceane', 'Evanthe', 'Zenobia', 'Eurydice', 'Quillan', 'Aeronwen',
  'Thorsten', 'Xiomara', 'Zephyrus', 'Ysolde'
]
\end{verbatim}
}

Note the flip role perturbation is somewhat adversarial as it goes against the common intuition that good role tends to tell the truth while bad role tends to lie. We indeed observe that the models would make a lot of mistakes under this perturbation, despite that the perturbed problem is perfect valid and unambiguous. However, the study of how model's bias impact its reasoning capability is not the main focus of this paper. So we keep this perturbation as reference but primarily focus on ``benign'' perturbations.

\subsection{Dataset Generation}
\label{app:data_gen_details}

\paragraph{\kk dataset}

During our data construction, we use the maximum width $W=2$ and depth $D=2$, and the number of persons in the puzzle $N=2,3,4,5,6,7,8$. 

\begin{figure}[!htb]
    \centering
    \begin{minipage}{0.45\textwidth}
         \centering
         \includegraphics[width=\linewidth]{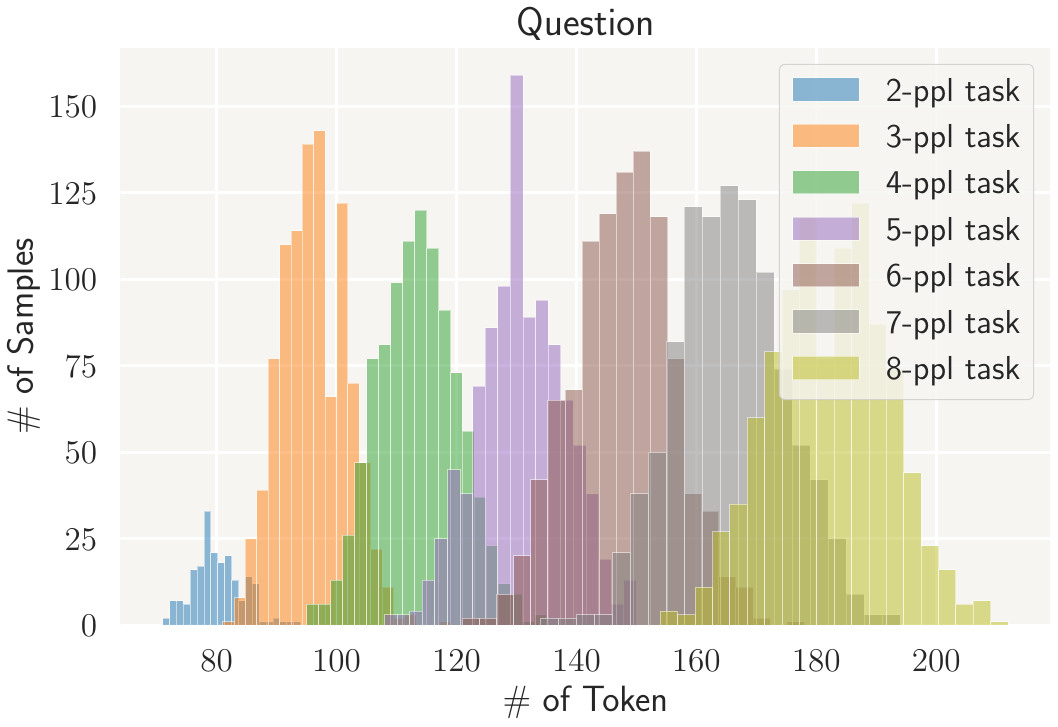}
    \caption{ \kk questions}
     \end{minipage}
     \begin{minipage}{0.45\textwidth}
         \centering
         \includegraphics[width=\linewidth]{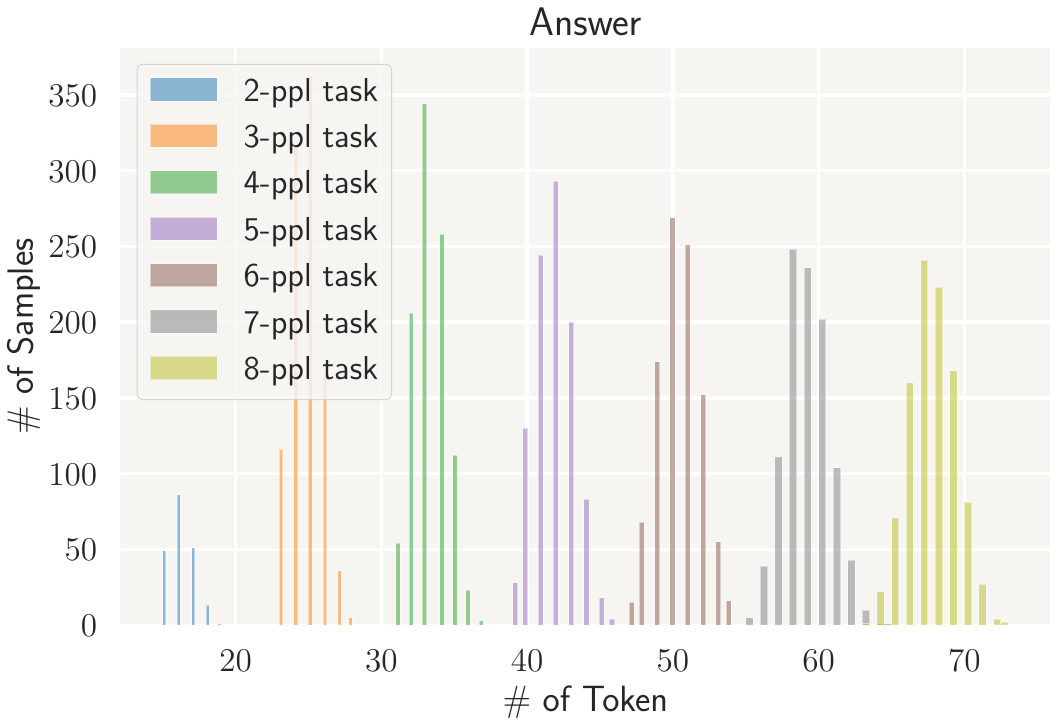}
    \caption{ \kk answers}
     \end{minipage}
      \begin{minipage}{0.31\textwidth}
         \centering
         \includegraphics[width=\linewidth]{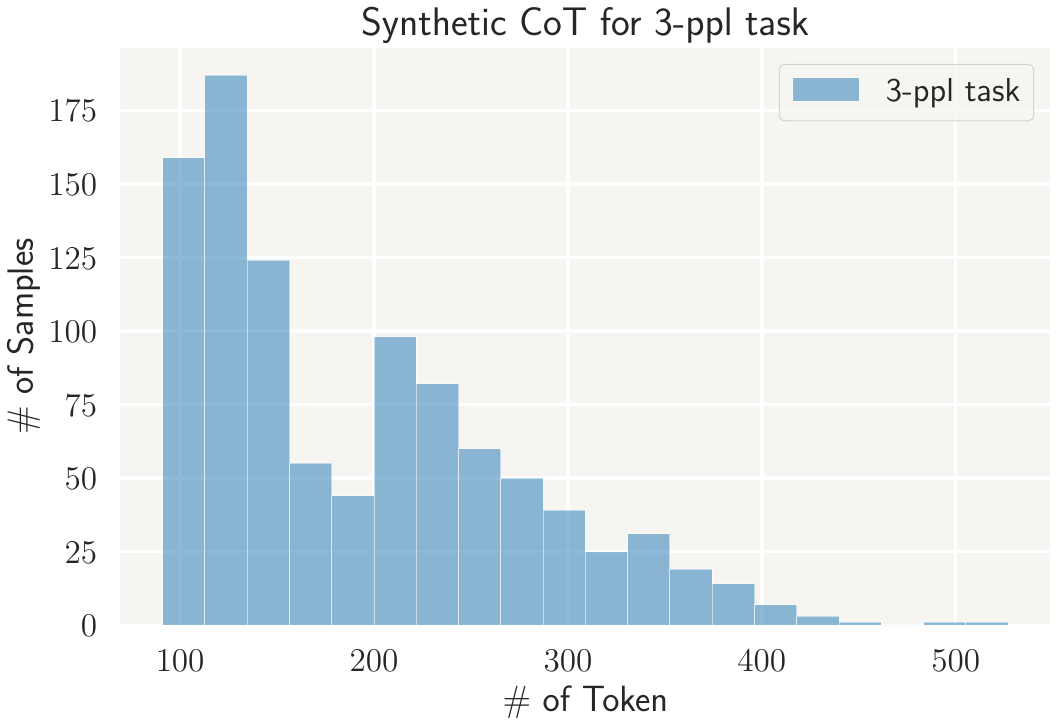}
    \caption{ 3-people \kk synthetic CoTs}
     \end{minipage}
      \begin{minipage}{0.31\textwidth}
         \centering
         \includegraphics[width=\linewidth]{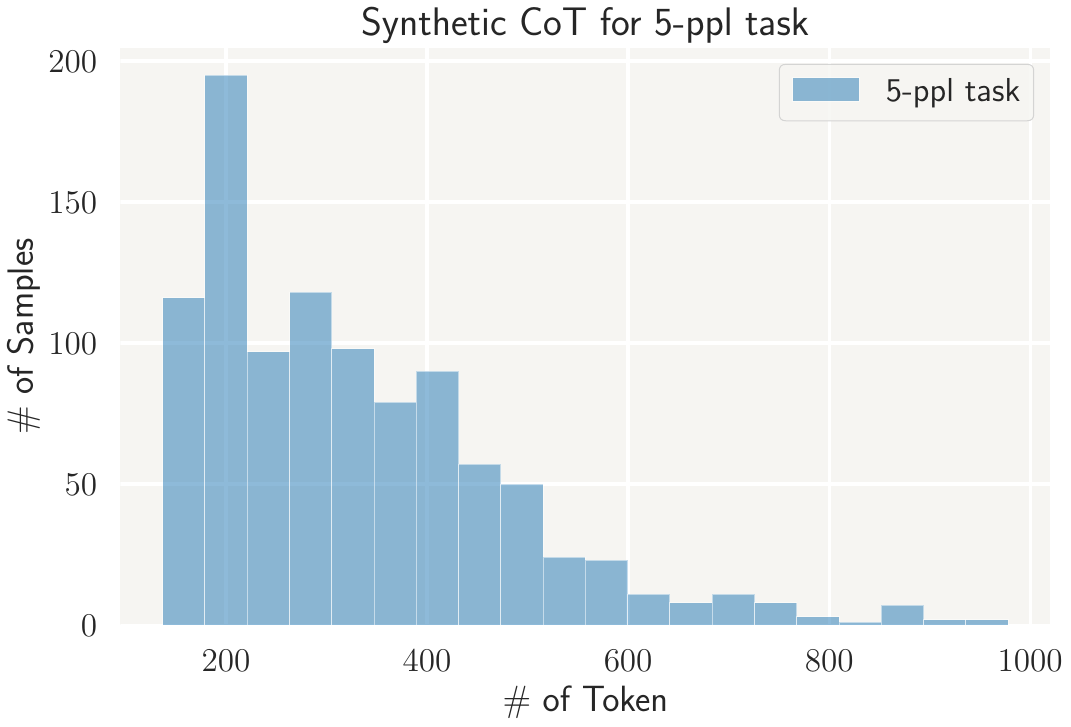}
     \caption{ 5-people \kk synthetic CoTs}
     \end{minipage}
     \begin{minipage}{0.31\textwidth}
         \centering
         \includegraphics[width=\linewidth]{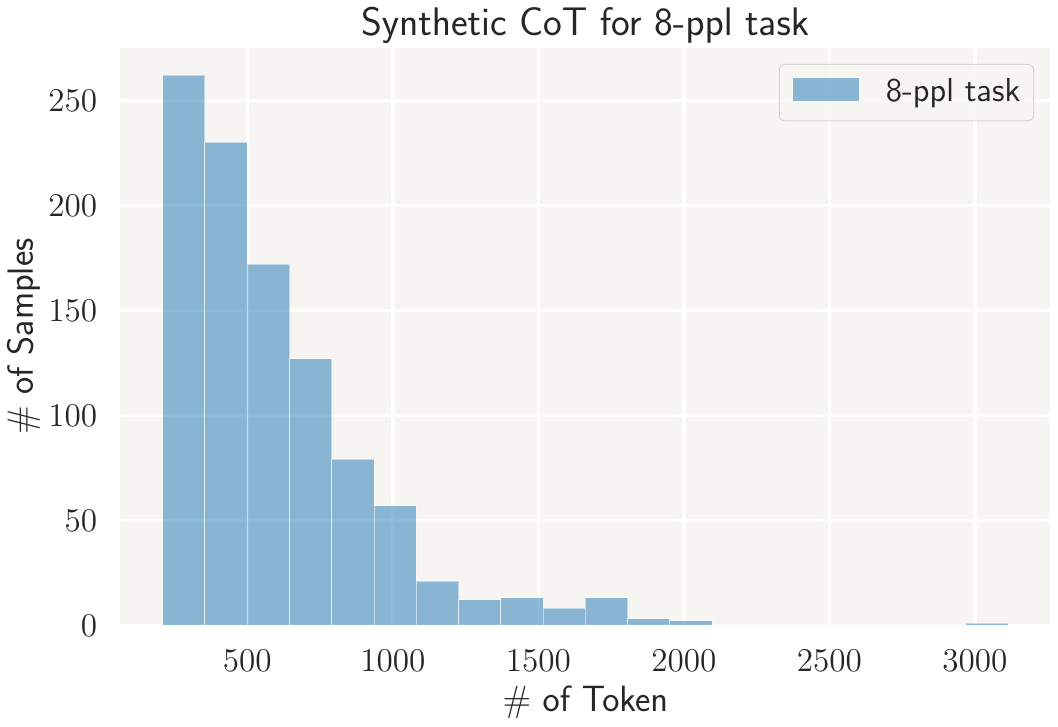}
     \caption{ 8-people \kk synthetic CoTs}
     \end{minipage}
    \caption{Length distributions of \kk training data.}
    \label{fig:train-data-stat}
\end{figure}

We present the length distributions of \kk training dataset in \cref{fig:train-data-stat}. The length distributions of the test dataset are similar to those of the training dataset.

\begin{figure}
    \centering
    \includegraphics[width=0.8\linewidth]{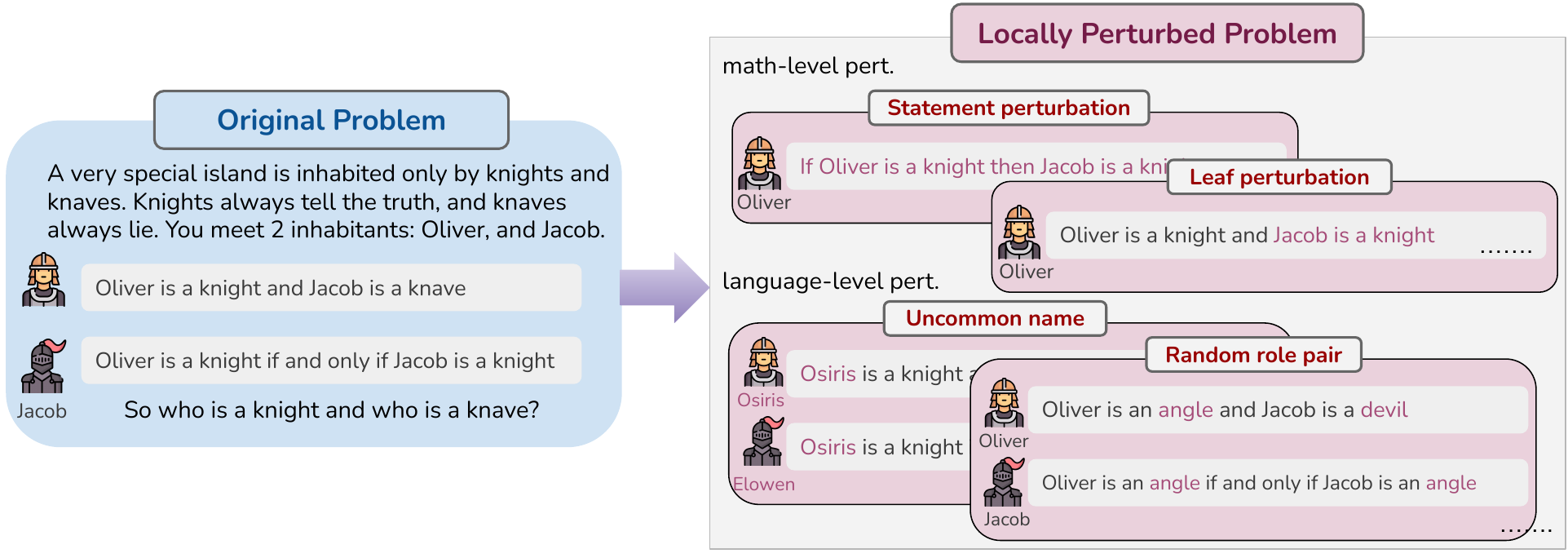}
    \caption{Comparison between different locally perturbed problems.}
    \label{fig:illustration_problem_perturb}
\end{figure}
\paragraph{Local perturbation}
\cref{tab:example_2ppl_puzzle} presents the example knight (truth-teller) and knave (liar) scenario involving two people: Liam and Aria, with corresponding logical statements, and converted English statements, questions, and answers. 
It also shows three versions of the problems: an original example, a leaf-perturbed version, and a statement-perturbed version. Specifically, (1) leaf perturbation changes a ``leaf'' in the logical tree - a single truth value. In this case, it flipped Jacob's status in Oliver's statement from knave (liar) to knight (truth-teller)
(2) Statement perturbation changes the entire structure of a statement. Here, it changed Oliver's statement entirely. 
Both perturbations result in changing the answer.  The leaf perturbation creates a subtle change in one statement that flips the logical outcome, while the statement perturbation changes the entire one statement. 

\begin{table}[!htb]
\centering
\vspace{-2mm}
\caption{2-person puzzle generation with the knight (telling-truth) and knave (lying) and comparison between original sample, leaf-perturbed sample, and statement-perturbed sample.
}
\label{tab:example_2ppl_puzzle}
\resizebox{\linewidth}{!}{%
\begin{tabular}{>{\centering\arraybackslash}m{3cm}|>{\centering\arraybackslash}m{5cm}|>{\centering\arraybackslash}m{5cm}|>{\centering\arraybackslash}m{5cm}}
\toprule
\textbf{Type} & \textbf{Example} & \textbf{Leaf Perturbed Example} & \textbf{Statement Perturbed Example} \\
\midrule
person &  \multicolumn{3}{c}{Oliver (person index 0), Jacob (person index 1)} \\
\midrule
logical statement 
& Oliver: ('and', ('telling-truth', 0), ('lying', 1)) \newline
  Jacob: (`$\Leftrightarrow$', ('telling-truth', 0), ('telling-truth', 1))
& Oliver: ('and', ('telling-truth', 0), \textbf{('telling-truth', 1)}) \newline
  Jacob: (`$\Leftrightarrow$', ('telling-truth', 0), ('telling-truth', 1))
& Oliver: \textbf{('$\xrightarrow{}$', ('telling-truth', 0), ('telling-truth', 1))} \newline
  Jacob: (`$\Leftrightarrow$', ('telling-truth', 0), ('telling-truth', 1))
  \\
\midrule
English statement 
& Oliver: Oliver is a knight and Jacob is a knave
\newline 
Jacob: Oliver is a knight if and only if Jacob is a knight
& Oliver: Oliver is a knight and \textbf{Jacob is a knight}
\newline 
Jacob: Oliver is a knight if and only if Jacob is a knight
& Oliver: \textbf{If Oliver is a knight then Jacob is a knight }
Jacob: Oliver is a knight if and only if Jacob is a knight
\\
\midrule
question
& A very special island is inhabited only by knights and knaves. Knights always tell the truth, and knaves always lie. You meet 2 inhabitants: Oliver, and Jacob. Oliver commented, "Oliver is a knight and Jacob is a knave". Jacob remarked, "Oliver is a knight if and only if Jacob is a knight". So who is a knight and who is a knave?
& A very special island is inhabited only by knights and knaves. Knights always tell the truth, and knaves always lie. You meet 2 inhabitants: Oliver, and Jacob. Oliver commented, "Oliver is a knight and Jacob is a knight". Jacob remarked, "Oliver is a knight if and only if Jacob is a knight". So who is a knight and who is a knave?
& A very special island is inhabited only by knights and knaves. Knights always tell the truth, and knaves always lie. You meet 2 inhabitants: Oliver, and Jacob. Oliver commented, "If Oliver is a knight then Jacob is a knight". Jacob remarked, "Oliver is a knight if and only if Jacob is a knight". So who is a knight and who is a knave?	\\
\midrule
answer &
(1) Oliver is a knight \newline
(2) Jacob is a knave 
&
(1) Oliver is a knight \newline
(2) \textbf{Jacob is a knight}
&
(1) Oliver is a knight \newline
(2) \textbf{Jacob is a knight} \\
\bottomrule
\end{tabular}
}
\end{table}

Moreover, we compare the math-level perturbation with language-level perturbation in \cref{fig:illustration_problem_perturb}.

As mentioned in \cref{sec:proposal}, the Perturber of the abstract puzzle module generates a perturbed puzzle with a \textit{unique} solution that is \textit{different} from the original puzzle, or until the maximum number of attempts is reached. We set this limit to 2000 attempts.
\begin{itemize}
    \item For statement perturbation, the Perturber can always return a valid perturbed puzzle due to the large perturbation space. 
    \item For leaf perturbation, since the process is restricted to a single leaf node, it may not always find a valid perturbed puzzle within the constraints of unique and different solution. Below are the detailed proportions of valid leaf perturbations on training samples (under 2000 max attempts for each sample):
76\% valid for 2-person task;
93.4\% valid for 3-person task;
95.4\% valid for 4-person task;
98.8\% valid for 5-person task;
99.5\% valid for 6-person task;
100\% valid for 7/8-person tasks.
\end{itemize}

\section{Experimental Setups}
\label{app:exp}

\subsection{Models}
\label{app:models}

\Cref{tab:endpoint} provides the details of the models evaluated in our study.

\begin{table}[ht]
    \centering
    \caption{HuggingFace links or endpoint specifications for evaluated models.}
    \label{tab:endpoint}
    \resizebox{\linewidth}{!}{
    \begin{tabular}{ll}
    \toprule
        {\bf Model} & {\bf Link} \\
    \midrule
        \llamathree & \url{https://huggingface.co/meta-llama/Meta-Llama-3-8B} \\
         Phi-3-mini & \url{https://huggingface.co/microsoft/Phi-3-mini-4k-instruct} \\
         Phi-3-medium & \url{https://huggingface.co/microsoft/Phi-3-medium-4k-instruct} \\
         NuminaMath-7B-CoT & \url{https://huggingface.co/AI-MO/NuminaMath-7B-CoT} \\
         Deepseek-Math-7B & \url{deepseek-ai/deepseek-math-7b-instruct}\\
        \claude & \url{https://www.anthropic.com/news/claude-3-5-sonnet}, \texttt{claude-3-5-sonnet-20240620} endpoint \\
        \gptfouromini  & \url{https://platform.openai.com/docs/models/}, \texttt{gpt-4o-mini-2024-07-18} endpoint \\
        \gptfouro & \url{https://platform.openai.com/docs/models/}, \texttt{gpt-4o-2024-05-13} endpoint \\
        \geminiflashtwo & \url{https://console.cloud.google.com/vertex-ai/model-garden}, \texttt{gemini-1.5-flash-002} endpoint \\
        \geminiprotwo & \url{https://console.cloud.google.com/vertex-ai/model-garden}, \texttt{gemini-1.5-pro-002} endpoint\\
    \bottomrule
    \end{tabular}
    }
\end{table}

\subsection{Experimental Details}
\label{app:exp_details}

\subsubsection{Evaluation}
By default, we utilize zero-shot direct prompting with task-specific instructions for open-ended question-answering. 
We employ the following prompt: 
\begin{promptbox}[0-shot Direct Prompting]
Your task is to solve a logical reasoning problem. You are given set of statements from which you must logically deduce the identity of a set of characters.\\

You must infer the identity of each character. At the end of your answer, you must clearly state the identity of each character by following the format:\\

CONCLUSION:\\
(1) ...\\
(2) ...\\
(3) ...\\

\#\#\# Question: \{question\}\\
\#\#\# Answer:
\end{promptbox}

In addition to the 0-shot direct prompting used in the main paper, we explore 0-shot Chain of Thought (CoT) prompting and 1-shot direct/CoT prompting and report the results in Appendix \cref{app:add_exp_results}.

\begin{promptbox}[0-shot CoT Prompting]
Your task is to solve a logical reasoning problem. You are given set of statements from which you must logically deduce the identity of a set of characters.\\ 

You must infer the identity of each character. First, explain your reasoning. At the end of your answer, you must clearly state the identity of each character by following the format:\\

CONCLUSION:\\
(1) ...\\
(2) ...\\
(3) ...\\

\#\#\# Question: \{question\}\\
\#\#\# Answer: Let's think step by step\\
\end{promptbox}

In addition, we utilize a specific CoT prompting format for instruction-tuned models: DeepSeek-Math-7B and NuminaMath-7B-CoT, as recommended by their developers:
\begin{promptbox}
Please reason step by step, and put your final answer within {\textbackslash}boxed\{\}.
\end{promptbox}
This replaces the previous prompt, "Let's think step by step."

\begin{promptbox}[1-shot Direct Prompting]
Your task is to solve a logical reasoning problem. You are given set of statements from which you must logically deduce the identity of a set of characters.\\

You must infer the identity of each character. At the end of your answer, you must clearly state the identity of each character by following the format:\\

CONCLUSION:\\
(1) ...\\
(2) ...\\
(3) ...\\

\#\#\# Question: A very special island is inhabited only by knights and knaves. Knights always tell the truth, and knaves always lie. You meet 2 inhabitants: Jack, and Sophia. Jack tells you that Sophia is not a knave. Sophia says that If Jack is a knight then Sophia is a knight. So who is a knight and who is a knave?\\
\#\#\# Answer:\\
CONCLUSION:\\
(1) Jack is a knight\\
(2) Sophia is a knight\\
\\

\#\#\# Question: \{question\}\\
\#\#\# Answer:\\
\end{promptbox}

\begin{promptbox}[1-shot CoT Prompting]
Your task is to solve a logical reasoning problem. You are given set of statements from which you must logically deduce the identity of a set of characters.\\

You must infer the identity of each character. First, explain your reasoning. At the end of your answer, you must clearly state the identity of each character by following the format:\\

CONCLUSION:\\
(1) ...\\
(2) ...\\
(3) ...\\

\#\#\# Question: A very special island is inhabited only by knights and knaves. Knights always tell the truth, and knaves always lie. You meet 2 inhabitants: Ella, and Penelope. In a statement by Ella: ``Ella is a knight or Penelope is a knight''. According to Penelope, ``Ella is a knave if and only if Penelope is a knight''. So who is a knight and who is a knave?\\
\#\#\# Answer: Let's think step by step, by considering whether each person is lying and if that leads to contradiction. Assume Ella is a knight. Penelope cannot be a knight, because this would contradict the claim of their own. Penelope cannot be a knave, because this would contradict the false claim of their own. We have exhausted all possibilities for Penelope, so let us go back and reconsider Ella. Assume Ella is a knave. Penelope cannot be a knight, because this would contradict the false claim of Ella. Assume Penelope is a knave. This leads to a feasible solution.\\
CONCLUSION:\\
(1) Ella is a knave\\
(2) Penelope is a knave\\
\\

\#\#\# Question: \{question\}\\
\#\#\# Answer: Let's think step by step\\
\end{promptbox}

In our evaluation process, we use greedy decoding with temperature $t=0$ for all models and a maximum token length of 2048.  

To assess the correctness, we implement keyword matching: a response is considered correct if each person's ground truth identity appears in the conclusion part of the model's output.

\subsubsection{Fine-tuning}
\label{app:ft_details}

\paragraph{Training instance} Each training instance in Direct FT includes the task instruction, question, and the correct conclusion. In CoT FT, each training instance includes the task instruction, question, synthetic reasoning steps, and the correct conclusion. Both formats are structured similarly to task instructions followed by a single demonstration used in 1-shot Direct Prompting or 1-shot CoT Prompting.

\paragraph{Training loss}  In Direct FT, the loss for each training instance is computed on the tokens that appear directly after ``\#\#\# Answer:\textbackslash n''. 
 In CoT FT, the loss is calculated on the tokens that appear directly after 
``\#\#\# Answer: Let's think step by step''.

\paragraph{Training hyperparameters}  For \llamathree fine-tuning, we used \revise{LoRA} fine-tuning with the following standard hyperparameters: a batch size of 4,  gradient accumulation steps of 8, and 5e-5 learning rate.
\revise{The LoRA configuration was set as follows: rank $r = 32$, scaling factor $\alpha = 32$, and dropout rate $0.05$. No quantization techniques were used.}
We fine-tune for a maximum of 100 epochs.  We primarily reported results before 50 epochs, as we found the model typically converged by then.

For \gptfouromini fine-tuning, we utilized the default hyperparameters provided by the OpenAI fine-tuning API. The model was fine-tuned for 5 epochs to achieve high accuracy within reasonable budget.

\paragraph{Reported Training accuracy} For \gptfouromini, the training accuracy for each $N$-people \kk task is calculated using 100 training samples due to budget constraints on API usage. For open-source \llamathree, the training accuracy is based on the full set of training samples.

\subsubsection{Probing} 
\label{app:probing}

As described in \Cref{subsec:probing}, in the probing experiments, we train logistic regression models on the model’s intermediate outputs from different transformer blocks, to distinguish between correct and incorrect statements. For each transformer block, we extract the MLP layer’s output. 

The correct/incorrect statements consist of a \kk puzzle and a conclusion about a character’s role in the puzzle. For example, considering the following 2-people \kk puzzle:

\begin{promptbox} A very special island is inhabited only by knights and knaves. Knights always tell the truth, and knaves always lie. You meet 2 inhabitants: Oliver, and Ethan. Oliver told you that Oliver is a knight or Ethan is a knave. In a statement by Ethan: ``Oliver is a knight''. So who is a knight and who is a knave?
\end{promptbox}

with the correct answer being
\begin{promptbox} Oliver is a knight, and Ethan is a knight.
\end{promptbox}

We can generate two correct statements: 
\begin{itemize}
    \item A very special island is inhabited only by knights and knaves. Knights always tell the truth, and knaves always lie. You meet 2 inhabitants: Oliver, and Ethan. Oliver told you that Oliver is a knight or Ethan is a knave. In a statement by Ethan: ``Oliver is a knight''. So who is a knight and who is a knave? \textcolor{ForestGreen}{\bf Oliver is a knight.}
    \item A very special island is inhabited only by knights and knaves. Knights always tell the truth, and knaves always lie. You meet 2 inhabitants: Oliver, and Ethan. Oliver told you that Oliver is a knight or Ethan is a knave. In a statement by Ethan: ``Oliver is a knight''. So who is a knight and who is a knave? \textcolor{ForestGreen}{\bf Ethan is a knight.}
\end{itemize}

And two incorrect statements:
\begin{itemize}
    \item A very special island is inhabited only by knights and knaves. Knights always tell the truth, and knaves always lie. You meet 2 inhabitants: Oliver, and Ethan. Oliver told you that Oliver is a knight or Ethan is a knave. In a statement by Ethan: ``Oliver is a knight''. So who is a knight and who is a knave? \textcolor{OrangeRed}{\bf Oliver is a knave.}
    \item A very special island is inhabited only by knights and knaves. Knights always tell the truth, and knaves always lie. You meet 2 inhabitants: Oliver, and Ethan. Oliver told you that Oliver is a knight or Ethan is a knave. In a statement by Ethan: ``Oliver is a knight''. So who is a knight and who is a knave? \textcolor{OrangeRed}{\bf Ethan is a knave.}
\end{itemize}

\subsubsection{Distinguishing Memorization from Reasoning}

For \gptfouromini and \llamathree, we calculate the memorization score for each training sample within each complete $N$-people \kk training dataset. As discussed in \Cref{sec:mem-reason-classification}, we omit samples where $\accd[{x}]=0$ and label the remaining samples based on whether they are consistently solved under perturbation. We then split the dataset into 80\%/20\% train/test sets and perform binary classification.

\subsubsection{Computation Resources}
The fine-tuning  experiments are conducted on 2 NVIDIA A100 GPU cards, each with 80GB of memory.  The LLM evaluation experiments can be conducted on one NVIDIA RTX A6000 GPU card with 48 GB of memory. 

\section{Additional Experimental Results}
\label{app:add_exp_results}

\subsection{Benchmark Performance of Off-the-shelf Models}
\label{appsubsec:eval_benchmark}
\paragraph{Off-the-shelf models}

\begin{figure}[!htb]
    \centering
    \begin{minipage}{0.24\textwidth}
         \centering
         \includegraphics[width=\linewidth]{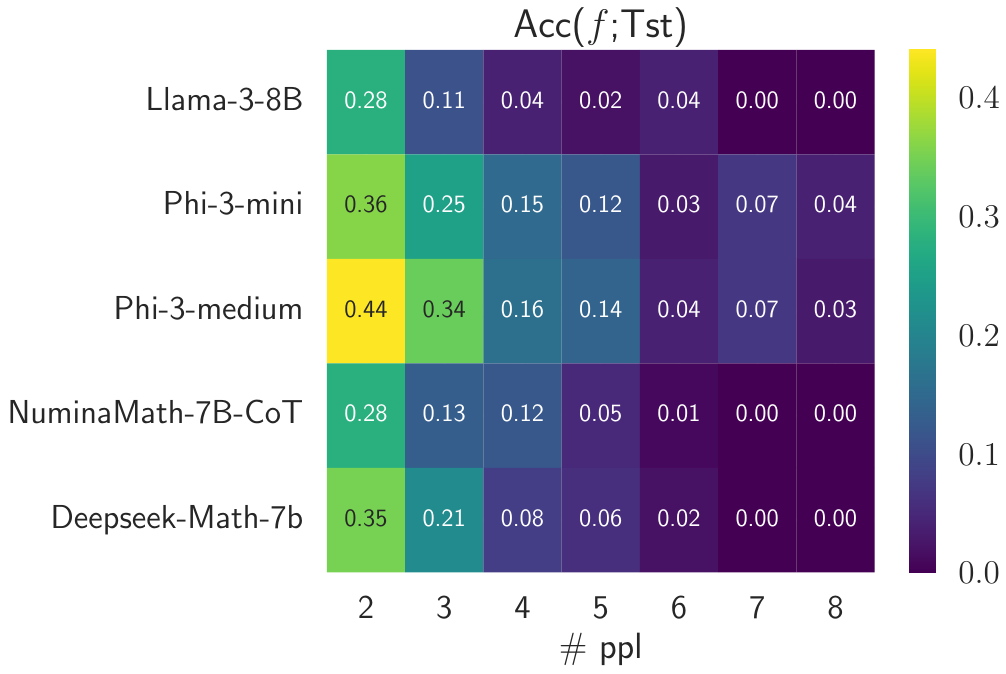}
    \subfigure{(a) 0-shot Direct prompting}
     \end{minipage}
    \begin{minipage}{0.24\textwidth}
         \centering
         \includegraphics[width=\linewidth]{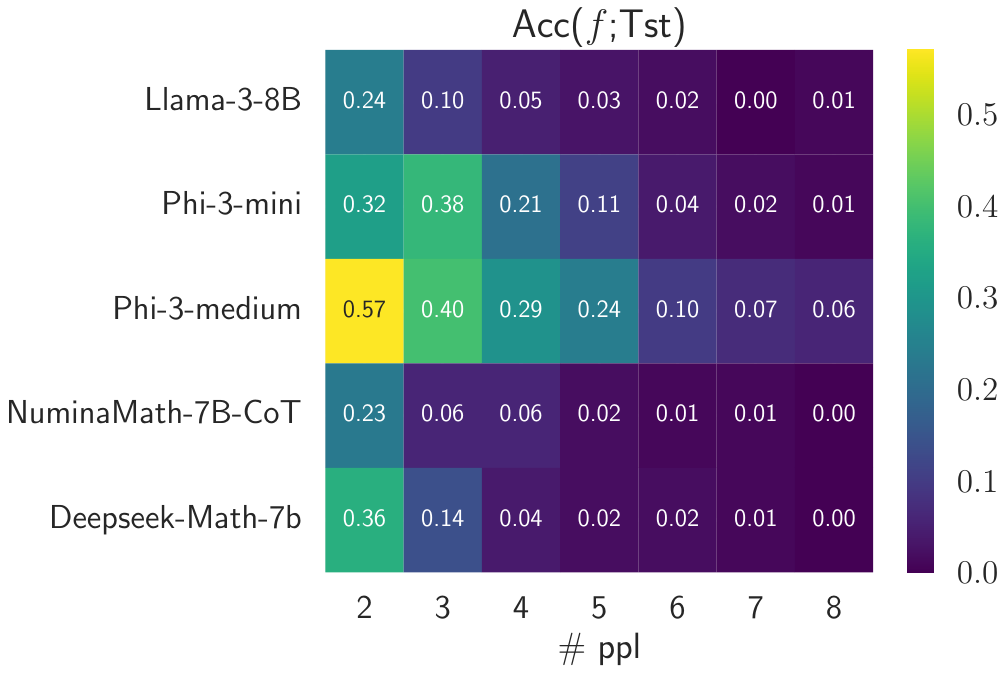}
    \subfigure{(b)  0-shot CoT prompting}
     \end{minipage} 
    \begin{minipage}{0.24\textwidth}
         \centering
         \includegraphics[width=\linewidth]{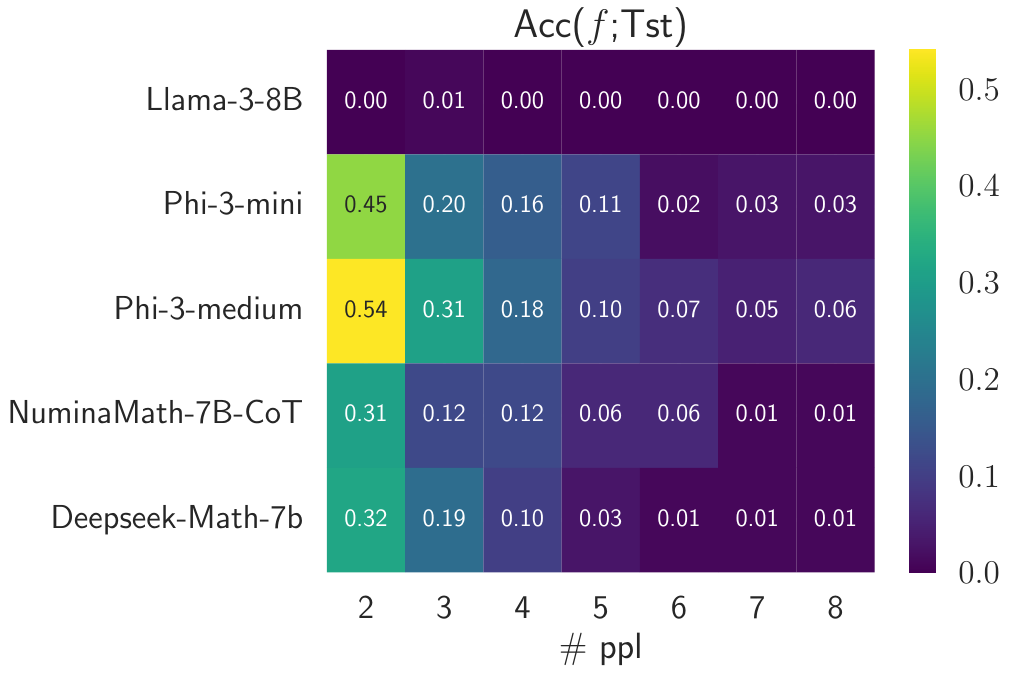}
    \subfigure{(c) 1-shot Direct prompting}
     \end{minipage} 
     \begin{minipage}{0.24\textwidth}
         \centering
         \includegraphics[width=\linewidth]{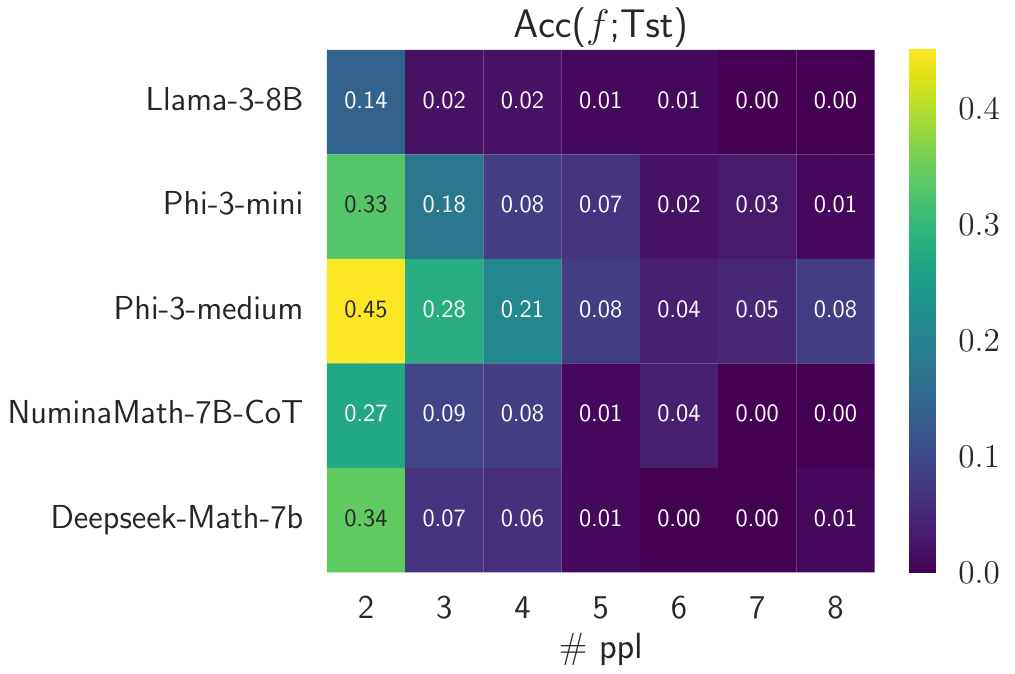}
         \subfigure{(d) 1-shot CoT prompting}
     \end{minipage} 
    \caption{$\ttacc$ and $\limemtt$  of off-the-shelf models under various prompt formats.}
    \label{fig:offtheshelf-prompt-format}
\end{figure}

We evaluate \llamathree, Phi-3-mini, Phi-3-medium, NuminaMath-7B-CoT, and Deepseek-Math-7B using 0/1-shot Direct/CoT prompting in \cref{fig:offtheshelf-prompt-format}. The results indicate that these open-source models exhibit poor accuracy on \kk tasks, particularly as the number of people in the \kk puzzles increases. Different prompting methods do not significantly enhance performance. 

Moreover, we evaluate \gptfouromini  under the self-consistency~\citep{wangself2023} where we query each puzzle 40 times under temperature 0.7. 
\cref{tab:selfconsistency} shows that self-consistency provides limited improvement on the 3-ppl task and fails to enhance performance on the more challenging 8-ppl task, suggesting that the model fundamentally struggles with solving such complex problems.

\begin{table}[ht]
    \centering
\caption{\revise{Self-consistency~\citep{wangself2023} can enhance the accuracy of \gptfouromini on the easy 2-ppl \kk task, but has limited improvement on 3-ppl task and fails to help on the challenging 8-ppl task, which suggests that the model cannot fundamentally solve such complex problems. 
}}
    \label{tab:selfconsistency}
    \resizebox{0.5\linewidth}{!}{
    \begin{tabular}{lccc}
    \toprule
    \multirow{2}{*}{Method} & \multicolumn{3}{c}{Test Accuracy} \\
    \cmidrule(lr){2-4}
    & 2-ppl & 3-ppl & 8-ppl \\
    \midrule
    Direct Prompting & 0.63 & 0.42 & 0.01 \\
    Direct Prompting + Self-consistency & 0.74 & 0.43 & 0.02 \\
    \bottomrule
    \end{tabular}
    }
\end{table}

\newpage
\subsection{Memorization Measurement}
\label{appsubsec:eval_mem}
\begin{figure}[!htb]
    \centering
        \begin{minipage}[!htb]{1\linewidth}
        \centering
        \includegraphics[width=1\linewidth]{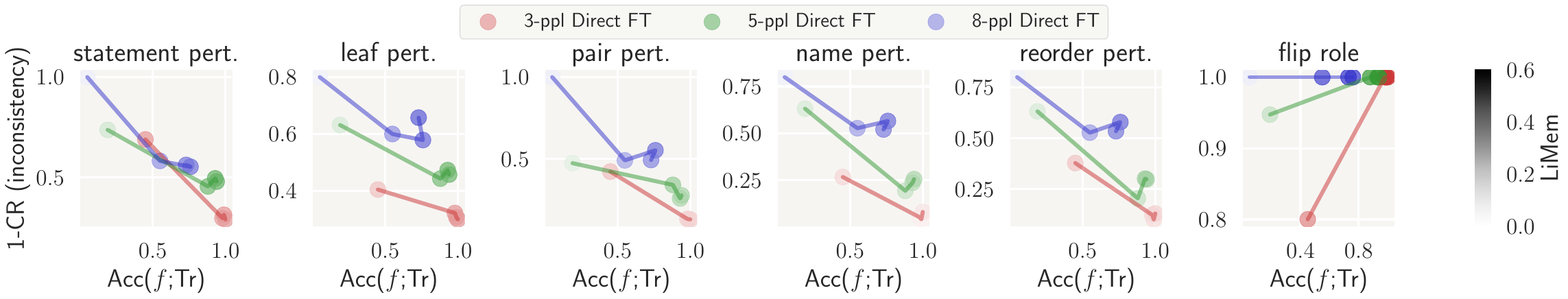}
        \includegraphics[width=1\linewidth]{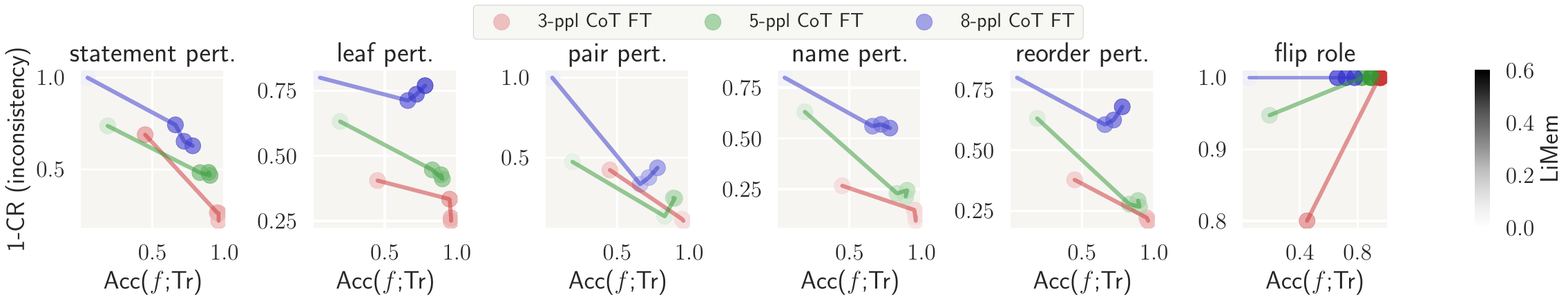}
        \caption{\small  Inconsistency ratio  ($y$-axis)  of fine-tuned \gptfouromini (first row: Direct FT; second row: CoT FT) decreases over training epochs  ($x$-axis), despite that the memorization becomes stronger as reflected by larger $\limemtr$ (deeper color). 
        }
        \label{fig:test_acc_mem_4omini}
    \end{minipage}
\end{figure}

\paragraph{Fine-tuned models}
As shown in \cref{fig:test_acc_mem_4omini},
the inconsistency ratio on correctly solved training puzzles ($y$-axis) of CoT-FTed or Direct-FTed \gptfouromini tends ot decreases over the training epochs ($x$-axis), despite that the memorization score $\limemtr$ on training samples also increases (i.e., a larger proportion of memorized samples in the training set).
The memorization score $\limemtr$ under role-flipping is significantly higher than other perturbation, possibly due to an internal bias that knights are truthful.

\begin{figure}[!htb]
    \centering
 \begin{minipage}[t]{\linewidth}
        \centering
         \includegraphics[width=0.32\linewidth]{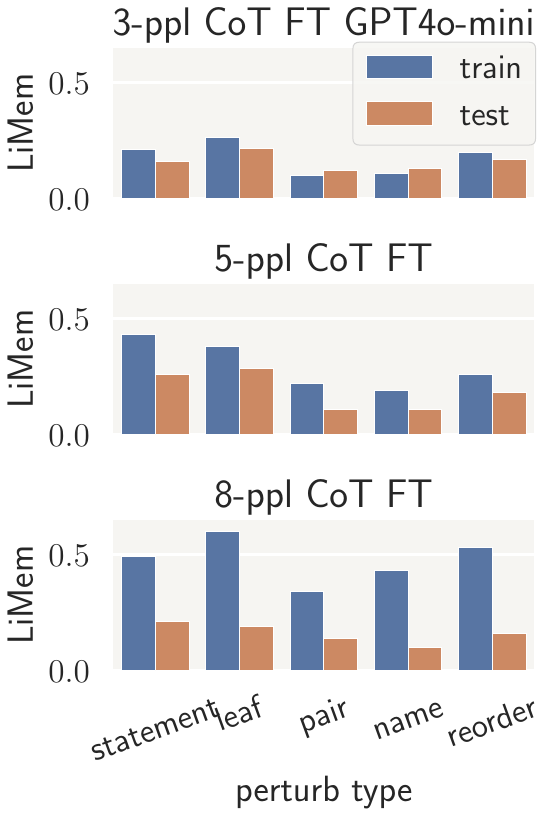}
        \includegraphics[width=0.32\linewidth]{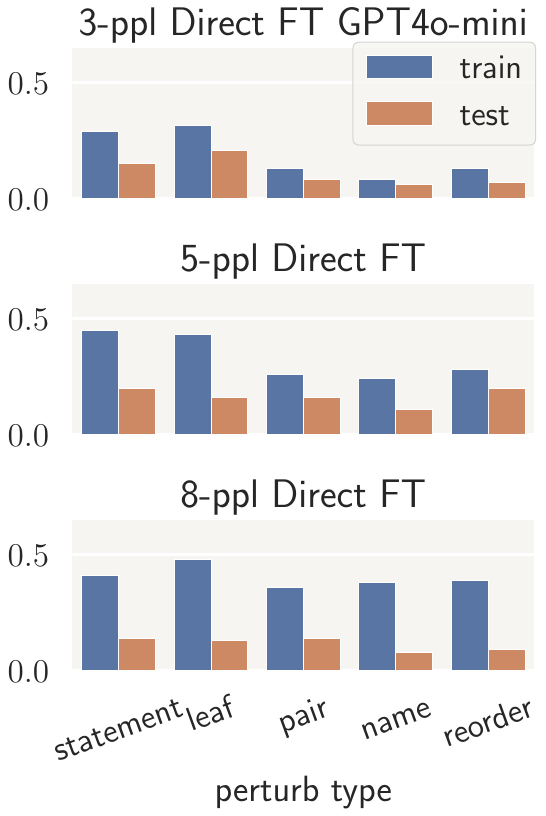}
        \includegraphics[width=0.32\linewidth]{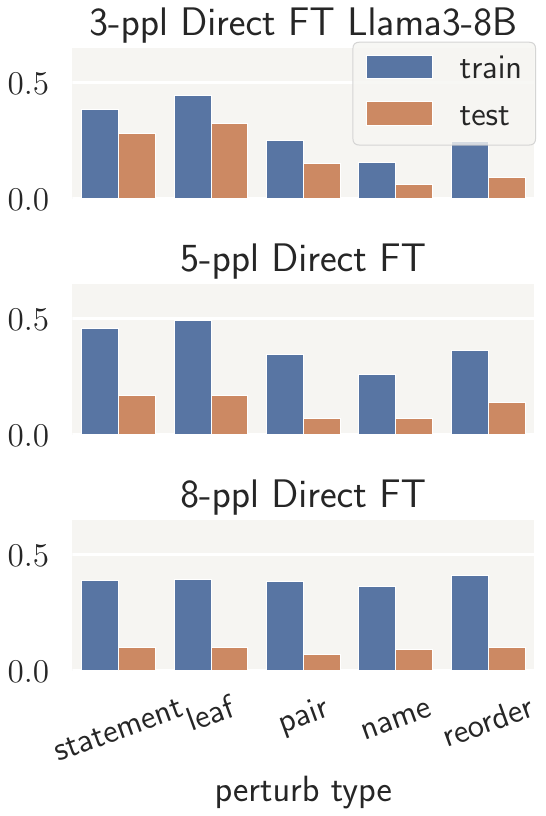}
        \vspace{-3mm}
        \caption{\small \revise{ Fine-tuned LLMs exhibit high memorization score on the training set under different perturbations, especially for hard tasks.
        The score on the test set can be smaller than on the training set.
        Models show stronger memorization under math-level perturbations compared to language-level perturbations.}
        }
    \label{fig:perturb_ablation_direct_ft_seperate_mem_score}
    \vspace{-3mm}
    \end{minipage}
\end{figure}

\begin{figure}[!htb]
    \centering
 \begin{minipage}[t]{\linewidth}
        \centering
         \includegraphics[width=0.32\linewidth]{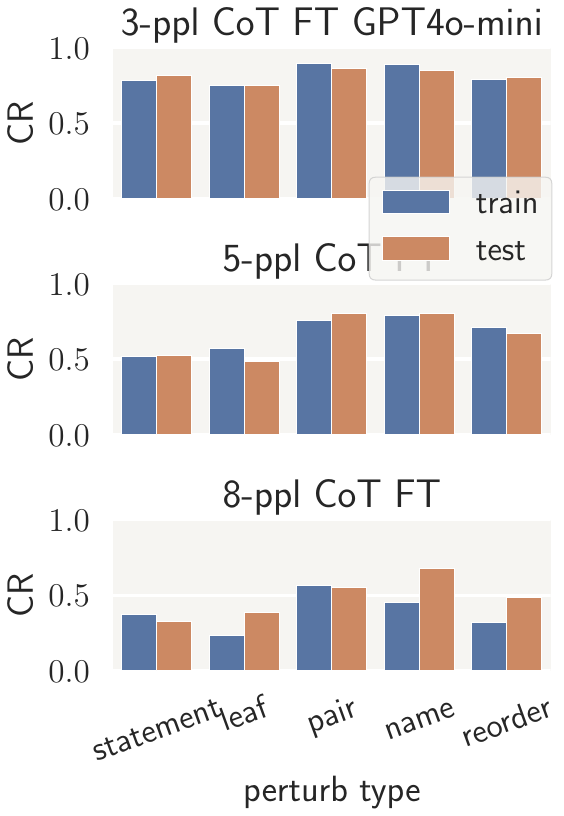}
        \includegraphics[width=0.32\linewidth]{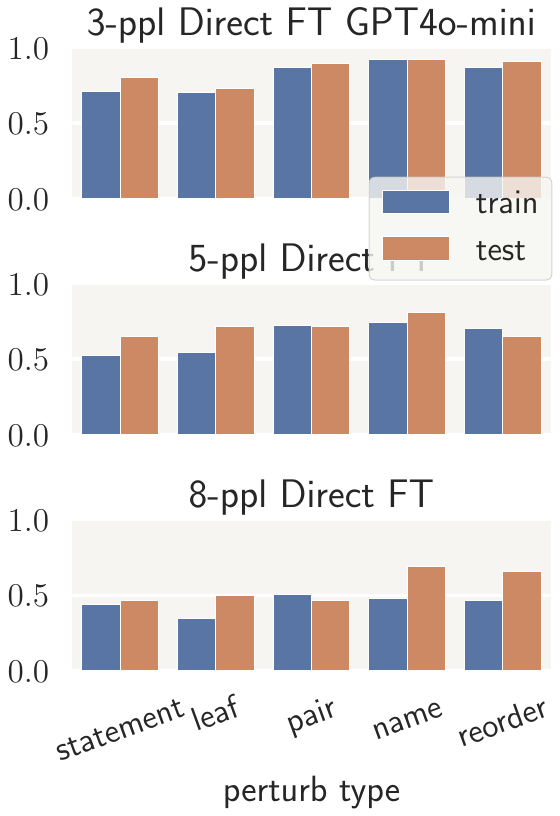}
        \includegraphics[width=0.32\linewidth]{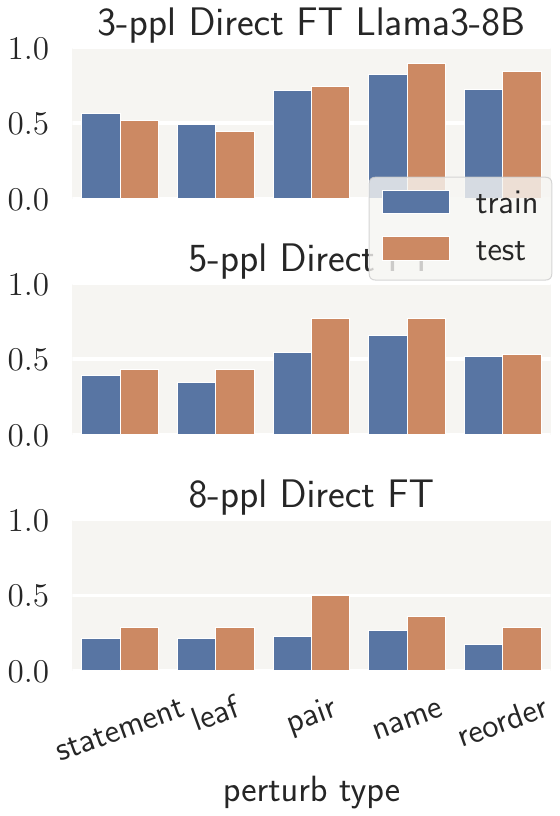}
        \vspace{-3mm}
        \caption{\small \revise{Consistency Ratio (CR $\uparrow$) under local perturbations. Fine-tuned LLMs generally demonstrate a higher consistency ratio on solved problems in the test set compared to the train set, particularly for challenging tasks such as 5/8-person puzzles. On the 3-person puzzle task, the consistency ratio between the train and test sets remains comparable. The consistency ratio generally is higher in easy tasks than in hard tasks.}
        }
        \label{fig:perturb_ablation_direct_ft_consist_ratio}
    \vspace{-3mm}
    \end{minipage}
\end{figure}

\begin{figure}[!htb]
    \centering
 \begin{minipage}[t]{\linewidth}
        \centering
         \includegraphics[width=0.4\linewidth]{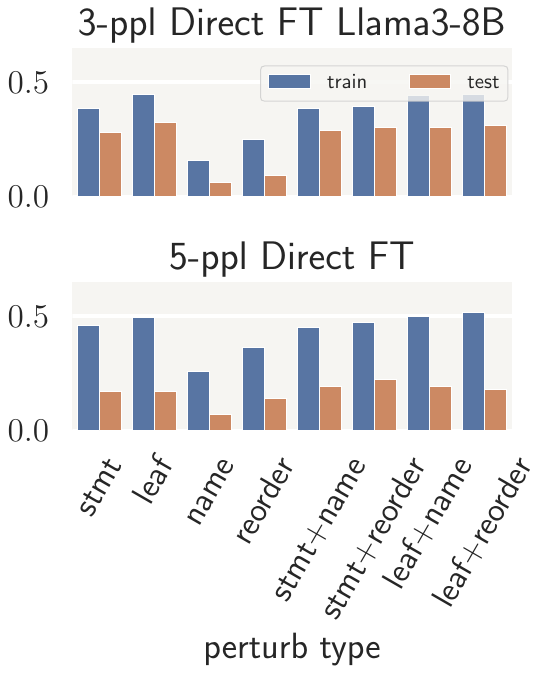}
        \vspace{-3mm}
       \caption{\small \revise{Memorization scores of Directly Fine-Tuned \llamathree under various math-level (statement, leaf) and language-level (name, reorder) perturbations. Combining math-level and language-level perturbations progressively can result in higher memorization scores (e.g., leaf + reorder), especially compared to applying language-level perturbations alone.}}
        \label{fig:perturb_ablation_direct_ft_stack_mem_score}
    \vspace{-3mm}
    \end{minipage}
\end{figure}

\clearpage
\subsection{Evaluation on Reasoning Capability}
\label{appsubsec:eval_reason}

\subsubsection{\llamathree}

\begin{figure}[!htb]
    \centering
    \begin{minipage}{\textwidth}
         \centering
         \includegraphics[width=\linewidth]{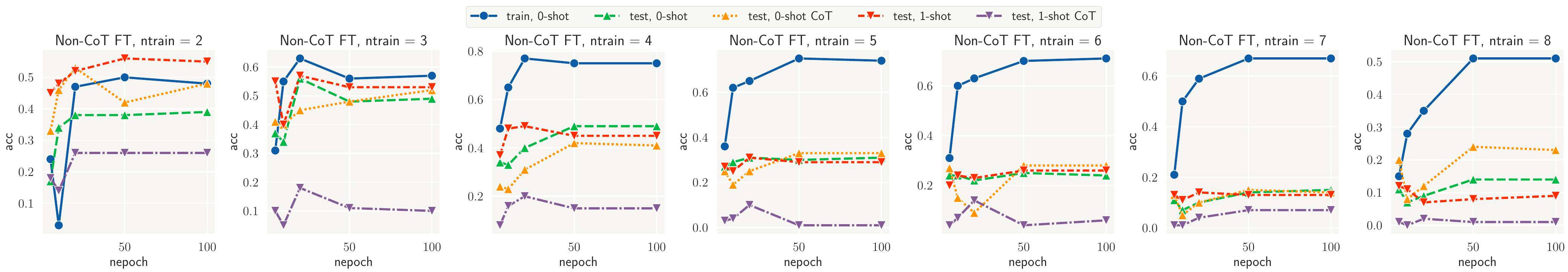}
         \subfigure{(a) Direct FT}
     \end{minipage} 
     \begin{minipage}{\textwidth}
         \centering
         \includegraphics[width=\linewidth]{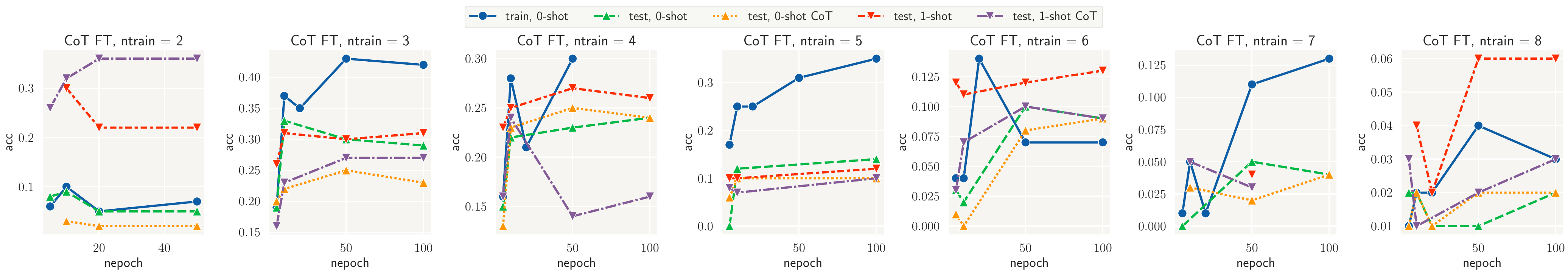}
         \subfigure{(b) CoT FT}
     \end{minipage}
    \caption{Train and test accuracy (under different evaluation configurations) for the \llamathree model fine-tuned on $N$-person tasks across multiple training epochs.
    }
    \label{fig:train-acc}
\end{figure}

\paragraph{Accuracy over epochs}
\Cref{fig:train-acc} reports the train and test accuracy (under different evaluation configurations) for the \llamathree model fine-tuned on $N$-person tasks across multiple training epochs.

\begin{figure}[!htb]
    \centering
    \begin{minipage}{\textwidth}
         \centering
         \includegraphics[width=\linewidth]{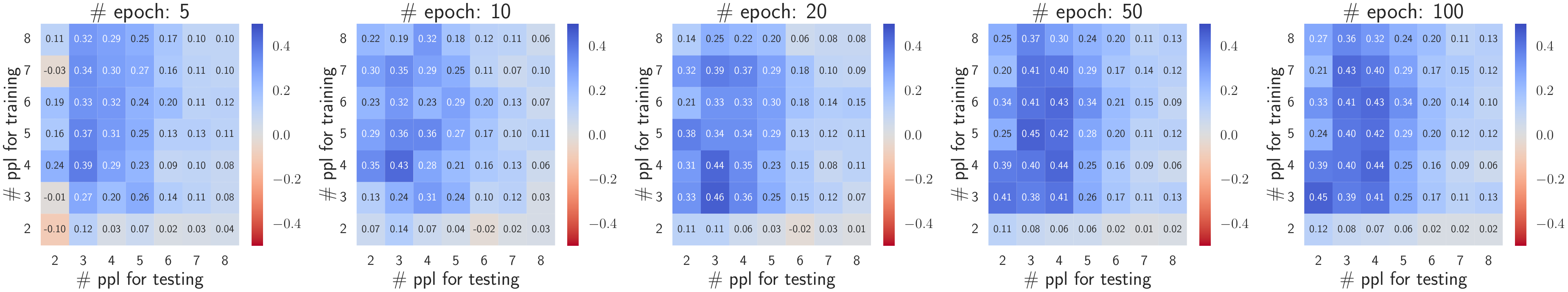}
         \subfigure{(a) 0-shot Direct Prompting}
     \end{minipage} 
    
    \begin{minipage}{\textwidth}
         \centering
         \includegraphics[width=\linewidth]{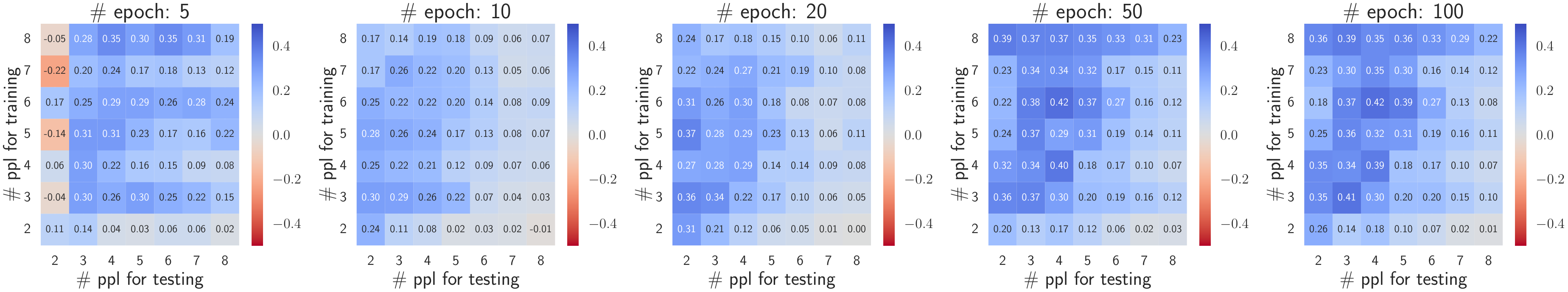}
         \subfigure{(b) 0-shot CoT Prompting}
     \end{minipage} 

     \begin{minipage}{\textwidth}
         \centering
         \includegraphics[width=\linewidth]{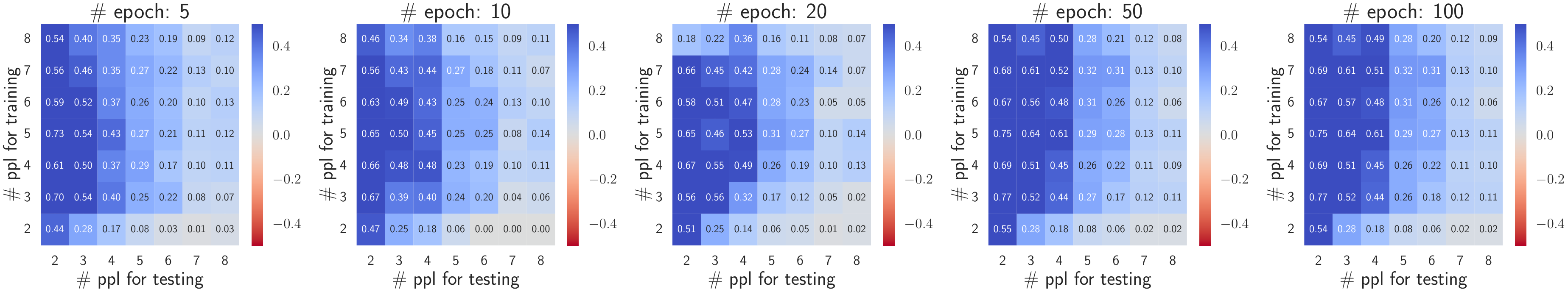}
         \subfigure{(c) 1-shot Direct Prompting}
     \end{minipage} 
    \caption{Improvement in test accuracy on $N$-person problems for \llamathree fine-tuned on $M$-person problems \textbf{with direct FT}, compared to the unfine-tuned model, under various evaluation configurations.}
    \label{fig:transfer-delta}
\end{figure}

\begin{figure}[!htb]
    \centering
    \begin{minipage}{\textwidth}
         \centering
         \includegraphics[width=\linewidth]{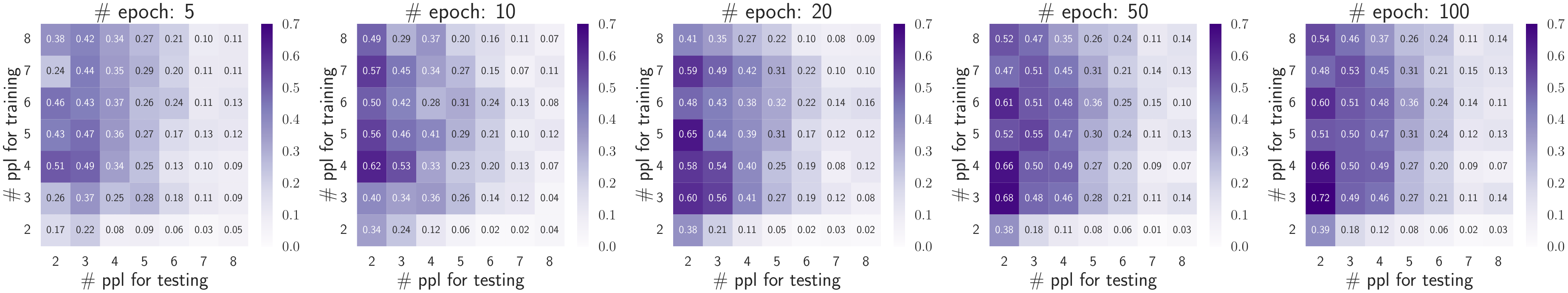}
         \subfigure{(a) 0-shot Direct Prompting}
     \end{minipage} 
    
    \begin{minipage}{\textwidth}
         \centering
         \includegraphics[width=\linewidth]{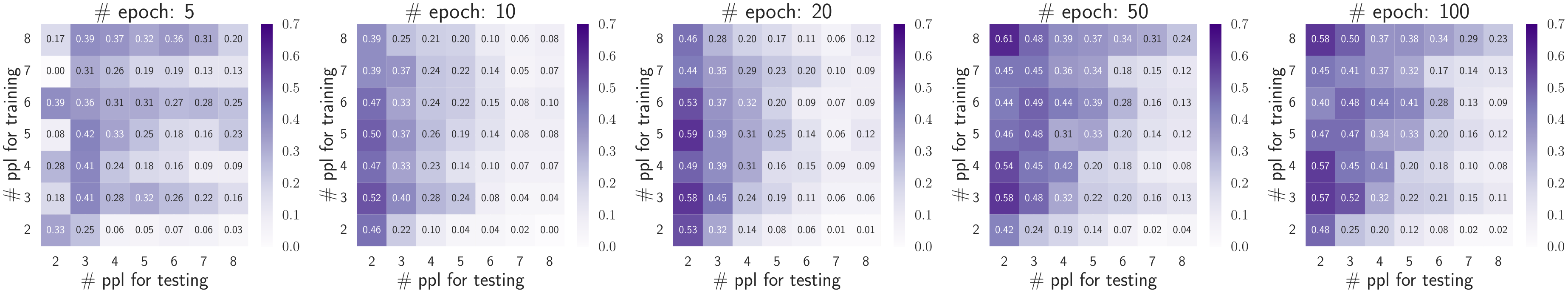}
         \subfigure{(b) 0-shot CoT Prompting}
     \end{minipage} 

     \begin{minipage}{\textwidth}
         \centering
         \includegraphics[width=\linewidth]{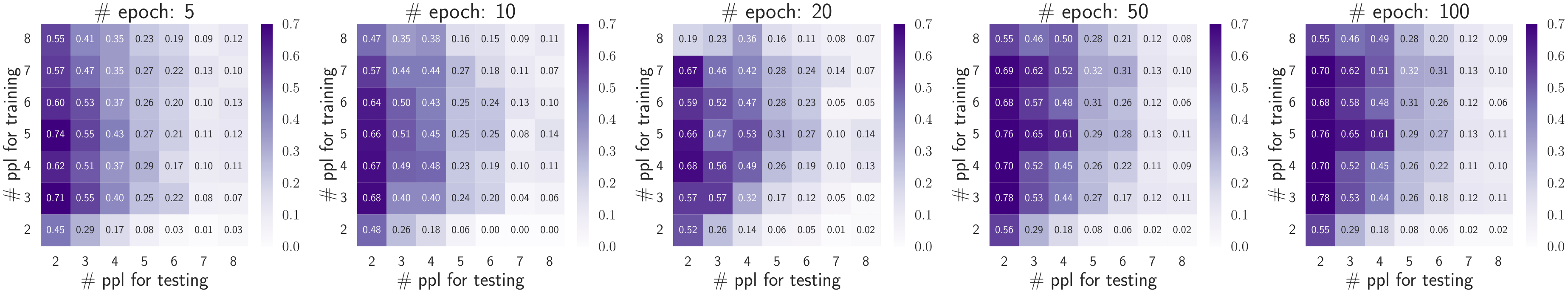}
         \subfigure{(c) 1-shot Direct Prompting}
     \end{minipage} 
    \caption{Test accuracy on $N$-person problems for \llamathree fine-tuned on $M$-person problems \textbf{with direct FT}, under various evaluation configurations.}
    \label{fig:transfer-absolute}
\end{figure}

\paragraph{Transferability}
We present the transferability results for the \kk task across different problem sizes and training epochs in \Cref{fig:transfer-delta} and \Cref{fig:transfer-absolute}. \Cref{fig:transfer-delta} shows the accuracy improvements relative to the baseline with no fine-tuning, while \Cref{fig:transfer-absolute} reports the absolute accuracy values.

\begin{figure}[!htb]
    \centering
    \includegraphics[width=\linewidth]{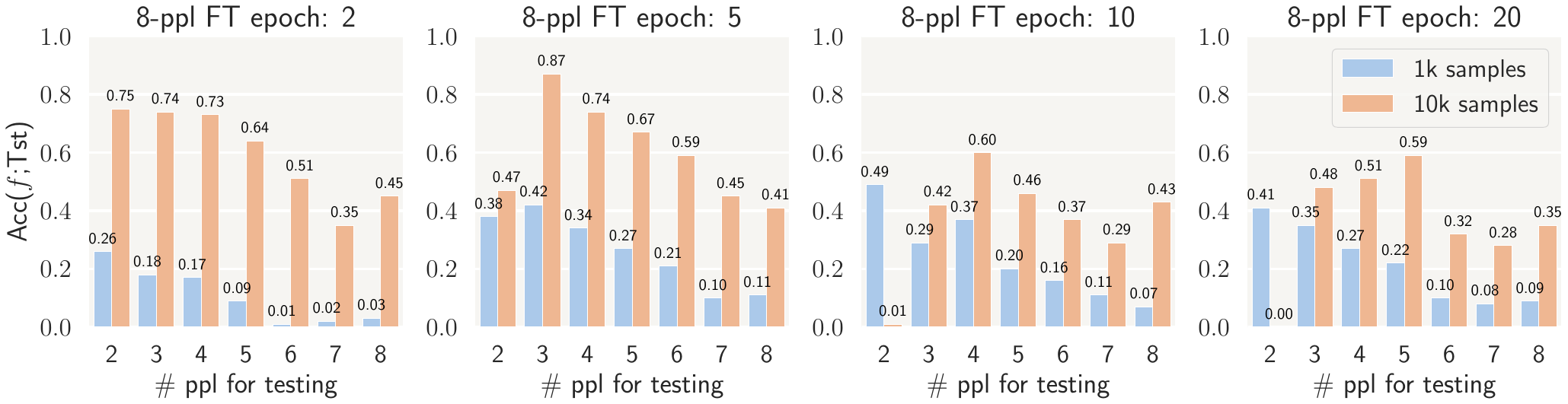}
    \caption{Transferability of \llamathree  Direct-FTed  on 1$k$/10$k$ samples at different epochs.}
    \label{fig:llama-10k-8ppl}
\end{figure}

\paragraph{Fine-tuning on 10$k$ 8-people \kk samples}
The results in \Cref{fig:llama-10k-8ppl} shows that $10k$ fine-tuning achieves significantly higher test accuracy than $1k$ fine-tuning on all tasks. 
\ft with $10k$ puzzles shows surprisingly high test accuracy, e.g., 87\% accuracy on 3-person tasks, where the un-FTed model has nearly 0 accuracy as shown in \cref{fig:eval_offshelf_lp_gap_test}. Notably, the models don't see reasoning steps during training and rely solely on memorizing answers. It also suggests that training on the hardest (8-person) tasks helps the model learn certain underlying rules that can be transferred to solve easier tasks.

However, the test accuracy drops for \llamathree  when {\ft}ing on $10k$ samples for overly long epochs, especially evaluated on 2-people \kk task, potentially due to overfitting to the more complicated 8-people training task.

\clearpage
\subsubsection{\gptfouromini}
\label{appsubsec:4omini_eval_reason}

\begin{figure}[!htb]
    \centering
    \begin{minipage}{0.48\textwidth}
         \centering
         \includegraphics[width=\linewidth]{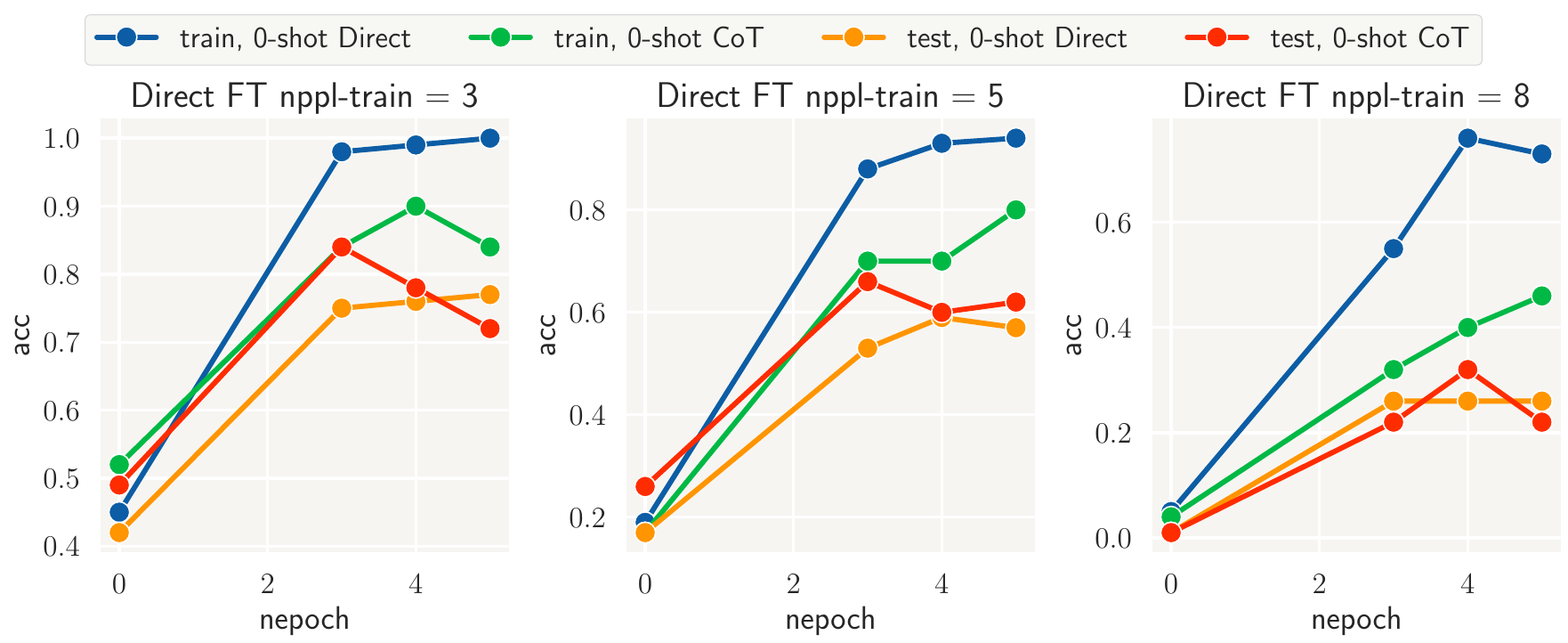}
         \subfigure{(a) Direct FT}
     \end{minipage} 
     \begin{minipage}{0.48\textwidth}
         \centering
         \includegraphics[width=\linewidth]{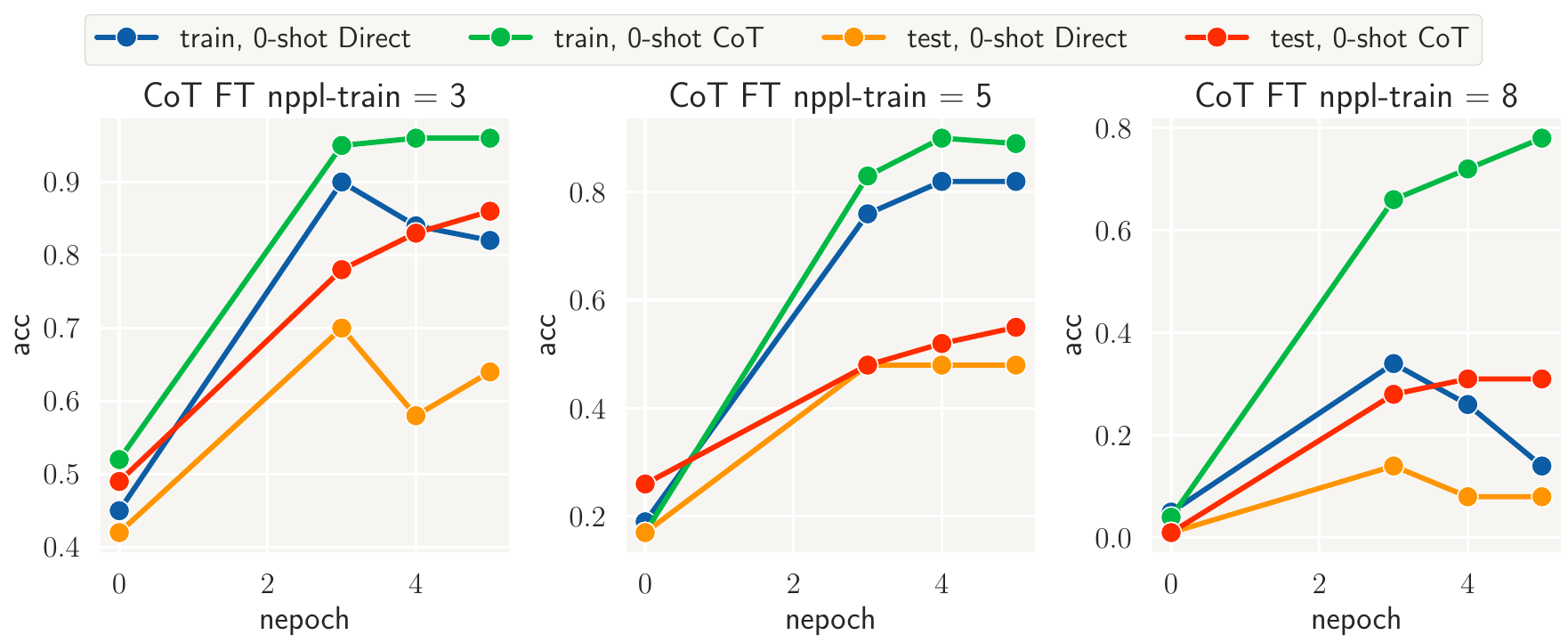}
         \subfigure{(b) CoT FT}
     \end{minipage}
    \caption{Train and test accuracy (under different evaluation configurations) for the \gptfouromini model fine-tuned on $N$-person tasks across multiple training epochs. }
    \label{fig:train-acc-4omini}
\end{figure}

\paragraph{Accuracy over epochs}
\Cref{fig:train-acc-4omini} reports the train and test accuracy (under different evaluation configurations) for the \gptfouromini model fine-tuned on $N$-person tasks across multiple training epochs.

Using the same paradigm for training and evaluation (i.e., \ft \& Direct Prompting,  \cotft \& CoT Prompting) usually achieves the best accuracy for \gptfouromini on training dataset and test dataset. 
We focus on 0-shot setting for \gptfouromini evaluation given its stronger capacity and higher accuracy than \llamathree.

\begin{figure}[!htb]
    \centering
     \begin{minipage}{\textwidth}
         \centering
         \includegraphics[width=0.7\textwidth]{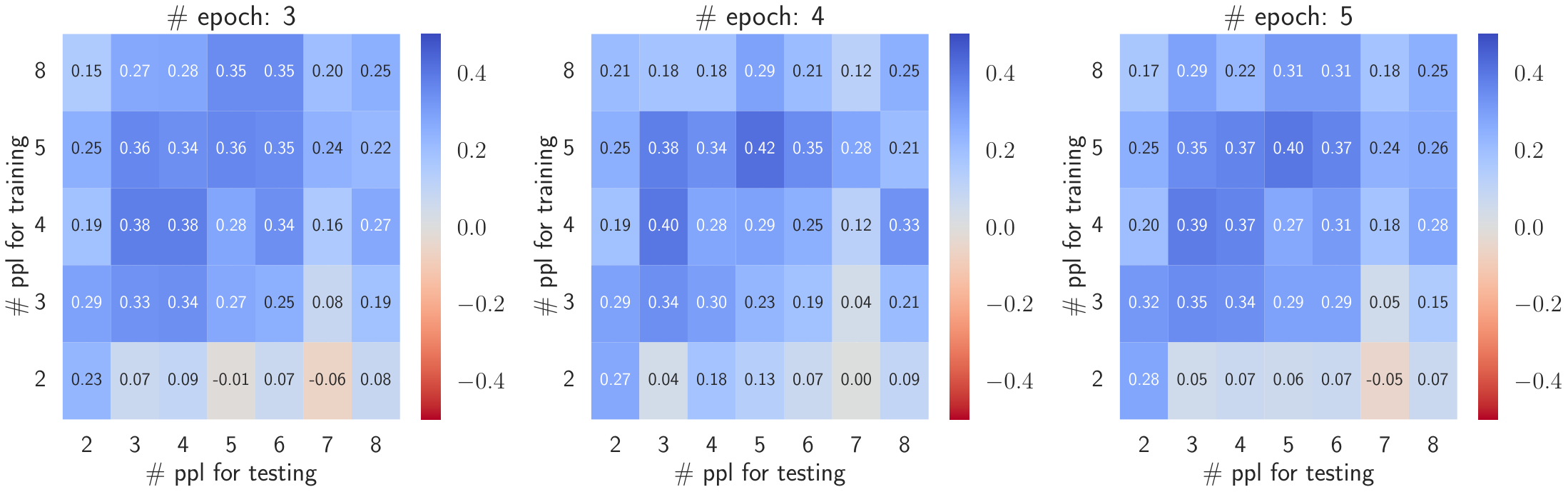}
         \subfigure{(a) \ft \& 0-shot Direct Prompting}
     \end{minipage} 
    \begin{minipage}{\textwidth}
         \centering
         \includegraphics[width=0.7\textwidth]{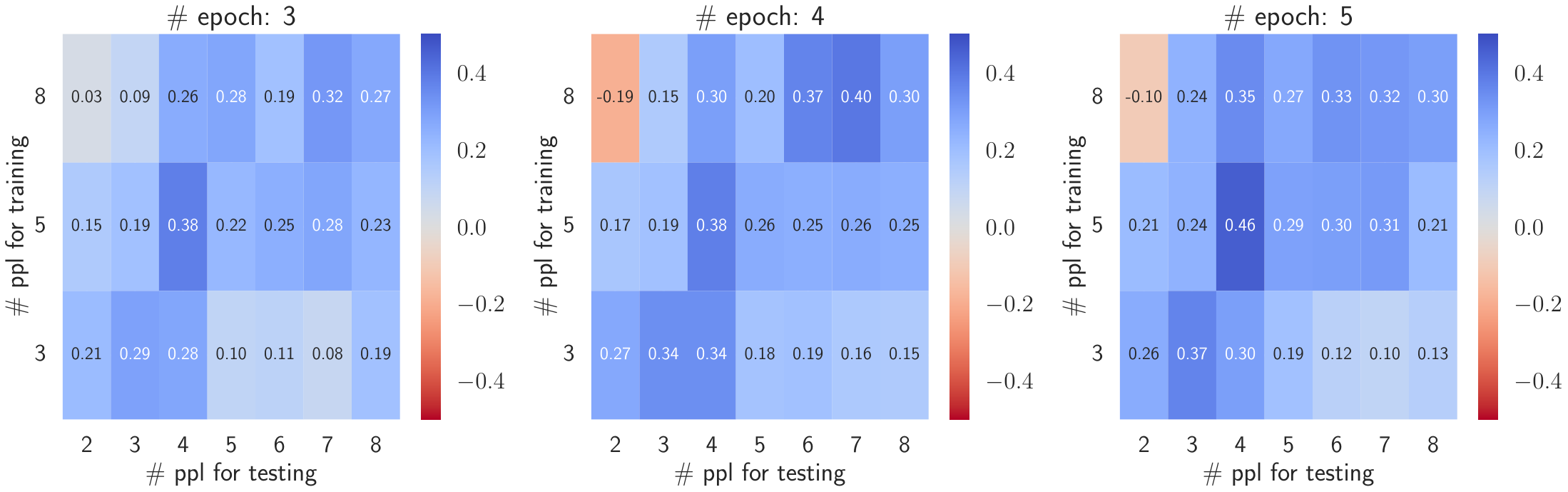}
         \subfigure{(b) \cotft \& 0-shot CoT Prompting}
     \end{minipage} 
    \caption{Improvement in test accuracy on $N$-person problems for \gptfouromini fine-tuned on $M$-person problems, under two finetuning/evaluation configurations.}
    \label{fig:transfer-delta-4omni}
\end{figure}

\begin{figure}[!htb]
    \centering
    \begin{minipage}{\textwidth}
         \centering
         \includegraphics[width=0.7\textwidth]{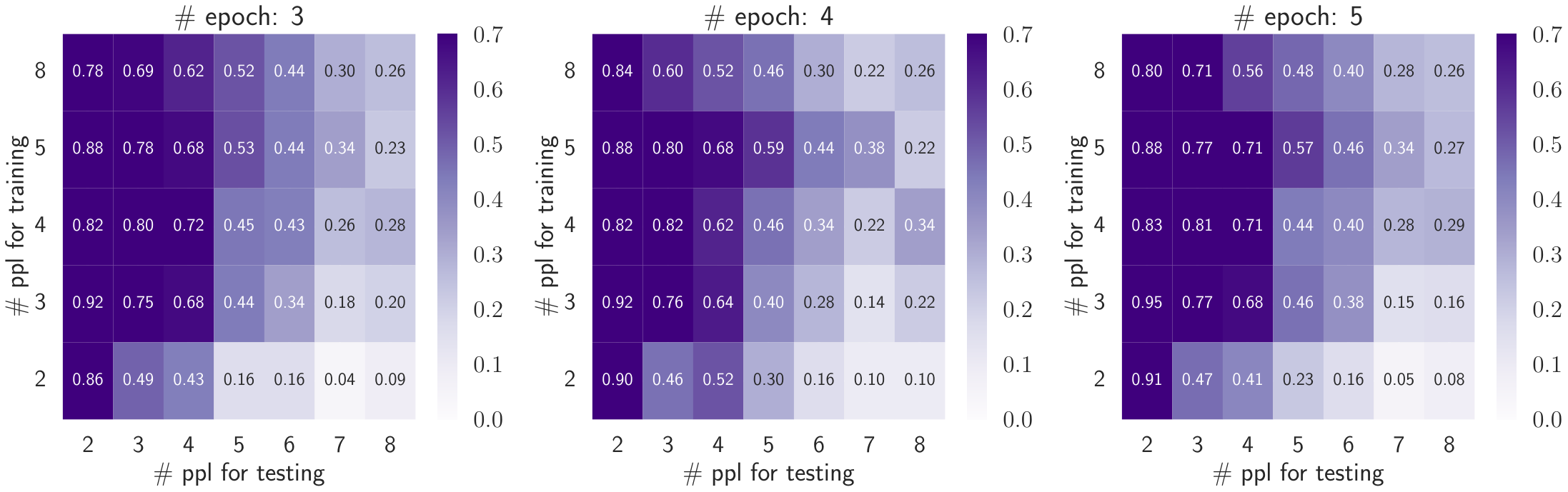}
         \subfigure{(a) \ft \& 0-shot Direct Prompting}
     \end{minipage} 
    
    \begin{minipage}{\textwidth}
         \centering
         \includegraphics[width=0.7\textwidth]{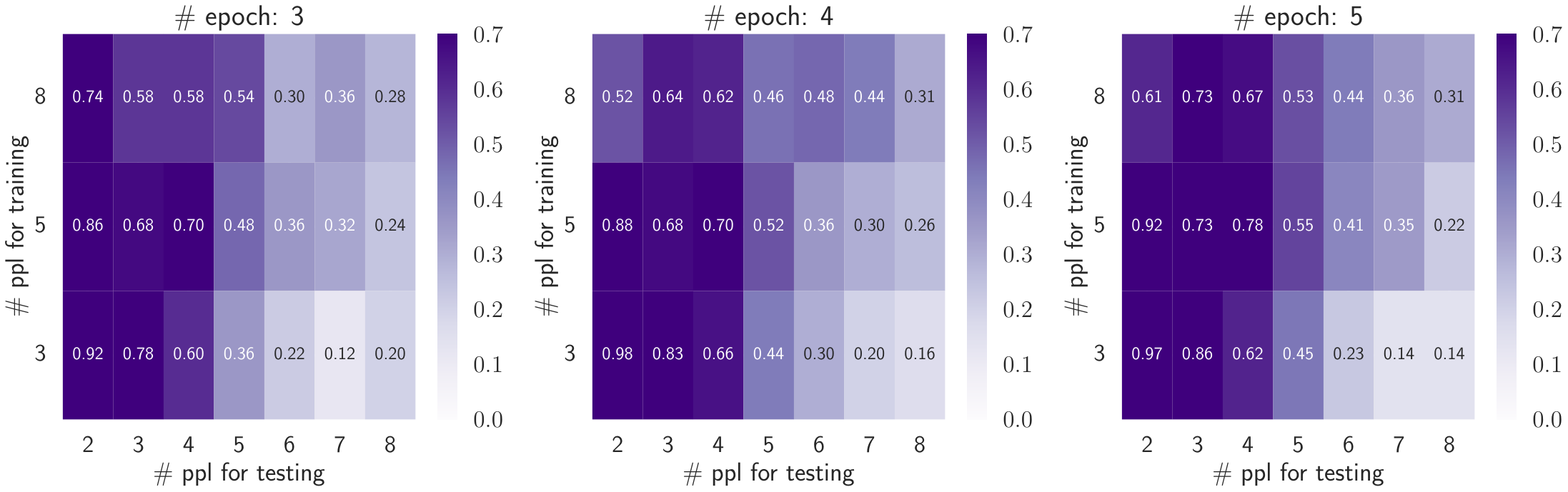}
         \subfigure{(b) \cotft \& 0-shot CoT Prompting}
     \end{minipage} 
    \caption{Test accuracy on $N$-person problems for \gptfouromini fine-tuned on $M$-person problems, under two finetuning/evaluation configurations.}
    \label{fig:transfer-absolute-4omni}
\end{figure}

\paragraph{Transferability}
We present the transferability results for the \kk task across different problem sizes and training epochs in \Cref{fig:transfer-delta-4omni} and \Cref{fig:transfer-absolute-4omni}. \Cref{fig:transfer-delta-4omni} shows the accuracy improvements relative to the baseline with no fine-tuning, while \Cref{fig:transfer-absolute-4omni} reports the absolute accuracy values.

\paragraph{Fine-tuning on 10$k$ 8-people \kk samples}

\begin{figure}[!htb]
    \centering
    \begin{minipage}{0.7\textwidth}
         \centering
         \includegraphics[width=\linewidth]{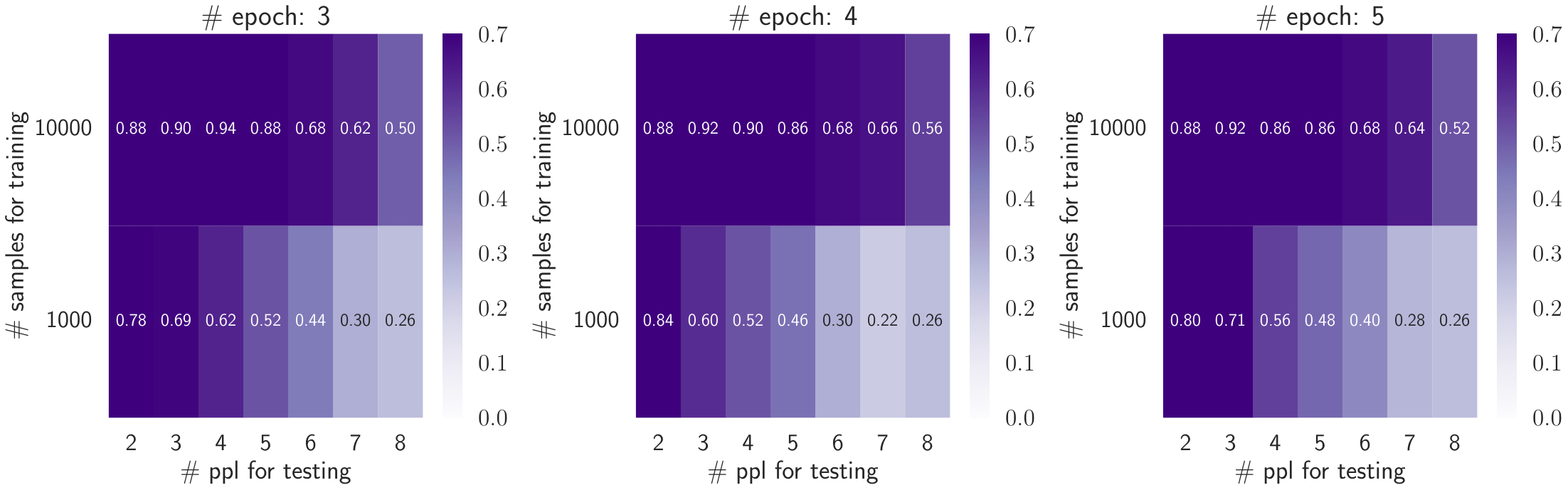}
         \subfigure{(a) \ft \& 0-shot Direct Prompting}
     \end{minipage} 
    \begin{minipage}{0.7\textwidth}
         \centering
         \includegraphics[width=\linewidth]{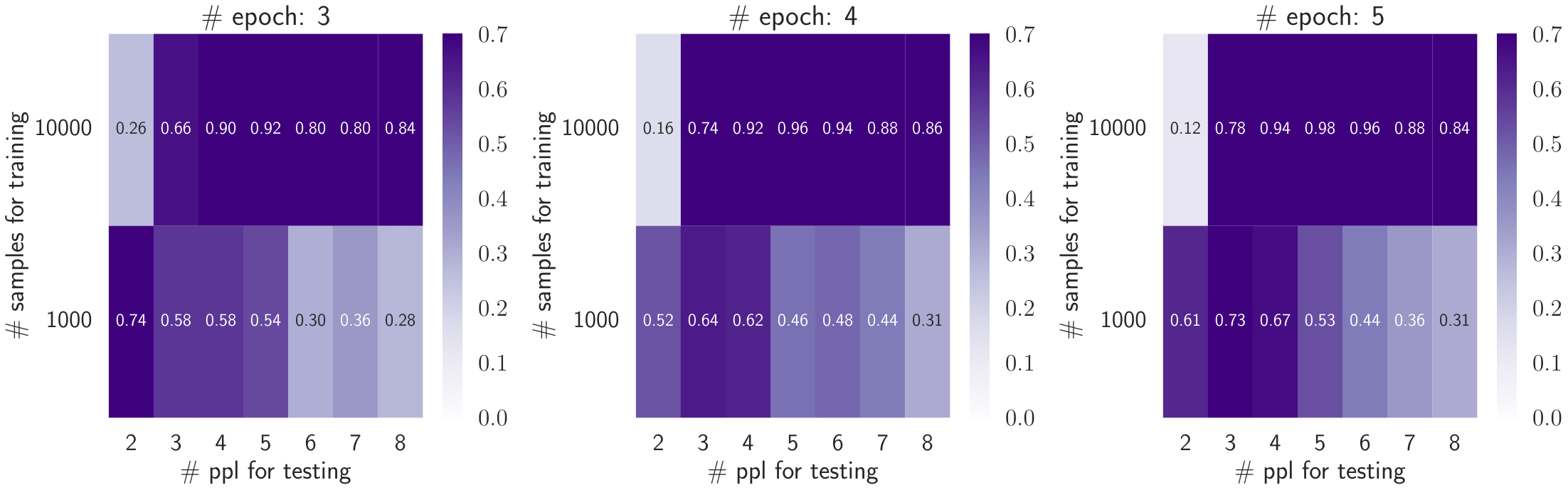}
         \subfigure{(b) \cotft \& 0-shot CoT Prompting}
     \end{minipage} 
    \caption{Test accuracy on $N$-person problems for \gptfouromini fine-tuned on 1$k$/10$k$ 8-people problems, under two finetuning/evaluation configurations.}
    \label{fig:transfer-absolute-4omni-8ppl-10k}
\end{figure}

We present the transferability results with absolute test accuracy for the \kk task across different 8-people task training sizes and training epochs in \Cref{fig:transfer-absolute-4omni-8ppl-10k}.  As shown, \gptfouromini achieves high accuracy on all tasks at early epochs (e.g., 3 epochs). 
We also find that \gptfouromini exhibits poor test accuracy on two-person testing puzzles when CoT-FTed on 10$k$ 8-people puzzles, unlike the Direct FTed model that have stable performance across all task.
In the failure case below, the CoT-FTed \gptfouromini gets stuck in a loop of listing assumptions and contradictions, resulting in long, repetitive responses without reaching a conclusion.

{\scriptsize
\begin{promptbox}[Failure case of \gptfouromini CoT FTed on 10$k$ 8-people \kk task]
\{omitted task instruction\}\\
\#\#\# Question: A very special island is inhabited only by knights and knaves. Knights always tell the truth, and knaves always lie. You meet 2 inhabitants: Logan, and Olivia. Logan was heard saying, ``Logan is a knight''. ``If Olivia is a knight then Logan is a knight'' - Olivia. So who is a knight and who is a knave?\\
\#\#\# Answer: Let's think step by step\\

\textcolor{gray}{\textit{// response from \gptfouromini}}\\
by considering whether each person is lying and if that leads to contradiction. Assume Logan is a knight. No contradiction is found in their claim that Logan is a knight. Assume Olivia is a knight. No contradiction is found in their claim that If Olivia is a knight then Logan is a knight. Assume this leads to no contradiction. No contradiction is found in their claim that If Olivia is a knight then Logan is a knight. This leads to no contradiction in their claim that Logan is a knight. This leads to no contradiction in their claim that If Olivia is a knight then Logan is a knight. This leads to no contradiction in their claim that Logan is a knight. This leads to no contradiction in their claim that If Olivia is a knight then Logan is a knight. This leads to no contradiction in their claim that If Olivia is a knight then Logan is a knight. This leads to no contradiction in their claim that Logan is a knight. This leads to no contradiction in their claim that If Olivia is a knight then Logan is a knight. This leads to no contradiction in their claim that Logan is a knight.
\end{promptbox}
}

\clearpage
\subsection{\ft with Wrong Answers}

\subsubsection{\llamathree}
\begin{figure}[!htb]
    \centering
    \begin{minipage}{\textwidth}
         \centering
         \includegraphics[width=\linewidth]{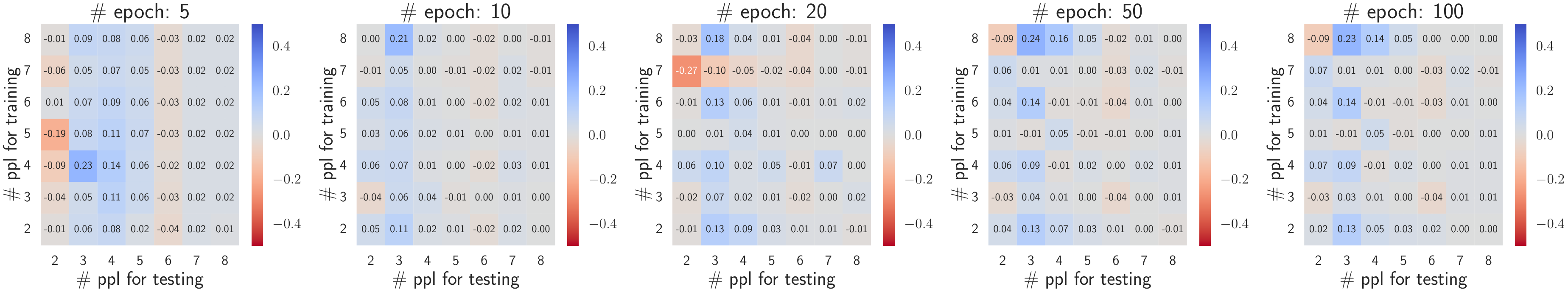}
        \subfigure{(a) 0-shot Direct Prompting}
     \end{minipage} 
    
    \begin{minipage}{\textwidth}
         \centering
         \includegraphics[width=\linewidth]{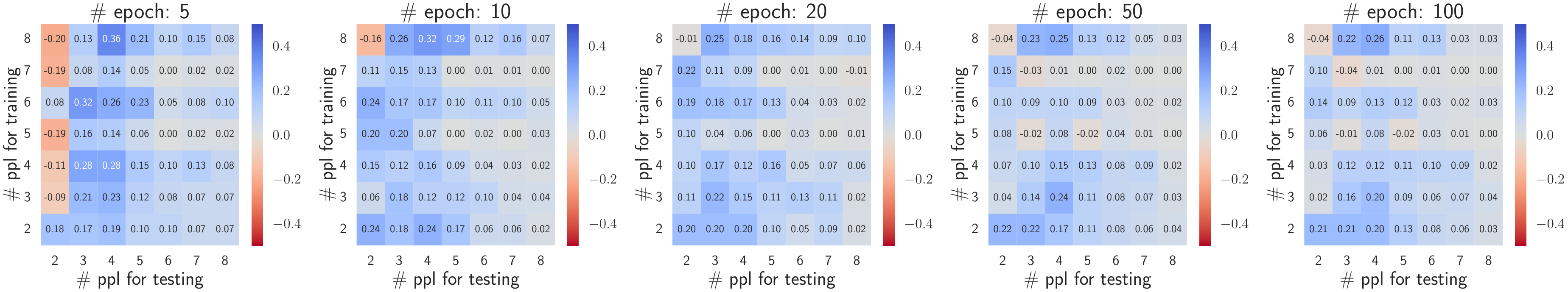}
         \subfigure{(b) 0-shot CoT Prompting}
     \end{minipage} 

     \begin{minipage}{\textwidth}
         \centering
         \includegraphics[width=\linewidth]{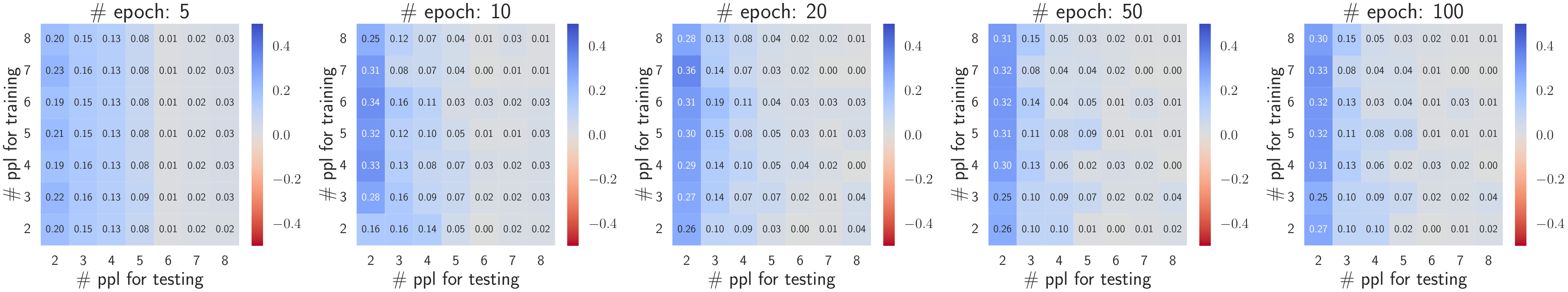}
         \subfigure{(c) 1-shot Direct Prompting}
     \end{minipage} 
    \caption{Improvement in test accuracy on $N$-person problems for \llamathree Direct FTed on $M$-person problems \textbf{with completely wrong answers}, compared to the unfine-tuned model, under various evaluation configurations.}
    \label{fig:wrong-ans-ft-100}
\end{figure}

\begin{figure}[!htb]
    \centering
    \begin{minipage}{\textwidth}
         \centering
         \includegraphics[width=\linewidth]{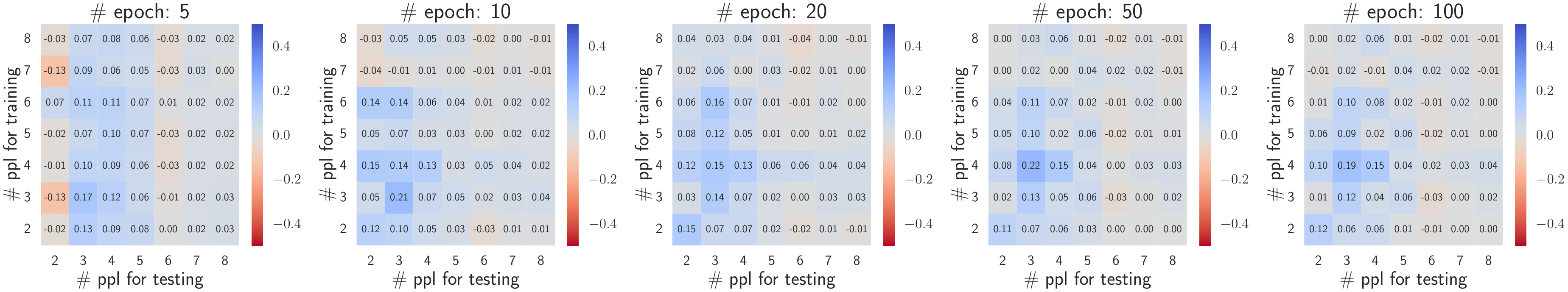}
         \subfigure{(a) 0-shot Direct Prompting}
     \end{minipage} 
    
    \begin{minipage}{\textwidth}
         \centering
         \includegraphics[width=\linewidth]{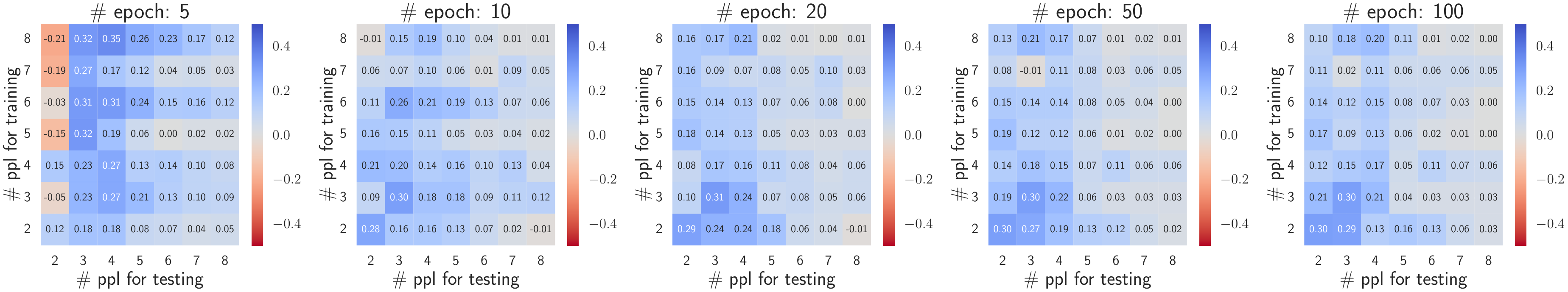}
         \subfigure{(b) 0-shot CoT Prompting}
     \end{minipage} 

     \begin{minipage}{\textwidth}
         \centering
         \includegraphics[width=\linewidth]{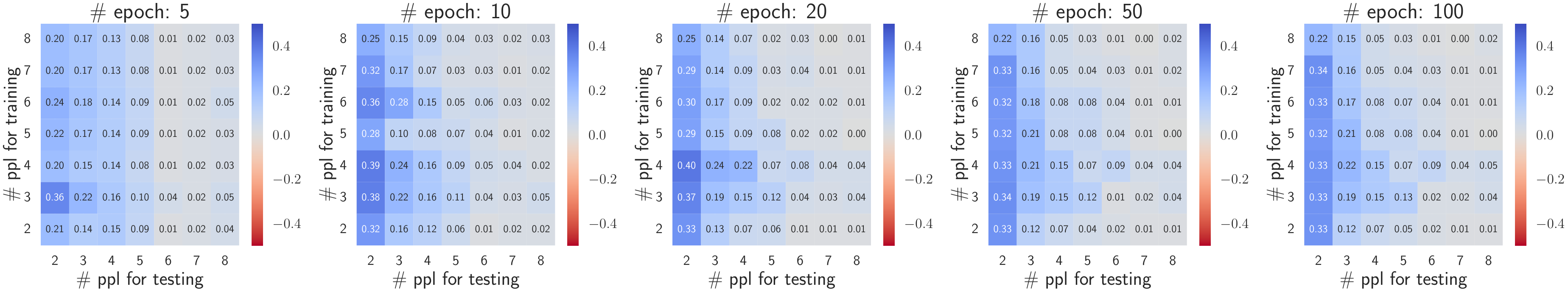}
         \subfigure{(c) 1-shot Direct Prompting}
     \end{minipage} 
    \caption{Improvement in test accuracy on $N$-person problems for \llamathree Direct FTed on $M$-person problems \textbf{with $\mathbf{75\%}$ wrong answers}, compared to the unfine-tuned model, under various evaluation configurations.}
    \label{fig:wrong-ans-ft-75}
\end{figure}

\begin{figure}[!htb]
    \centering
    \begin{minipage}{\textwidth}
         \centering
         \includegraphics[width=\linewidth]{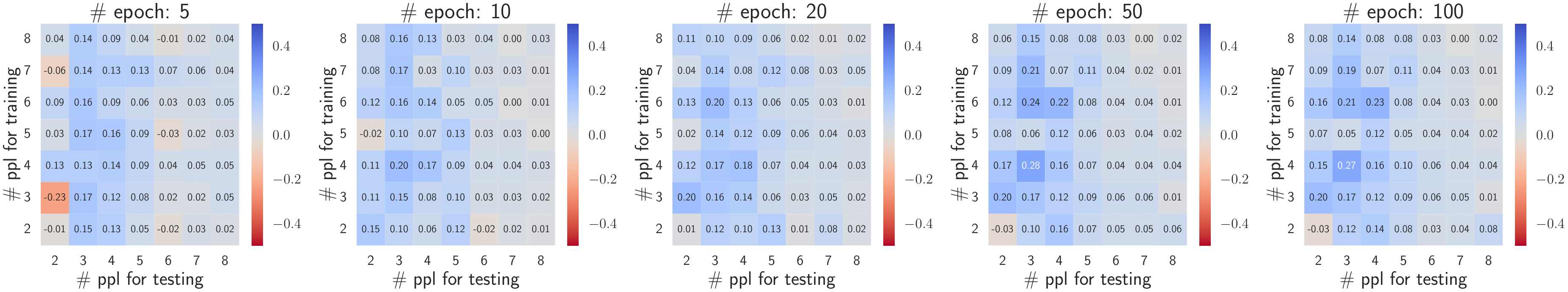}
         \subfigure{(a) 0-shot Direct Prompting}
     \end{minipage} 
    
    \begin{minipage}{\textwidth}
         \centering
         \includegraphics[width=\linewidth]{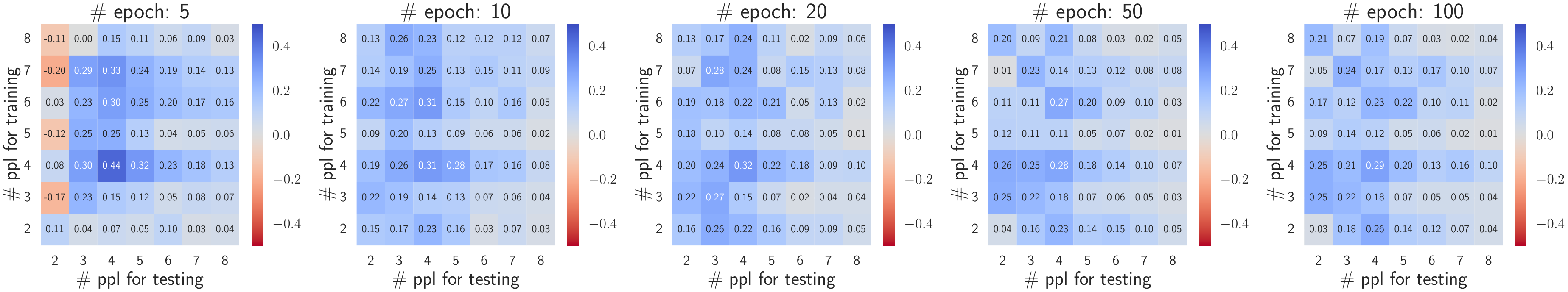}
         \subfigure{(b) 0-shot CoT Prompting}
     \end{minipage} 

     \begin{minipage}{\textwidth}
         \centering
         \includegraphics[width=\linewidth]{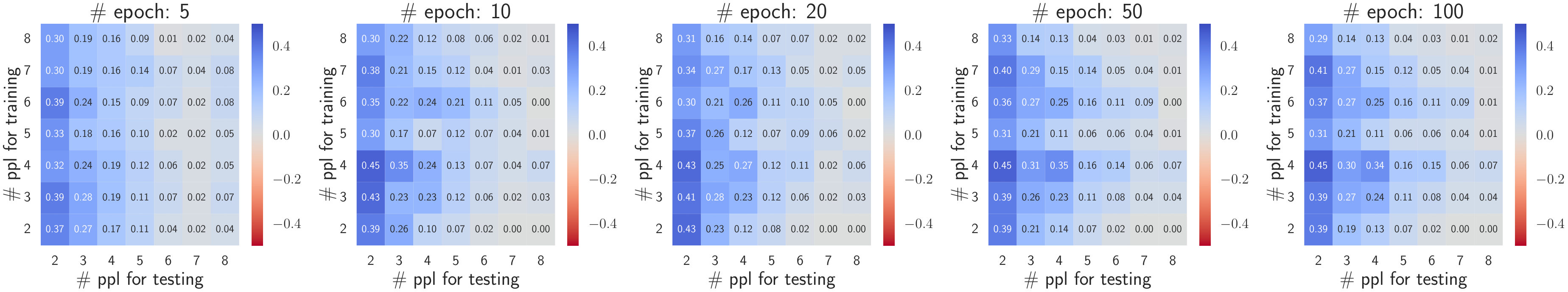}
         \subfigure{(c) 1-shot Direct Prompting}
     \end{minipage} 
    \caption{Improvement in test accuracy on $N$-person problems for \llamathree Direct FTed on $M$-person problems \textbf{with $\mathbf{50\%}$ wrong answers}, compared to the unfine-tuned model, under various evaluation configurations.}
    \label{fig:wrong-ans-ft-50}
\end{figure}

\Cref{fig:wrong-ans-ft-100}, \Cref{fig:wrong-ans-ft-75} and \Cref{fig:wrong-ans-ft-50} show the results of \ft with $100\%$, $75\%$ and $50\%$ incorrect answers for the \llamathree model across different prompting setups. Consistent with our earlier findings in \Cref{subsec:wrong-ans-ft}, fine-tuning with incorrect answers still significantly improves \kk performance, especially with 0-shot CoT prompting or 1-shot direct prompting.

\clearpage
\subsubsection{\gptfouromini}

\begin{figure}[!htb]
    \centering
    \includegraphics[width=\linewidth]{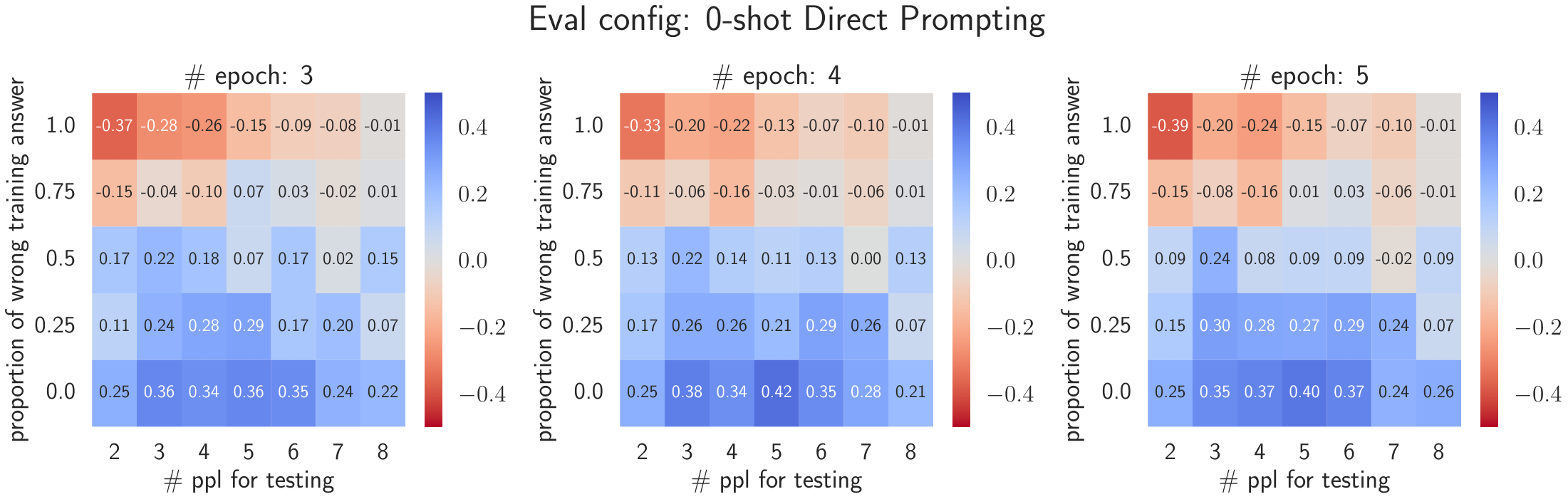}
    \caption{Improvement in test accuracy on $N$-people problems for \gptfouromini fine-tuned on 5-people problems with different proportion of wrong answers, compared to the unfine-tuned model.  \ft with 50\% wrong answers still improves \kk performance.}
    \label{fig:wrong-ans-ft-4omini-ep345}
\end{figure}

\Cref{fig:wrong-ans-ft-4omini-ep345} displays the results of direct fine-tuning using 5-people training \kk puzzles for the \gptfouromini model, containing varying percentages of incorrect answers in the dataset: $100\%, 75\%, 50\%,25\%, 0\%$. This is evaluated across different epochs in the five-person puzzle. As noted in \Cref{subsec:wrong-ans-ft}, when the training dataset includes 50\% or fewer samples with incorrect answers, fine-tuning can still enhance \kk's performance across various testing tasks.

\clearpage
\subsection{Probing}

\begin{figure}[!htb]
    \centering
    \includegraphics[width=\linewidth]{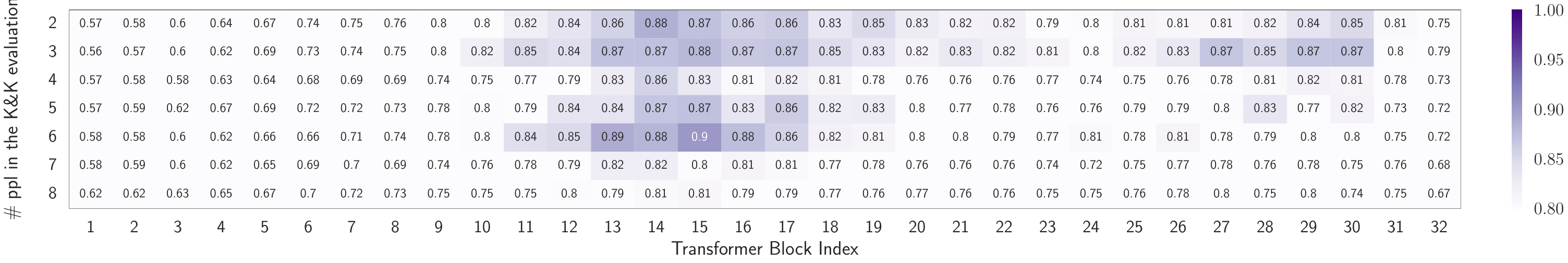}
    \caption{Probing accuracy of \kk puzzles with different number of people in testing puzzles across different layers of the un-finetuned \llamathree transformer model.
    }
    \label{fig:probe-base}
\end{figure}

We report the probing accuracy for the un-fine-tuned \llamathree model in \Cref{fig:probe-base}. As shown, without fine-tuning, the model demonstrates relatively low probing accuracy, with values usually below $90\%$.

\clearpage

\subsection{Distinguishing Memorization from Reasoning}

\begin{figure}[h]
    \centering  
    \begin{minipage}{\textwidth}
         \centering
         \includegraphics[width=\linewidth]{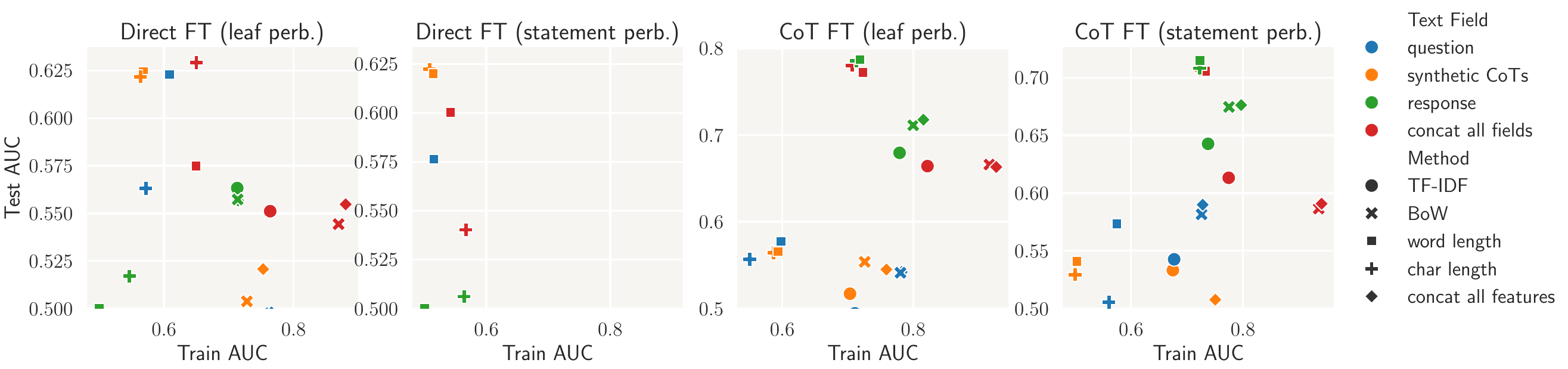}
          \subfigure{(a) 3-person puzzles for \gptfouromini.}
     \end{minipage}  
    \begin{minipage}{\textwidth}
         \centering
         \includegraphics[width=\linewidth]{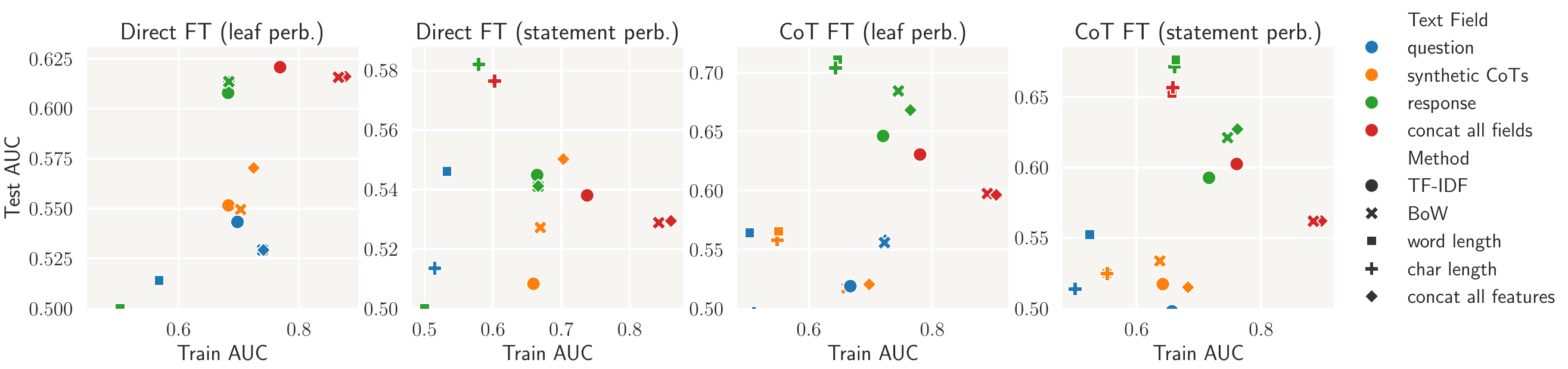}
          \subfigure{(b)5-person puzzles for \gptfouromini.}
     \end{minipage}  
     \begin{minipage}{\textwidth}
           \centering
         \includegraphics[width=0.8\linewidth]{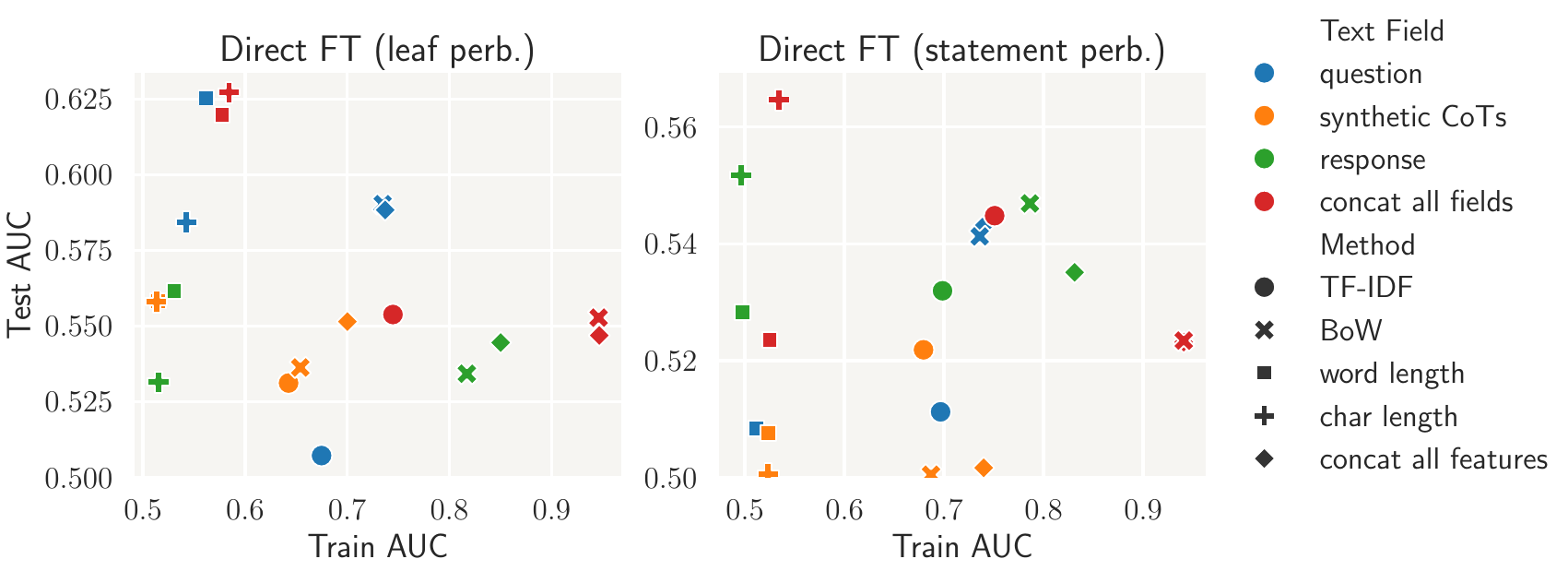}
          \subfigure{(c) 3-person puzzles for \llamathree.}
     \end{minipage}  
     \begin{minipage}{\textwidth}
           \centering
         \includegraphics[width=0.8\linewidth]{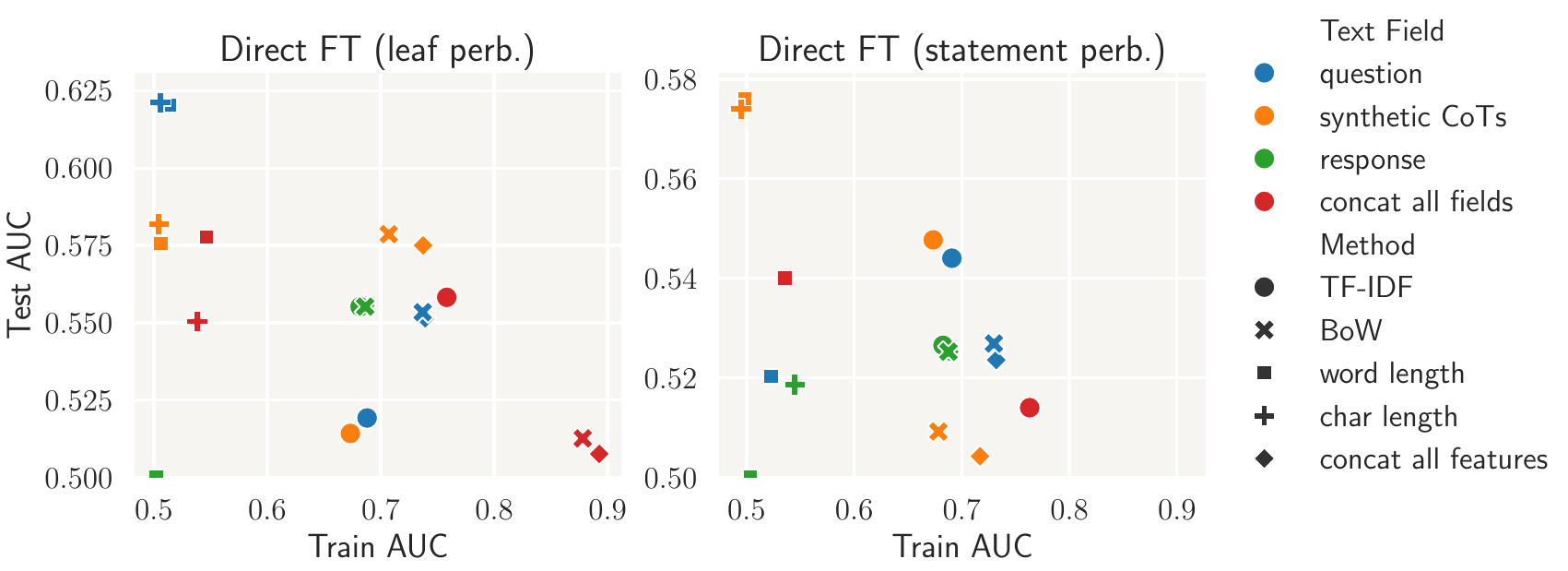}
          \subfigure{(d) 5-person puzzles for \llamathree.}
     \end{minipage}  
    \caption{AUC for predicting whether $N$-person puzzles can be consistently solved under perturbations based on puzzle-based indicators.}
    \label{fig:identify_robust_puzzle_more}
\end{figure}

\paragraph{Puzzle-based indicators}
\cref{fig:identify_robust_puzzle_more} shows the train and test AUC for predicting whether $N$-person puzzles can be consistently solved by a specific model under perturbations, using puzzle-based indicators. The results indicate that length-related features are useful for distinguishing memorization from reasoning. Notably, the test AUC is generally higher for CoT FTed \gptfouromini compared to Direct FTed \gptfouromini. 

\begin{figure}[ht]
    \vspace{-2mm}
    \centering
    \includegraphics[width=\linewidth]{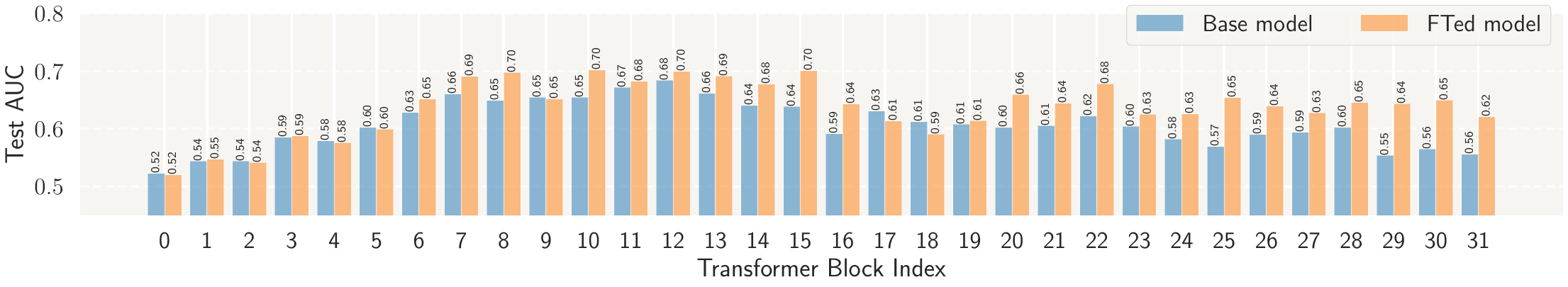}
    \vspace{-8mm}
    \caption{Test AUC for predicting 3-people puzzles based on whether they are consistently solved under leaf perturbation by the \llamathree model Direct-FTed. The embeddings across different layers of the fine-tuned \llamathree provide more distinguishable signals than those of the un-FTed model, leading to 0.7 AUC at the middle layers.
    Results under more tasks and perturbations are in \Cref{fig:identify_robust_llama_more}. 
    }
    \label{fig:identify_robust_llama}
    \vspace{-2mm}
\end{figure}

\begin{figure}[h]
    \centering
    
    \begin{minipage}{\textwidth}
         \centering
        \includegraphics[width=\linewidth]{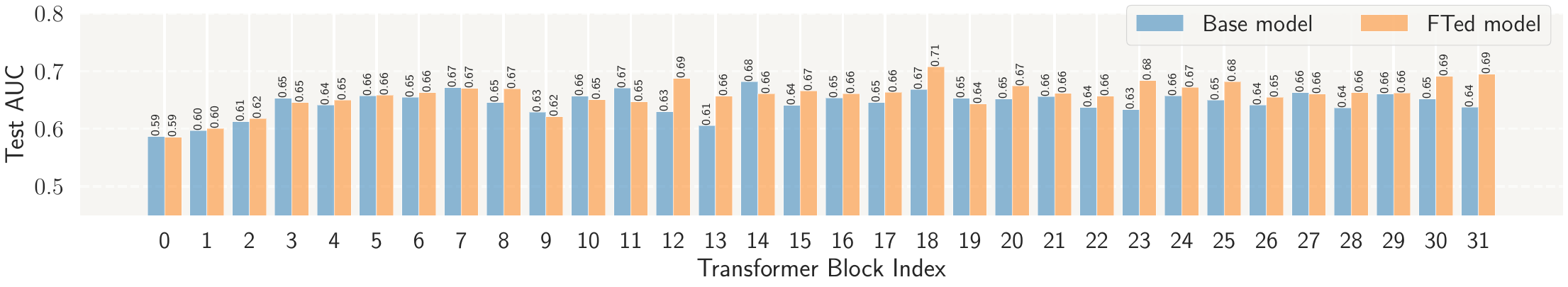}
         \subfigure{ (a) 3-person puzzles under statement perturbation.}
     \end{minipage}  
     \begin{minipage}{\textwidth}
         \centering
         \includegraphics[width=\linewidth]{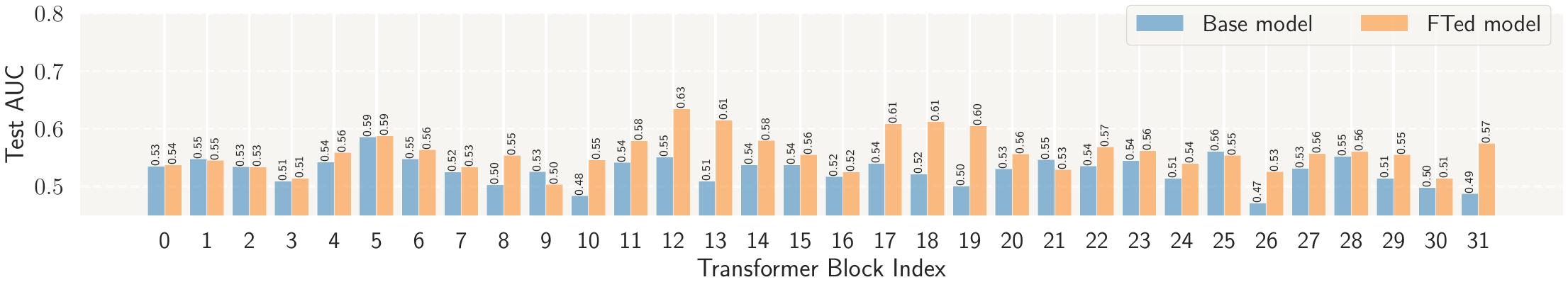}
         \subfigure{ (b) 5-person puzzles under leaf perturbation.}
     \end{minipage}      
    \begin{minipage}{\textwidth}
         \centering
        \includegraphics[width=\linewidth]{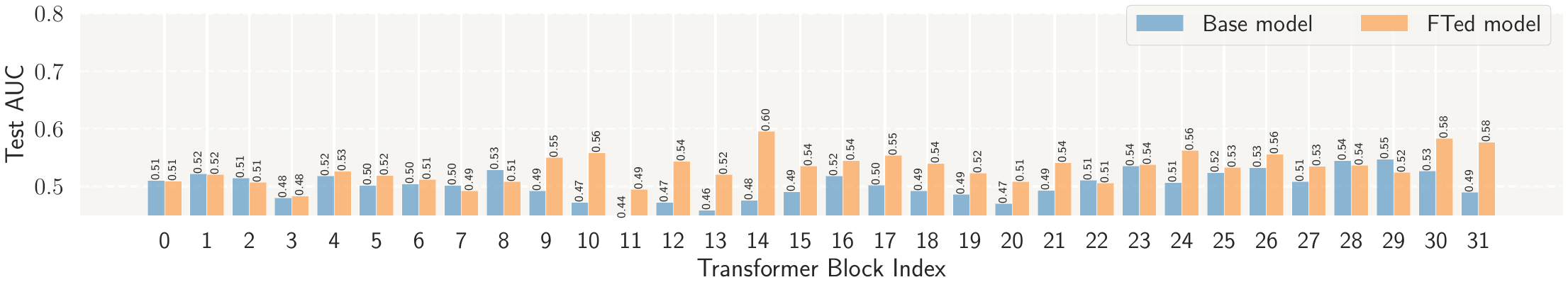}
         \subfigure{ (c) 5-person puzzles under statement perturbation.}
     \end{minipage}       
    \caption{Test AUC for predicting whether $N$-person puzzles can be consistently solved under perturbations by Direct-FTed \llamathree models.}
    \label{fig:identify_robust_llama_more}
\end{figure}

\paragraph{Model-based indicators}
We report test AUC for classifying puzzles based on whether they are consistently solved under leaf/statement perturbation by the \llamathree model Direct-FTed on the 3/5-person task. As shown in \cref{fig:identify_robust_llama} and \cref{fig:identify_robust_llama_more}, the embeddings across different layers of the fine-tuned \llamathree provide more distinguishable signals for memorized samples than those of the base model.

\end{document}